\documentclass[sigconf]{aamas} 

\usepackage{balance}        
\usepackage[utf8]{inputenc} 
\usepackage[T1]{fontenc}    
\usepackage{hyperref}       
\usepackage{url}            
\usepackage{booktabs}       
\usepackage{amsfonts}       
\usepackage{nicefrac}       
\usepackage{tikz}

\usepackage{graphics, graphicx}
\usepackage{xparse, xspace}
\usepackage[english]{babel}
\usepackage{upgreek}

\usepackage{tabularx}
\usepackage{enumerate, enumitem}
\usepackage{wrapfig}
\usepackage[subrefformat=parens, labelformat=parens]{subcaption}\usepackage{subcaption}

\usepackage{setspace}
\usepackage[linesnumbered,ruled,noend]{algorithm2e}
\SetKwBlock{Cycle}{}{}
\SetAlCapHSkip{0em}

\usepackage{adjustbox}
\usepackage{diagbox}
\usepackage{multirow}
\usepackage{colortbl}
\usepackage{makecell}
\usepackage{xcolor}
\usepackage{multicol}
\usepackage[strict]{changepage}

\setcopyright{ifaamas}
\acmConference[AAMAS '24]{Proc.\@ of the 23rd International Conference
on Autonomous Agents and Multiagent Systems (AAMAS 2024)}{May 6 -- 10, 2024}
{Auckland, New Zealand}{N.~Alechina, V.~Dignum, M.~Dastani, J.S.~Sichman (eds.)}
\copyrightyear{2024}
\acmYear{2024}
\acmDOI{}
\acmPrice{}
\acmISBN{}

\usepackage{xifthen}
\usepackage{amsmath}
\usepackage{amsfonts}

\newcommand\Tstrut{\rule[2ex]{0pt}{3pt}}    
\newcommand\Bstrut{\rule[-1.5ex]{0pt}{3pt}}   

\makeatletter
\protected\def\xvcenter{%
  \hbox\bgroup$\everyvbox{\everyvbox{}\aftergroup\m@th\aftergroup$\aftergroup\egroup}%
  \vcenter
}

\DeclareRobustCommand{\midscript}[1]{
  \mathchoice{\mid@script\scriptstyle{#1}}
    {\mid@script\scriptstyle{#1}}
    {\mid@script\scriptscriptstyle{#1}}
    {\mid@script\scriptscriptstyle{#1}}
}
\newcommand{\mid@script}[2]{
  \vcenter{\hbox{$\m@th#1#2$}}
}

\makeatletter

\usepackage{color}

\usepackage{pifont}
\newcommand{\cmark}{\ding{51}}%
\newcommand{\xmark}{\ding{55}}%
\newcommand{\bad}{{\color{red!80!black}\scalebox{1.25}{\xmark}}}    
\newcommand{\good}{{\color{green!80!black}\scalebox{1.25}{\cmark}}} 

\makeatletter
\newcommand{\colorlabel}[1]{%
	\global\tag@true%
	\nonumber%
	\refstepcounter{equation}%
	\gdef\df@tag{\maketag@@@{{\color{#1}(\theequation)}}\def\@currentlabel{\theequation}}}
\makeatother

\newcommand{\bigmid}{\;\ifnum\currentgrouptype=16 \middle\fi|\;} 
\newcommand{\mytilde}[0]{\mathds{\raise.17ex\hbox{$\scriptstyle\sim$}}} 

\DeclareMathOperator{\EV}{\mathbb{E}}
\newcommand{\EVV}[2]{\EV_{#1}\!\left[{#2}\right]}

\newcommand{\realspace}{\mathbb R}
\newcommand{\naturalspace}{\mathbb N}

\providecommand{\joinstatespace}{\mathcal S}
\providecommand{\joinactionspace}{\mathcal A}

\providecommand{\rewardmodel}{\mathcal{R}}
\providecommand{\probmodel}{\mathcal{P}}

\providecommand{\rewardundefined}{\bot}

\providecommand{\statespace}{\mathcal S}
\providecommand{\actionspace}{\mathcal A}
\providecommand{\envstatespace}{\mathcal S^\textsc{e}}
\providecommand{\envactionspace}{\mathcal A^\textsc{e}}
\providecommand{\envrewardmodel}{\mathcal{R}^\textsc{e}}
\providecommand{\envprobmodel}{\mathcal{P}^\textsc{e}}

\providecommand{\monstatespace}{\mathcal S^\textsc{m}}
\providecommand{\monactionspace}{\mathcal A^\textsc{m}}
\providecommand{\monrewardmodel}{\mathcal{R}^\textsc{m}}
\providecommand{\monprobmodel}{\mathcal{P}^\textsc{m}}
\providecommand{\monitormodel}{\mathcal{M}}

\providecommand{\trajspace}{\mathbb{T}}
\providecommand{\monspace}{\mathbb{M}}
\providecommand{\indistingspace}{\statespace\actionspace\probmodel\monrewardmodel}
\providecommand{\indisting}{\mathbb{I}}

\providecommand{\rmon}{r^\textsc{m}}
\providecommand{\amon}{a^\textsc{m}}

\providecommand{\smon}{s^\textsc{m}}

\providecommand{\qenv}{Q^\textsc{e}}
\providecommand{\qmon}{Q^\textsc{m}}
\providecommand{\pienv}{\pi^\textsc{e}}
\providecommand{\pimon}{\pi^\textsc{m}}

\providecommand{\qenvpi}{Q^{\pi^{{\textsc{e}}}}\!}
\providecommand{\qenvpiseq}{Q^{\textcolor{sequential}{\pi^{{\textsc{e}}}}}\!}

\providecommand{\qmonenvpi}{Q^{\pi^{{\textsc{m}}}\pi^{{\textsc{e}}}}\!}
\providecommand{\qmonenvpiseq}{Q^{\pi^{{\textsc{m}}}\textcolor{sequential}{\pi^{{\textsc{e}}}}}\!}
\providecommand{\qmonenvpistar}{Q^{\pi^{{\textsc{m}}}\pi^{{\textsc{e}*}}}\!}
\providecommand{\qmonenvpistarseq}{Q^{\pi^{{\textsc{m}}}\textcolor{sequential}{\pi^{{\textsc{e}*}}}}\!}
\providecommand{\qenvstar}{Q^{\pi^{{\textsc{e}*}}}\!}
\providecommand{\qenvstarseq}{Q^{\textcolor{sequential}{\pi^{{\textsc{e}*}}}}\!}

\providecommand{\pienvstar}{\pi^{{\textsc{e}*}}}
\providecommand{\pimonstar}{\pi^{{\textsc{m}*}}}

\providecommand{\rprox}{\hat{r}^\textsc{e}}

\providecommand{\renv}{r^\textsc{e}}
\providecommand{\aenv}{a^\textsc{e}}
\providecommand{\senv}{s^\textsc{e}}

\providecommand{\horizon}{\infty}

\newtheorem{Definition}{Definition}
\newtheorem{Lemma}{Lemma}
\newtheorem{Proposition}{Proposition}

\newtheorem{Proof}{Proof}
\newtheorem{Property}{Property}

\definecolor{oracle}{rgb}{0.12156862745098039, 0.4666666666666667, 0.7058823529411765}
\definecolor{model}{rgb}{1.0, 0.4980392156862745, 0.054901960784313725}
\definecolor{sequential}{rgb}{0.17254901960784313, 0.6274509803921569, 0.17254901960784313}
\definecolor{joint}{rgb}{0.8392156862745098, 0.15294117647058825, 0.1568627450980392}
\definecolor{ignore}{rgb}{0.5803921568627451, 0.403921568627451, 0.7411764705882353}
\definecolor{zero}{rgb}{0.5490196078431373, 0.33725490196078434, 0.29411764705882354}

\acmSubmissionID{507}

\title[Monitored Markov Decision Processes]{Monitored Markov Decision Processes}

\author{Simone Parisi}
\affiliation{
  \institution{University of Alberta}
  \city{Edmonton}
  \country{Canada}}
\email{parisi@ualberta.ca}

\author{Montaser Mohammedalamen}
\affiliation{
  \institution{University of Alberta}
  \city{Edmonton}
  \country{Canada}}
\email{mohmmeda@ualberta.ca}

\author{Alireza Kazemipour}
\affiliation{
  \institution{University of Alberta}
  \city{Edmonton}
  \country{Canada}}
\email{kazemipo@ualberta.ca}

\author{Matthew E. Taylor}
\affiliation{
  \institution{University of Alberta \\ Alberta Machine Intelligence Institute}
  \city{Edmonton}
  \country{Canada}}
\email{matthew.e.taylor@ualberta.ca}

\author{Michael Bowling}
\affiliation{
  \institution{University of Alberta \\ Alberta Machine Intelligence Institute}
  \city{Edmonton}
  \country{Canada}}
\email{mbowling@ualberta.ca}

\begin{abstract}
In reinforcement learning (RL), an agent learns to perform a task by interacting with an environment and receiving feedback (a numerical reward) for its actions. 
However, the assumption that rewards are always observable is often not applicable in real-world problems. For example, the agent may need to ask a human to supervise its actions or activate a monitoring system to receive feedback. 
There may even be a period of time before rewards become observable, or a period of time after which rewards are no longer given. In other words, there are cases where the environment generates rewards in response to the agent's actions but the agent cannot observe them. In this paper, we formalize a novel but general RL framework --- \emph{Monitored MDPs} --- where the agent cannot always observe rewards. We discuss the theoretical and practical consequences of this setting, show challenges raised even in toy environments, and propose algorithms to begin to tackle this novel setting. This paper introduces a powerful new formalism that encompasses both new and existing problems
and lays the foundation for future research.
\end{abstract}


\begin{document}

\pagestyle{fancy}
\fancyhead{}

\maketitle

\section{Introduction}
\label{sec:intro}

Reinforcement learning (RL) has developed into a powerful setting where agents can tackle a variety of tasks, including games~\citep{schrittwieser2020mastering}, robotics~\citep{kober2013reinforcement}, medical applications~\citep{yu2021reinforcement}, and user engagement~\citep{gauci2018horizon}.
Autonomous agents trained with RL learn by trial and error: they are deployed in an environment, try different actions, and receive a numerical reward depending on the outcome of their actions. More interactions lead to more data as the agent tries to maximize its rewards.
Traditionally, RL frames the environment-agent interaction as a Markov Decision Process (MDP), where rewards are assumed to be observable after every action. 
This is in stark contrast with many real-world situations, where the agent may need additional instrumentation (e.g., cameras or specialized sensors) or a human expert to observe the reward~\citep{zanzotto2019human}. 
If the instrumentation breaks or the expert is unavailable, the agent does not observe any reward for its actions, even though the efficacy of its behavior is still important.
Or even if the agent observes rewards, these could be imperfect due to human mistakes or faulty instrumentation~\citep{li2023mind}. 
\textit{In other words, there will be situations where the agent cannot observe the exact rewards generated by the environment to judge its actions.}
This paper argues that such circumstances should be part of the problem specification, suggesting an extension to MDPs is needed, as agents that ignore these complications can result in real-world failures.

\begin{figure*}[t]
    \centering
    \begin{subfigure}[t]{0.32\textwidth}
        \centering
        \includegraphics[width=0.87\linewidth]{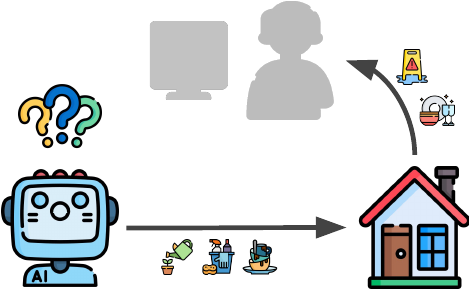}
        \end{subfigure}
    \hfill
    \begin{subfigure}[t]{0.32\textwidth}
        \centering
        \includegraphics[width=0.87\linewidth]{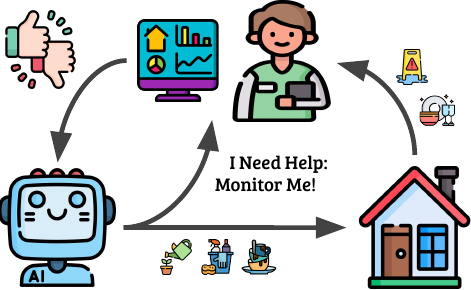}
    \end{subfigure}
    \hfill
    \begin{subfigure}[t]{0.32\textwidth}
        \centering
        \includegraphics[width=0.87\linewidth]{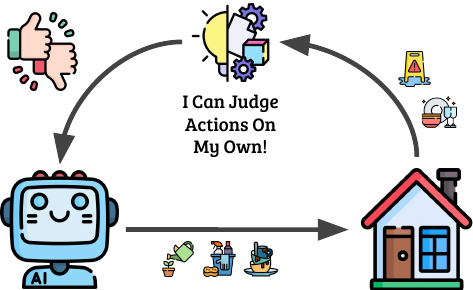}
    \end{subfigure}
    \caption{\label{fig:intro}
    \textbf{Example of Monitored MDP. The agent is tasked with household chores but needs the owner or home sensors to observe rewards.}  
    {If the owner is not home or the sensors are unavailable \textit{(left)}, the agent will not receive positive rewards for cleaning dishes or negative rewards for spilling water.} 
    {Thus, the agent must learn how to seek monitoring --- where to move for sensory feedback or when the owner is home \textit{(center)} --- and act appropriately even when monitoring is unavailable (e.g., be cautious when not being monitored).} 
    {Eventually, the agent can judge actions on its own without any monitoring \textit{(right)}.}}
\end{figure*}

Consider the situation shown in Figure~\ref{fig:intro}. Here, an autonomous agent is tasked with household chores, and the quality of its behavior is provided through feedback from the homeowner and smart sensors.
However, this reward feedback is not always observable, as the owner may not be present, the sensors may not have full coverage of the home, or even be malfunctioning. 
In such situations, the agent should not interpret the lack of reward as meaning that all behavior is equally desirable. Neither should it think that avoiding monitoring or intentionally damaging sensors is an effective way to avoid negative feedback. 
Further, the agent may need to reason about how to 
seek the most useful feedback, such as planning exploratory actions when the owner is home or in well-monitored rooms. 
Ideally, such an agent will eventually learn to judge its own actions without the need for human feedback or home sensors.

We argue that to autonomously learn to complete tasks in such real-world settings, RL agents need a comprehensive framework where 1) the agent cannot always observe rewards even though its behaviour should still seek to maximize the unobserved reward; 2) the agent may need to explicitly act to observe rewards, yet whether it observes the reward or not may not be fully under its control; 3) the process that determines the observation of rewards, namely \textit{the monitor}, is itself something that can be learned or modelled by the agent; and 4) the reward provided to the agent may be imperfect. 
Most importantly, even if the agent does not receive explicit rewards through the monitor, its actions are still impactful: \textit{the environment always generates rewards in response to the agent's actions, even when the agent cannot observe them because the monitor --- that communicates rewards to the agent --- could be unavailable or faulty.}

To the best of our knowledge, there is no framework that fully formalizes this problem setting in the RL literature. \emph{Active RL} tackles similar problems~\citep{schulze2018active, krueger2020active}, but it is limited to cases where there are explicit binary actions that deterministically control when reward is observed. 
\emph{Partial monitoring} addresses the problem of learning with limited feedback, but only in bandits~\citep{littlestone1994weighted, bartok2014partial, auer2002nonstochastic, lattimore2019information}. 
In \textit{sparse-reward RL}, the agent receives a zero-reward after each action until eventually receiving a more informative reward~\citep{ladosz2022exploration}, but rewards are always observable. 
In \textit{cautious RL}, rewards may not be observable~\citep{mohammedalamen2021learning} but the agent has no means to control their observability. 

This paper formalizes \textit{Monitored Markov Decision Processes (Mon-MDPs)}, a novel RL framework that accounts for unobservable rewards by introducing the \textit{monitor}, a separate MDP that dictates when and how the agent sees the rewards, and that the agent can affect with dedicated actions. 
We discuss the theoretical and practical consequences of unobservable rewards, present toy environments, and provide algorithms to illustrate the resulting challenges.
We believe that Mon-MDPs allow to formalize the complexity of many real-world tasks, provide a unifying view of existing areas of research, and lay the foundation for new research directions.

\section{Problem Formulation}
\label{sec:preliminaries}

A Markov Decision Process (MDP) is a mathematical framework for sequential decision-making, defined by the tuple $\langle \joinstatespace, \joinactionspace, \rewardmodel, \probmodel, \upgamma\rangle$. An agent interacts with an environment by repeatedly observing a state $s_t \in \joinstatespace$, taking an action $a_t \in \joinactionspace$, and observing a bounded reward $r_t \in \realspace$. 
The state dynamics are governed by the Markovian transition function $\probmodel(s_{t+1} \mid a_t,s_{t})$, while the reward is determined by the reward function $r_t \sim \rewardmodel(s_t,a_t)$. Both functions are unknown to the agent, whose goal is to act to maximize the sum of discounted rewards ${\scriptstyle\sum}_{t=1}^\horizon \upgamma^{t-1} r_t$, 
where $\upgamma \in [0,1)$ is the discount factor that describes the trade-off between immediate and future rewards.\footnote{The constraint $\upgamma < 1$ ensures the infinite sum is well-defined. Alternatively, one could allow $\upgamma = 1$ but with restrictions on the MDP, e.g., absorbing states.}

\subsection{Monitored MDPs}
\label{subsec:mon_mdps}
\begin{figure}[h!]
    \centering
    \includegraphics[width=0.95\columnwidth]{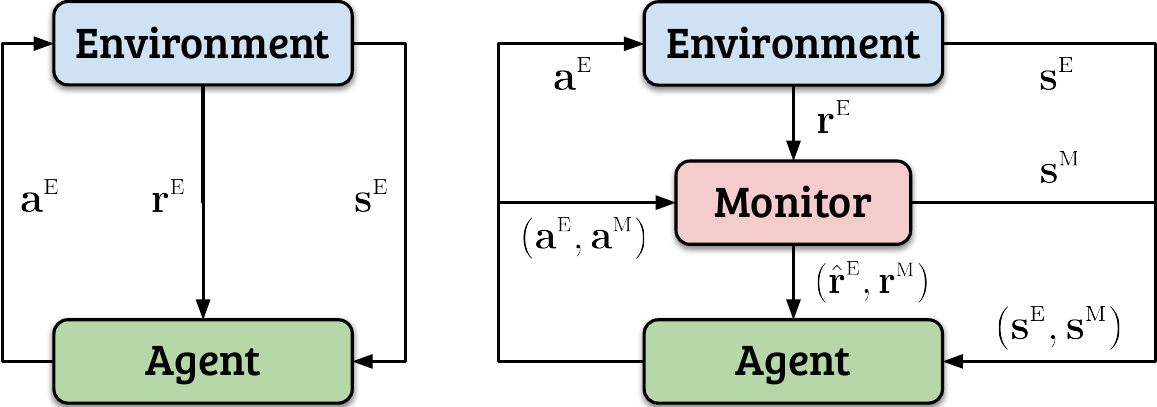}
    \caption{\label{fig:mdp+mmdp}In classic MDPs (left), the agent directly observes environment rewards $\renv$. In Mon-MDPs (right), the agent cannot. 
    {Instead, it receives proxy rewards $\rprox$ via the \textit{monitor}. Like the environment, the monitor is governed by a Markovian transition function and has its own rewards $\rmon$. 
    At every step, the agent observes the state of both the environment and the monitor, and executes actions affecting both. 
    The agent's goal is to maximize the cumulative sum of rewards $(\renv + \rmon)$, while observing $\rprox$ instead of $\renv$. If the reward is unobservable, the agent receives $\rprox = \rewardundefined$.}
    }
\end{figure}

In {Monitored MDPs (Mon-MDPs)} the observability of the reward is governed by the \textit{monitor}, a separate Markovian decision process. 
Formally, a Mon-MDP is defined by the tuple
\begin{equation*}
\langle 
\envstatespace, 
\envactionspace, 
\envprobmodel, 
\envrewardmodel, 
\monitormodel,
\monstatespace,
\monactionspace,
\monprobmodel,
\monrewardmodel,
\upgamma
\rangle.
\end{equation*}
The tuple $\langle \envstatespace, \envactionspace, \envprobmodel, \envrewardmodel, \upgamma \rangle$ is the same as classic MDPs where the superscript $\textsc{e}$ stands for ``environment.'' 
However, the \textit{environment reward} $\renv_t \sim \envrewardmodel(\senv_t, \aenv_t)$ is not directly observable. Instead, the agent observes a \textit{proxy reward} $\rprox_t \sim \monitormodel(\renv_t, \smon_{t}, \amon_t)$, where $\monitormodel$ is the \textit{monitor function}, $\smon \in \monstatespace$ is the monitor state, and $\amon \in \monactionspace$ is the monitor action.
Even so, the monitor function is not guaranteed to always show a reward and the agent may receive $\rprox_t = \rewardundefined$, i.e., ``\textit{unobservable reward}'' (i.e., the monitor dictates what the agent sees about the environment reward according to its current state and action). The monitor state follows the Markovian transition function $\monprobmodel(\smon_{t+1} \mid \smon_t, \senv_t, \amon_t, \aenv_t)$,\footnote{The monitor states and actions may be as simple as ``reward is (not) observable'' and ``do (not) ask for reward'', or more heterogeneous. For example, the agent could ask for rewards by pushing buttons, collecting and using a device, or uncovering objects. Similarly, the monitor state may include the position of items, the battery of a device, or the distance of the agent from environment sensors. The next monitor state $\smon_{t+1}$, indeed, depends on both the current monitor pair $(\smon_t, \amon_t)$ and environment pair $(\senv_t, \aenv_t)$. 
Consequently, the proxy reward also depends on both the environment and the monitor. In Figure~\ref{fig:intro}, to receive rewards from the home sensors, the sensor has to be active (monitor state) and the agent near enough (environment state).
}
and executing monitor actions yields a bounded \textit{monitor reward} $\rmon_t \sim \monrewardmodel(\smon_t, \amon_t)$.\footnote{The monitor reward can represent a cost (e.g., if monitoring consumes resources), but it is not constrained to be negative. Just like the environment reward, its design depends on the agent's desired behavior.}
We will use \textit{monitor} to refer to the tuple $\langle \monitormodel, \monstatespace, \monactionspace, \monprobmodel, \monrewardmodel \rangle$, and \textit{monitor function} to refer to $\monitormodel$. The monitor together with the environment tuple $\langle \envstatespace, \envactionspace, \envprobmodel, \envrewardmodel, \upgamma \rangle$ is the \textit{Mon-MDP}.
Figure~\ref{fig:mdp+mmdp} shows a diagram of the Mon-MDP in contrast to the standard MDP framework.

\clearpage

In Mon-MDPs, the agent repeatedly executes a joint action $(\aenv_t, \amon_t)$ according to the joint state $(\senv_t, \smon_t)$. In turn, the environment and monitor states change and produce a joint reward $(\renv_t, \rmon_t)$, but the agent observes $(\rprox_t, \rmon_t)$. 
The agent's goal is to select joint actions to maximize ${\scriptstyle\sum}_{t=1}^\horizon \upgamma^{t-1} \left(\renv_t + \rmon_t\right)$ \emph{even though it only observes} $\rprox_t$ instead of $\renv_t$. 
As shown in the remainder of the paper, not observing directly $\renv_t$ (and possibly not observing any reward if $\rprox_t = \rewardundefined$) makes the optimization non-trivial, creating challenging --- \textit{and potentially impossible} --- problems, which we discuss further in Section~\ref{sec:mmdp}.

\subsection{\textbf{Why RL Needs Mon-MDPs}}
\label{subsec:remarks}

As in MDPs, the agent is always being judged, i.e., the environment always generates rewards in response to the agent's actions. In Mon-MDPs, however, the agent does not observe these rewards directly, but instead observes the proxy rewards given by the monitor. Most importantly, the monitor does not affect the environment reward --- how the agent is judged --- but only what the agent observes. 
In Figure~\ref{fig:intro}, if the agent spills water on the floor, there is a clear ``bad'' feedback that the owner (i.e., the monitor) would give to the agent. However, if the owner is not home, the agent would not receive it. 
Nonetheless, the action that spilled water is still undesirable.

An alternative to the Mon-MDP framing of such a situation may be to formalize it as an MDP with delayed reward: the act of spilling water when the owner is not home causes no immediate feedback, but later when the owner returns there is negative feedback for the floor being wet. 
Such a framing either dictates a non-Markovian reward function or places a significant burden on the state representation to capture long sequences of actions, as well as stressing the agent's ability to do credit assignment. 
Further, suppose that the water dries before the owner returns, or that the agent purposefully only spills water where the owner will not notice. The agent may never receive any 
negative feedback for this behavior --- a behavior that is undesirable.
\textit{The owner's lack of feedback should not be interpreted by the agent as indifference} to a behavior that would otherwise result in negative feedback when monitored. Instead, \textit{the agent should understand that spilling water (an environment action) is undesirable regardless of the owner's presence (the monitor state).} 

Another alternative to the Mon-MDP framing may be to avoid explicitly representing the monitor with its own states, actions, 
and rewards, as partially decoupled from the environment. 
However, this separation --- where the monitor and its state {do not affect the environment rewards} --- is the basis for the agent to know that hiding spilled water from the owner does not avoid (unobserved) negative rewards. 
An alternative that forces the environment and monitor state together into a single state variable (and similarly for environment and monitor actions) would only be able to achieve the same effect through carefully constructed generalization bias in the agent's representation, which is its own challenges. 

There are further advantages for separating out the monitor process.
First, whether the details of the monitor are known in advance by the agent is a design decision, and keeping them separate allows clarity to what is known and what must be learned through interaction.
Second, explicitly reasoning about the monitor and environment separately facilitates better exploration and more advanced behavior. For example, if the outcome of some states or actions has not been monitored, the agent could either (a) avoid them, or (b) intentionally explore them when it knows it will gain monitoring feedback. For instance, if the agent learns that it can reliably observe rewards when the owner is home, it could try new actions for which it does not know the reward. In contrast, if the owner is not home, the agent may choose to act cautiously. 
Third, decoupling the monitor and the environment creates new opportunities for task transfer, where only the monitor is different or only the environment is.  
For example, the agent may first learn in a Mon-MDP and then be deployed in a similar --- but entirely unmonitored --- environment, or may be assigned new tasks under the same monitor. E.g., the agent in Figure~\ref{fig:intro} could be assigned new chores within the same house, and reasoning about the presence of the owner or home sensors would allow to learn more efficiently. 

We argue that current RL frameworks do not capture the complexity of these --- and many more --- real-world tasks. In contrast, Mon-MDPs provide a comprehensive framework that can be applied to a large variety of challenging real-world complexities. 
In the next section, we discuss related work that capture some of these complexities, and show how Mon-MDPs can be seen as a general framework encompassing existing areas of research. 

\begin{figure*}[t]
\begin{minipage}[c]{0.35\textwidth}
    \centering
    \includegraphics[width=0.95\linewidth]{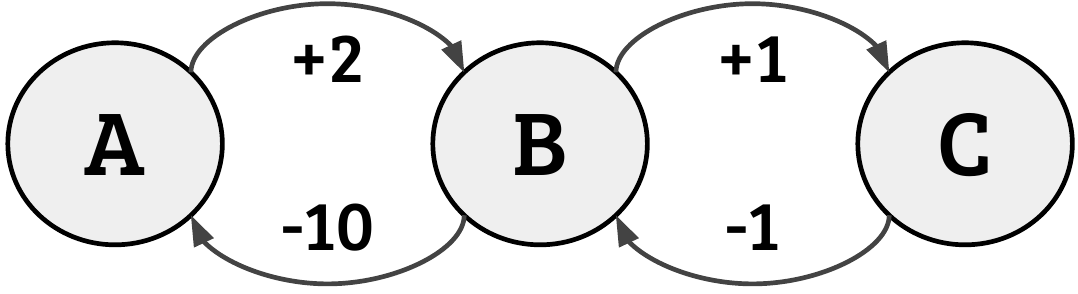}
\end{minipage}
\hfill
\begin{minipage}[c]{0.63\textwidth}
    \vspace*{-5pt}
    \caption{\label{fig:simple_monmdp}
    \textbf{Example showing why learning with unobservable reward is non-trivial.}
    {The agent moves between states \texttt{A}, \texttt{B}, and \texttt{C}, but rewards can be observed only with monitoring. Otherwise, the agent observes $\rprox_t = \rewardundefined$. If the agent interprets the lack of reward as meaning that all actions are equally good, it could believe that transitioning from \texttt{B} to \texttt{A} is as good as transitioning from \texttt{B} to \texttt{C}.}
    }
\end{minipage}
\end{figure*}

\subsection{Related Work}
\label{subsec:related}

In \textbf{partially observable MDPs (POMDPs)}, the environment state is not directly observable. The presence of an unobservable component may induce the reader to think that Mon-MDPs are related to POMDPs~\citep{astrom1965optimal}. However, not observing the reward rather than the state is not just cosmetics. 
In Mon-MDPs, the agent cannot judge its own actions without rewards, but can still identify the environment state (and, thus, explore). 
In contrast, in POMDPs the agent must learn to identify the state from its observations to provide an appropriate input for its actions~\citep{golowich2022learning}. 
POMDPs and Mon-MDPs are not mutually exclusive, though, and we can formalize Mon-POMDPs where both rewards and states are not observable.

In \textbf{sparse-reward RL}, the agent receives meaningful rewards rarely and zero-rewards otherwise. This makes exploration hard, especially if the reward is given only at task completion. 
To compensate for the sparsity of rewards, {intrinsic motivation} relies on auxiliary rewards such as bonuses for hard-to-predict states~\citep{schmidhuber1991possibility, stadie2015incentivizing, pathak2017curiosity}, rarely-visited states~\citep{strehl2008analysis, bellemare2016unifying, parisi2022long}, or impactful actions~\citep{raileanu2020ride, parisi2021interesting}. At first it may seem that intrinsic motivation could be a complete solution to Mon-MDPs --- the agent could use auxiliary rewards when environment rewards are unobservable. 
However, in Mon-MDPs the problem is not the sparsity of rewards, but their unobservability, i.e., receiving $\rprox_t = \rewardundefined$. 
While auxiliary rewards may improve exploration, they cannot replace $\renv_t$ --- that remains unobservable --- and thus the agent cannot directly maximize the sum of environment rewards.  Nonetheless, techniques for sparse-reward MDPs will likely still be valuable 
in Mon-MDPs. We return to this in Section~\ref{sec:future}.

In \textbf{cautious and risk-averse RL}, the agent faces some form of uncertainty, either aleatoric (inherent randomness of the environment) or epistemic (due to the agent's ignorance). 
In Mon-MDPs, the unobservability of the reward can be seen as epistemic uncertainty. Thus, cautious and risk-averse methods could be used to reason about rewards uncertainty. 
However, the general setting of cautious and risk-averse RL is fundamentally different from Mon-MDPs, as either the reward is always observable~\citep{zintgraf2019varibad, zhang2020cautious} or never~\citep{mohammedalamen2021learning}.

In \textbf{human-in-the-Loop (HITL)}, a human helps the agent in making the correct decisions, e.g., by providing rewards or suggesting appropriate actions~\citep{knox2009interactively, christiano2017deep, macglashan2017interactive}. 
One may see the monitor as the human in HILT.
However, in HITL, typically there is either (a) \emph{never} an environment reward, e.g., 
the MDP has no reward function 
and all guidance comes from the human; or (b) \emph{always} an environment reward, and the human provides additional guidance to the agent. In neither of these settings there is an environment reward that the agent can only sometimes see.  Yet, many of these techniques deal with imperfect (human) rewards, which are a factor in Mon-MDPs.

In \textbf{Bayesian persuasion}, rewards depend on states, actions, and an external parameter~\citep{kamenica2011bayesian, bernasconi2022sequential, gan2022bayesian, wu2022sequential}. 
One agent (the \textit{sender}) cannot influence the environment state, but its actions determine what another agent (the \textit{receiver}) observes about the external parameter. 
This relationship recalls the monitor-agent's in Mon-MDPs, in the sense that one affects what the other observes. However, the two frameworks are fundamentally different. 
In Bayesian persuasion, there are two decision-makers (receiver and sender) with possibly conflicting goals --- the sender affects the receiver's observations for its own good. In Mon-MDPs, the monitor is a fixed process like the environment, and the agent (the only decision-maker) has one goal --- to maximize the sum of monitor and environment rewards. 

In \textbf{active RL (ARL)}, the agent must pay a cost to observe either the state~\citep{bellinger2021active} or the reward~\citep{schulze2018active, krueger2020active, tucker2023bandits}. ARL is perhaps the closest framework to Mon-MDPs but its setting is simpler. To the best of our knowledge, ARL considers only binary actions to request rewards, constant request costs, and perfect reward observations. By contrast, in Mon-MDPs (a) the observed reward depends on the monitor --- a process with its own states, actions, and dynamics; (b) there may be no direct action to request rewards, and requests may fail; (c) the monitor reward is not necessarily a cost; and (d) the monitor can be imperfect and modify the reward. 
For these reasons, Mon-MDPs can be considered a more general form of ARL. 

In \textbf{partial monitoring} for multi-armed bandits, the agent must maximize the payoffs of its actions, while unable to observe the exact payoffs~\citep{littlestone1994weighted, bartok2014partial, auer2002nonstochastic, lattimore2019information}.
E.g., the agent may observe only payoffs for some bandit arms but not all, or payoffs within a range.
Mon-MDPs can be considered a partial monitoring problem, as the agent has to maximize the cumulative sum of partially observable rewards. 
However, to the best of our knowledge, Mon-MDPs are the first example of partial monitoring in sequential decision-making.

\section{Mon-MDPs Optimality}
\label{sec:mmdp}
In Mon-MPDs, the agent's goal is to maximize the sum of cumulative rewards $(\renv_t + \rmon_t)$ while receiving $(\rprox_t, \rmon_t)$. 
However, issues arise when $\rprox_t$ is \textit{unobservable} ($\rprox_t = \rewardundefined$). 
On one hand, we cannot simply replace $\renv_t$ with $\rprox_t$ as the sum would be ill-defined. 
On the other hand, we cannot replace $\rprox_t = \rewardundefined$ with an arbitrary value or even just ignore it (this could result in suboptimal or even dangerous agent's behavior, as we show in Section~\ref{subsec:algos}). 
To address this problem, we will first define optimality in Mon-MDPs, and then we discuss under what conditions convergence to optimality is guaranteed (despite the presence of unobservable rewards).

\subsection{Policy Optimality}
\label{subsec:mmdp_optimality}

We define $a_t \!\coloneqq\! (\aenv_t, \amon_t)$, $s_t \!\coloneqq\! (\senv_t, \smon_t)$, and $r_t \!\coloneqq\! \renv_t + \rmon_t$ as the joint action, joint state, and joint reward, respectively. 
Although $r_t$ may not be observable to the agent, it is well-defined --- we can formalize the problem like a classic MDP with policy $\pi(a_t | s_t)$ and sum of discounted rewards ${\scriptstyle\sum}_{t=1}^\horizon \upgamma^{t-1} r_t$. 
Similarly, we can define an optimal policy $\pi^*$ as a policy maximizing the Q-function $Q^{\pi}(s,a)$, i.e.,
\begin{align}
\pi^* & \coloneqq \arg\max_\pi \: Q^{\pi}(s,a), \label{eq:max_pi}
\\
\textrm{where} \:\:
Q^{\pi}(s_t,a_t) & \coloneqq \mathbb{E}\Bigl[\sum\nolimits_{i=t}^\horizon \upgamma^{i-t} r_{i} \;\Big|\; \pi, \probmodel, s_t, a_t\Bigr],  \label{eq:q}
\end{align}
where $\probmodel(s_{t+1} \mid s_t, a_t) = \monprobmodel(\smon_{t+1} \mid \smon_t, \amon_t, \senv_t, \aenv_t) \, \probmodel(\senv_{t+1} \mid \senv_t, \aenv_t)$. 
This problem is well-defined and the existence of at least one optimal policy is guaranteed under assumptions already satisfied by Mon-MDPs.\footnote{These standard assumptions include stationary reward and transition functions, bounded rewards, and a discount factor $\upgamma \in [0, 1)$~\citep{puterman1994markov}.}
Notice that the monitor function $\monitormodel$ does not appear here --- for every Mon-MDP, there \emph{always} exists an optimal policy. 

\begin{figure*}[htp]
\begin{minipage}[c]{0.71\textwidth}
    \centering
    \begin{subfigure}[t]{0.31\textwidth}
        \centering
        \includegraphics[width=0.9\linewidth]{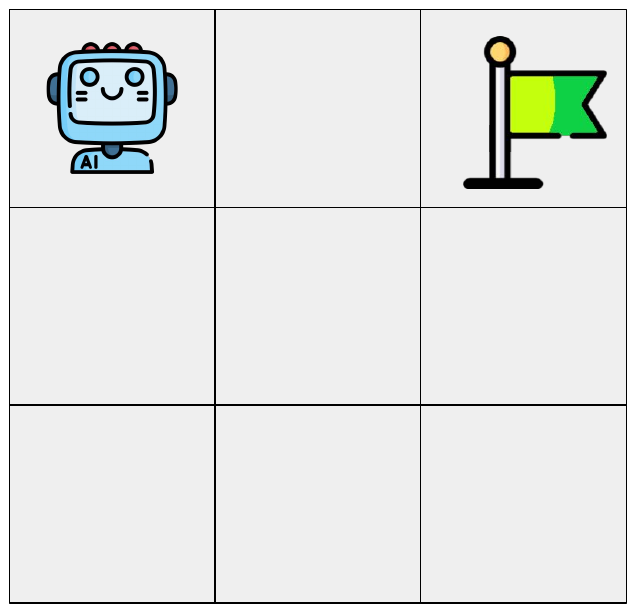}
        \caption{\label{fig:grid_easy}\textit{Simple.} {
        The reward $\renv_t$ is 1 (goal) or 0 (otherwise).
        To observe $\rprox_t = \renv_t$, the agent must do $\amon_t = \texttt{MONITOR ME}$ and pay $\rmon_t = -0.2$. Otherwise $\rprox_t = \rewardundefined$.
        }}
        \end{subfigure}
    \hfill
    \begin{subfigure}[t]{0.31\textwidth}
        \centering
        \includegraphics[width=0.9\linewidth]{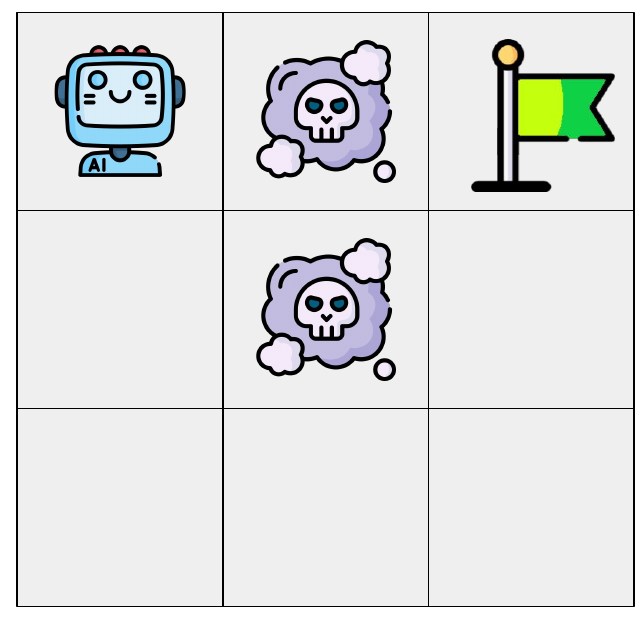}
        \caption{\label{fig:grid_medium}\textit{Penalty.} {
        Like the Simple version, but there are penalty cells that give $\renv_t = -10$.}}
    \end{subfigure}
    \hfill
    \begin{subfigure}[t]{0.31\textwidth}
        \centering
        \includegraphics[width=0.9\linewidth]{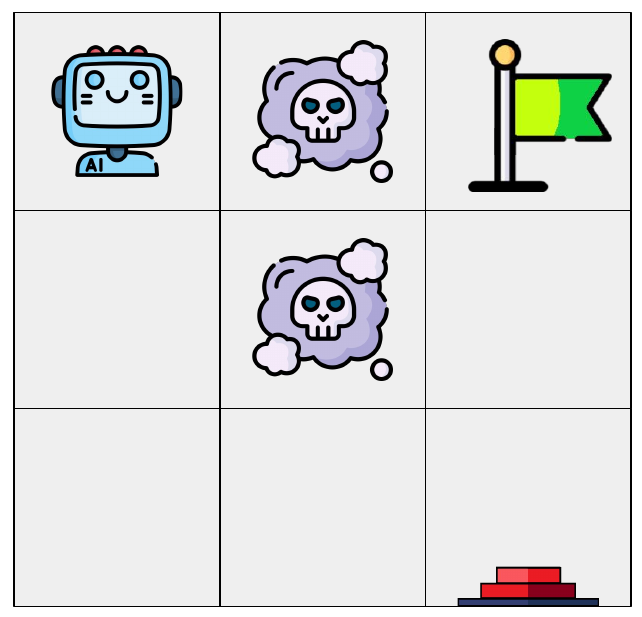}
        \caption{\label{fig:grid_hard}\textit{Button.} {
        Monitoring is turned $\texttt{ON}$ by hitting the red button with $\aenv_t = \texttt{DOWN}$. 
        When monitoring is $\texttt{ON}$, $\rmon_t = -0.2$ and $\rprox_t = \renv_t$ until the button is hit again.}
        }
    \end{subfigure}
\end{minipage}
\hfill
\begin{minipage}[c]{0.265\textwidth}
    \vspace*{-6pt}
    \caption{\label{fig:grid_env}\textbf{In our Mon-MDPs,} the agent starts in the top-left cell and has to reach the goal avoiding penalty cell. There are nine states $\senv$ (one for each cell) and four actions $\aenv$ ($\texttt{LEFT}, \texttt{RIGHT}, \texttt{UP}, \texttt{DOWN}$). 
    Rewards $\renv$ are 1 (goal), -10 (penalty cells), and 0 (otherwise). 
    Episodes end when the agent reaches the goal or after 50 steps. We propose three levels of difficulty depending on the presence of penalty cells and on the type of monitor. \vspace*{-1pt}
    }
\end{minipage}
\end{figure*}

An optimal policy exists, but can the agent actually learn it? The environment reward $\renv_t$ is not always observable and the agent sees $\rprox_t$ instead. 
How should it treat unobservable rewards $\rprox_t = \rewardundefined$? 
This problem is non-trivial. Consider a Mon-MDP whose deterministic environment is shown in Figure~\ref{fig:simple_monmdp}. Every time the agent moves, it can ask to be monitored or not with $\amon \in \{\texttt{MONITOR ME, NO-OP}\}$.
The agent observes $\rprox_t = \renv_t$ when moving 
if $\amon_t = \texttt{MONITOR ME}$, and $\rprox_t = \rewardundefined$ otherwise. 
The monitor reward is constant: $\rmon_t = 0$.  
The optimal policy for sufficiently large $\upgamma$ moves between \texttt{B} and \texttt{C} for an undiscounted cumulative reward of 0. 
What should the agent think when it does not ask to be monitored and sees $\rprox_t = \rewardundefined$? Should it assume $\rewardundefined = 0$? If so, the agent will believe it can avoid negative rewards by not asking to be monitored, and choose to move between \texttt{A} and \texttt{B} for an (apparent, but incorrect) undiscounted cumulative reward of 2. 
Or, what if the monitor function clips the environment reward to $[-1, 1]$,
a common practice in RL~\citep{mnih2015human, van2016learning}?  
In this case, the agent will conclude that moving between \texttt{A} and \texttt{B} yields the same rewards of moving between \texttt{B} and \texttt{C}. 

This example shows that convergence to an optimal policy depends on the observability of proxy rewards, how the agent treats $\rprox_t = \rewardundefined$, and if the monitor function alters environment rewards. In the extreme case, solving a Mon-MDP may be \emph{hopeless}, such as the case of a monitor function always returning $\rewardundefined$, where no agent could ever learn an optimal policy.
The next section discusses some properties of ``well-behaved'' Mon-MDPs, sufficient to guarantee the existence of an algorithm that converges to an optimal policy.  In Appendix~\ref{app:sec:taxonomy}, we give a formal treatment of \emph{solvable} and \emph{unsolvable} Mon-MDPs as well as interesting settings between these two.

\subsection{Convergence to an Optimal Policy}
\label{subsec:solving_monmdps}

\begin{Property}[Ergodic Mon-MDP]
\label{def:ergodic_mmdp}
A Mon-MDP is ergodic if any joint state $(\senv, \smon)$ can be reached by any other joint state $(\senv, \smon)'$ given infinite exploration. This implies that every state will be visited infinitely often given infinite exploration.\footnote{This is a generalization of the definition of ergodic MDPs~\citep{puterman1994markov}.} 
\end{Property}

\begin{Property}[Ergodic Monitor Function]
\label{def:env_ergodic_mon}
A monitor function $\rprox \sim \monitormodel(\renv, \smon, \amon)$ is ergodic if for all environment pairs $(\senv, \aenv)$ the proxy reward will be observable ($\rprox \neq \rewardundefined$) given infinite exploration.
\end{Property}

\begin{Property}[Truthful Monitor Function]
\label{def:truthful_mon}
A monitor function $\rprox \sim \monitormodel(\renv, \smon, \amon)$ is truthful if $\,\forall t$ either $\rprox_t = \renv_t$ or $\rprox_t = \rewardundefined$.  
\end{Property}

\begin{Proposition}[Sufficient Conditions for Convergence to an Optimal Policy]
\label{prop:convergence}
There exist an algorithm such that for any Mon-MDP with finite states and actions satisfying Properties~\ref{def:ergodic_mmdp},~\ref{def:env_ergodic_mon}, and~\ref{def:truthful_mon}, the algorithm converges to an optimal policy of that Mon-MDP. 
\end{Proposition}

We prove Proposition \ref{prop:convergence} in Section~\ref{subsec:proof}, after examining candidate 
algorithms in Section~\ref{subsec:algos}. 
For now, we remark that Properties~\ref{def:ergodic_mmdp},~\ref{def:env_ergodic_mon}, and~\ref{def:truthful_mon} guarantee that the agent will observe every environment reward infinitely often, even though not for every monitor state and action. 
In other words, the agent can still learn an optimal policy \textit{even if there are situations where it cannot observe the rewards}.

\section{Empirical Analysis of Mon-MDPs}
\label{sec:experiments}
In this section, we show practical challenges an agent faces in Mon-MDPs, and why methods used in MDPs fail to converge to an optimal policy. 
We start by introducing toy environments and monitors: in some the agent can use a monitor action to immediately be monitored, while in others it must execute certain environment actions to activate monitoring.
We then introduce algorithms that account for unobservable rewards $\rprox_t = \rewardundefined$ in different ways. 
Source code is available at \url{https://github.com/AmiiThinks/mon_mdp_aamas24}.

\subsection{The Environment and The Monitors}
\label{subsec:envs}

We study Mon-MDPs in the 3$\times$3 gridworld shown in Figure~\ref{fig:grid_env}, where the agent has to reach the goal ($\renv_t = 1$) using actions $\aenv \in \{\texttt{LEFT, DOWN, RIGHT, UP}\}$ while avoiding penalties ($\renv_t = -10$).
However, rewards are not always observable. 
We consider three Mon-MDPs of increasing difficulty, that differ in the environment reward and the monitor dynamics (more details in Appendix~\ref{app:subsec:mon_details}).
\begin{itemize}[leftmargin=1em, itemsep=1pt, topsep=3pt]
\item \textbf{Simple grid (Simple).} Together with $\aenv$, the agent selects $\amon \in \{\texttt{MONITOR ME, NO-OP}\}$. With $\amon_t = \texttt{MONITOR ME}$, the agent observes $\rprox_t = \renv_t$ and pays a cost ($\rmon_t = -0.2$). Otherwise it receives $\rprox_t = \rewardundefined$ at no cost ($\rmon_t = 0)$. 
The optimal policy brings the agent to the goal ($\renv_t = 1$), never asking to be monitored. 
\item \textbf{Grid with penalties (Penalty).} Same monitor as Simple, but the grid now has cells with negative environment reward ($\renv_t = -10$). The optimal policy brings the agent to the goal, avoiding penalty cells and never asking to be monitored. 
\item \textbf{Grid with penalties and button (Button).} There is no \texttt{MONITOR ME} action. Instead, monitoring is determined by the monitor state $\smon \!\in \!\{\texttt{ON, \!\!\!\!\! OFF}\}$.  The agent can change $\smon$ by hitting the red button in the rightmost state with $\aenv_t = \texttt{DOWN}$. At the start of an episode, the monitor state is set randomly.  Every time step $\smon_t = \texttt{ON}$, the agent pays a cost ($\rmon_t = -0.2$). 
The optimal policy brings the agent to the goal, avoiding penalty cells and turning \texttt{OFF} the monitor along the way if it was \texttt{ON} at the start of the episode.
\end{itemize}
In the Simple and Penalty Mon-MDPs, the agent can observe rewards with the explicit \textit{monitor} action $\amon_t = \texttt{MONITOR ME}$. 
In the Button Mon-MDP, instead, the agent must use a sequence of \textit{environment} actions $\aenv_t$ to start (or stop) observing rewards. 
This, and not being able to observe rewards for a period of time, has important consequences as we show in the next section.

\subsection{The Algorithms}
\label{subsec:algos}

We present algorithms based on Q-Learning~\citep{watkins1992q}. 
Given samples $(s_t, a_t, r_t, s_{t+1})$, Q-Learning updates are
\vspace*{-12pt}
\begin{equation}
    Q(s_t, a_t) \leftarrow (1 - \upalpha_t) Q(s_t, a_t) + \upalpha_t (r_t + \upgamma \overbrace{\max_a Q(s_{t+1}, a)}^{\text{greedy policy}}), \label{eq:qlearning} 
\end{equation}
where $\upalpha_t$ is the learning rate, and we wrote $Q$ in place of $Q^\pi$ for the sake of simplicity. 
In classic MDPs, Q-Learning is guaranteed to converge to an optimal greedy policy using $\varepsilon$-greedy exploration with an appropriate learning rate $\upalpha_t$ and exploration $\varepsilon_t$ schedules~\citep{dayan1992convergence}. 

In Mon-MDPs, we observe $\hat r_t \coloneqq (\rprox_t, \rmon_t)$ --- how do we update $Q(s_t,a_t)$ when $\rprox_t = \rewardundefined$?
We consider a set of Q-Learning variants that differ in how they treat $\rprox_t = \rewardundefined$, and show that the resulting policies can be very different (and often suboptimal).
For each variant, we show the \textit{greedy policy} learned after 10,000 steps in each of our three Mon-MDPs. Figures show the action taken in every cell --- including monitoring actions. Figures outlined in {red} are suboptimal policies, while figures outlined in {green} are optimal.
Over 100 seeds, the algorithms always converged to these policies within 10,000 steps when $\upgamma = 0.99$. For all details about the algorithms variants, please see Appendix~\ref{app:sec:alg_details}.

\begin{figure}[t]
    \begin{subfigure}[b]{\columnwidth}
        \centering
        \includegraphics[trim=0 41pt 0 0, width=0.9\columnwidth]{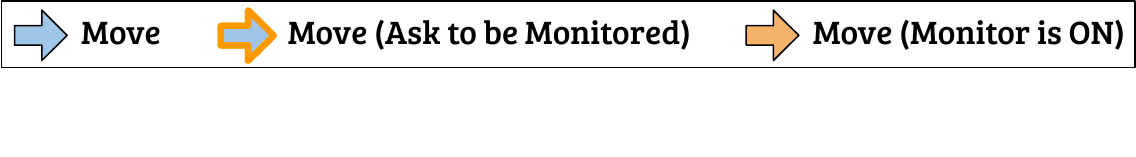}
    \end{subfigure}
    \\
    \begin{subfigure}[b]{.32\columnwidth}
        \centering
        \includegraphics[width=\columnwidth]{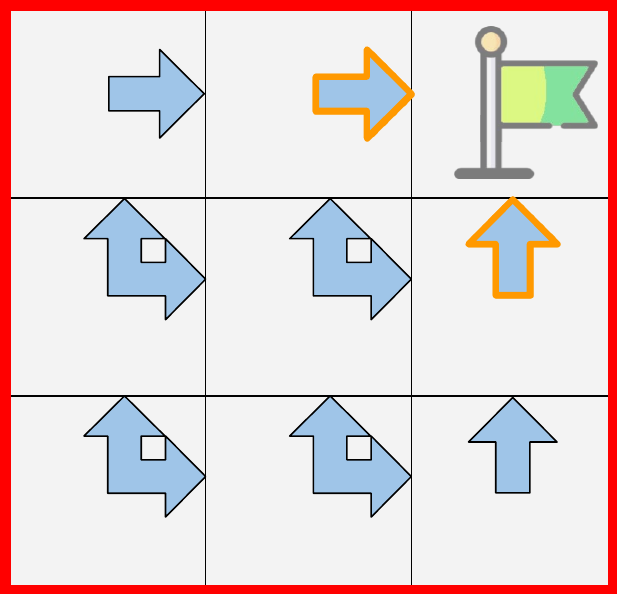}
        \caption{\label{fig:policies_zero_easy}Simple \bad}
    \end{subfigure}
    \hfill
    \begin{subfigure}[b]{.32\columnwidth}
        \centering
        \includegraphics[width=\columnwidth]{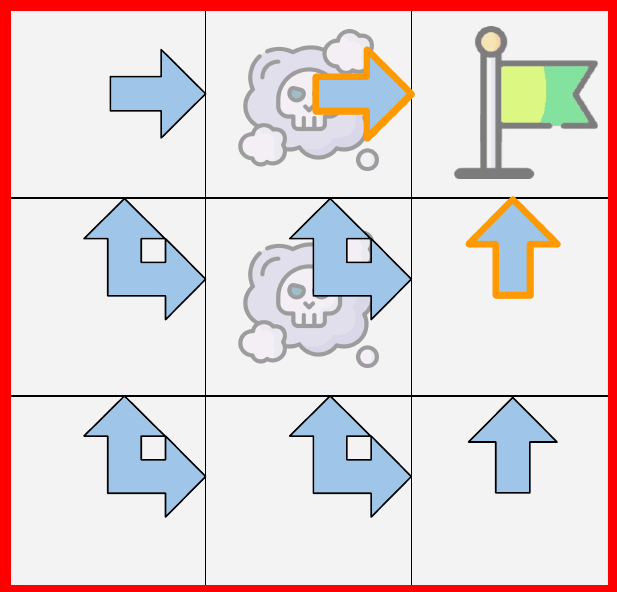}
        \caption{\label{fig:policies_zero_medium}Penalty \bad}
    \end{subfigure}
    \hfill
    \begin{subfigure}[b]{.32\columnwidth}
        \centering
        \includegraphics[width=\columnwidth]{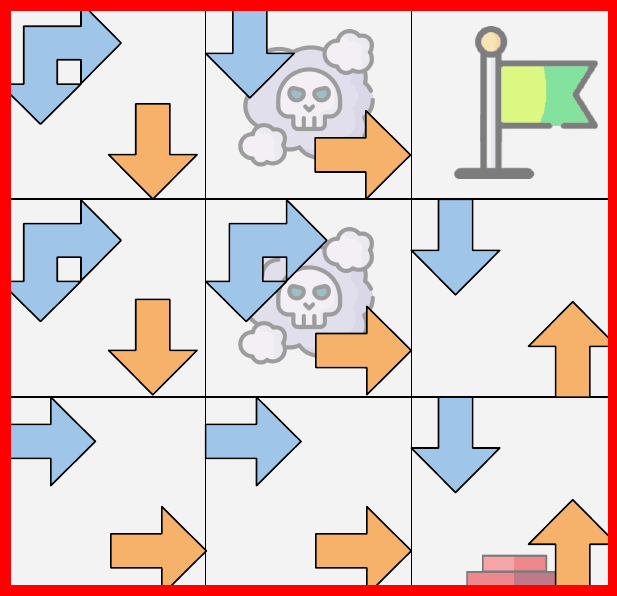}
        \caption{\label{fig:policies_zero_hard}Button \bad}
    \end{subfigure}
    \caption{\label{fig:policies_zero}\textbf{If $\rewardundefined = 0$,} the agent asks to be monitored only when rewards are positive, ``ignoring'' negative rewards.}
\end{figure}
\subsubsection{\textbf{Algorithm 1: Assign $\boldsymbol{\rewardundefined = 0}$.}}
\label{subsubsec:zero}
Our first algorithm assumes that unobservable rewards $\rprox_t = \rewardundefined$ have a constant value of 0. This can be seen as treating the Mon-MDP as a sparse-reward MDP, where most rewards are 0. 
Figure~\ref{fig:policies_zero} shows that the policy ends up ignoring negative rewards, asking to be monitored only when positive rewards can be observed. 
In the Simple Mon-MDP, the agent asks $\amon_t = \texttt{MONITOR ME}$ as it moves to the goal ($\renv_t = 1$). 
In the Penalty Mon-MDP, the agent does not avoid penalty cells ($\renv_t = -10$), ``pretending'' that walking over them gives $\renv_t = 0$ by not asking to be monitored. 
In the Button Mon-MDP, the agent learns to press the button --- again, without avoiding penalty cells --- and then goes to the goal.  All of the learned policies across all three Mon-MDPs are suboptimal.
In Appendix~\ref{app:subsec:ablation_zero}, we show that this algorithm performs poorly for different values assigned to $\rewardundefined$.

\begin{figure}[b]
    \begin{subfigure}[b]{.32\columnwidth}
        \centering
        \includegraphics[width=\columnwidth]{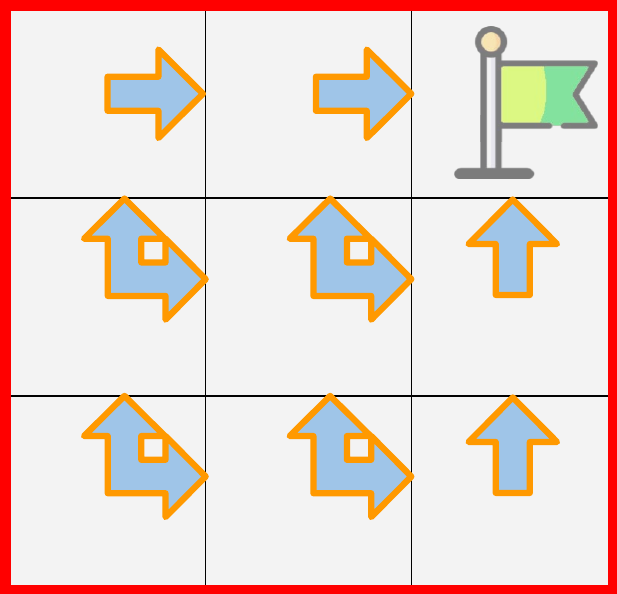}
        \caption{\label{fig:policies_ignore_easy}Simple \bad}
    \end{subfigure}
    \hfill
    \begin{subfigure}[b]{.32\columnwidth}
        \centering
        \includegraphics[width=\columnwidth]{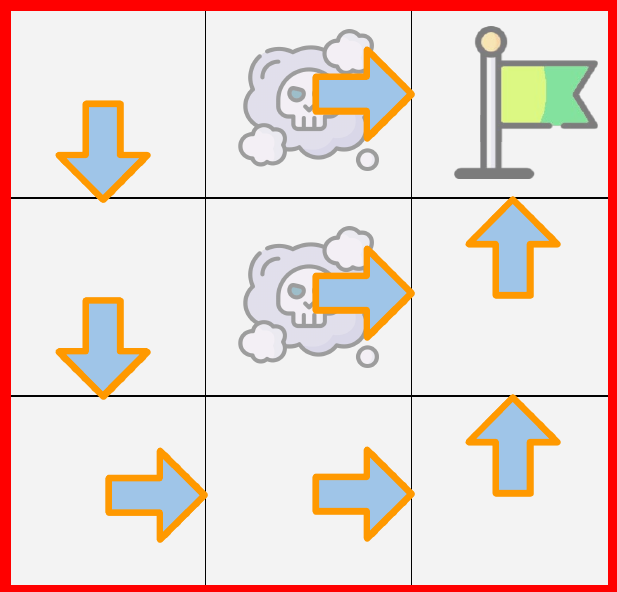}
        \caption{\label{fig:policies_ignore_medium}Penalty \bad}
    \end{subfigure}
    \hfill
    \begin{subfigure}[b]{.32\columnwidth}
        \centering
        \includegraphics[width=\columnwidth]{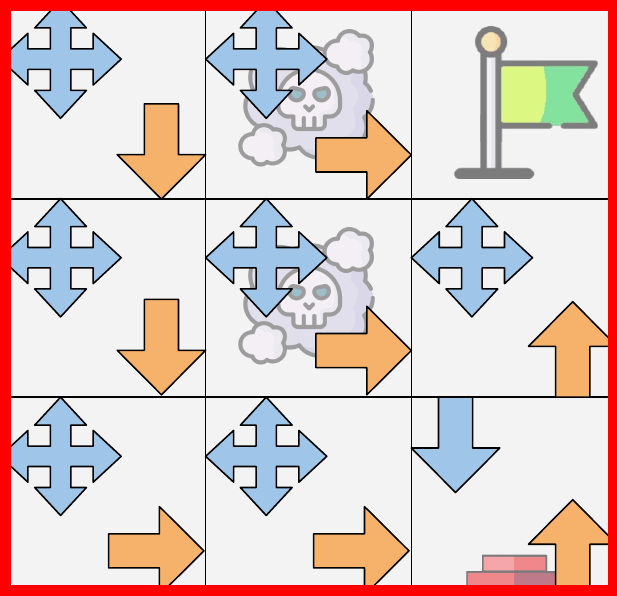}
        \caption{\label{fig:policies_ignore_hard}Button \bad}
    \end{subfigure}
    \caption{\label{fig:policies_ignore}\textbf{Ignoring updates when $\rprox_t = \rewardundefined$}, results in a policy that can navigate only when monitored.}
\end{figure}
\subsubsection{\textbf{Algorithm 2: Ignore $\boldsymbol{\rprox_t = \rewardundefined}$.}}
\label{subsubsec:ignore}
Instead of assigning an arbitrary value to unobservable rewards, the algorithm does not update the Q-function when $\rprox_t = \rewardundefined$. This could be considered a safe strategy, as the agent disregards samples with incomplete information. 
However, as shown in Figure~\ref{fig:policies_ignore}, the resulting policy ends up always seeking monitoring. In the Simple and Penalty Mon-MDPs, the agent executes $\amon_t = \texttt{MONITOR ME}$ in every state. 
In the Button Mon-MDP, when the monitor is \texttt{ON} the agent walks to the goal without pressing the button. However, when the monitor is \texttt{OFF}, the policy acts randomly. 
This happens because the Q-function is never updated when the monitor is \texttt{OFF}, as receiving $\rprox_t = \rewardundefined$ precludes any update.  
As a result, its learned policy depends only on the Q-function's initialization.  
When all Q-values initialized to the same value, the policy is random, as shown in Figure~\ref{fig:policies_ignore_hard}.

\begin{figure}[t]
    \begin{subfigure}[b]{\columnwidth}
        \centering
        \includegraphics[trim=0 41pt 0 0, width=0.9\columnwidth]{fig/policies/legend_v5.pdf}
    \end{subfigure}
    \\
    \begin{subfigure}[b]{.32\columnwidth}
        \centering
        \includegraphics[width=\columnwidth]{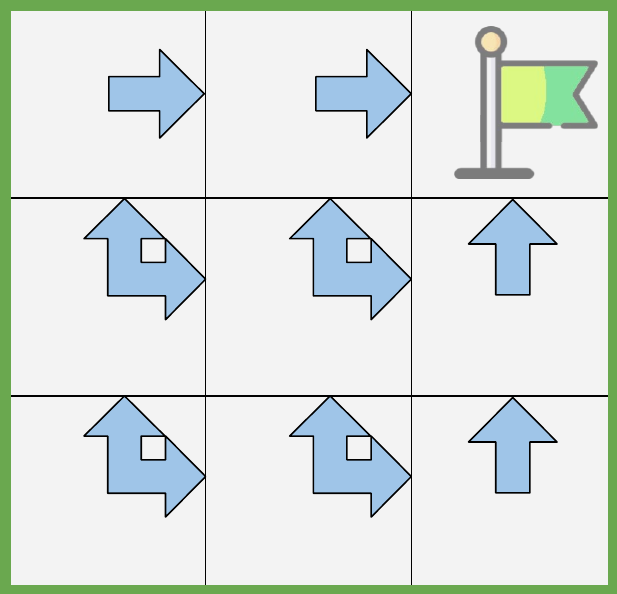}
        \caption{\label{fig:policies_joint_easy}Simple \good}
    \end{subfigure}
    \hfill
    \begin{subfigure}[b]{.32\columnwidth}
        \centering
        \includegraphics[width=\columnwidth]{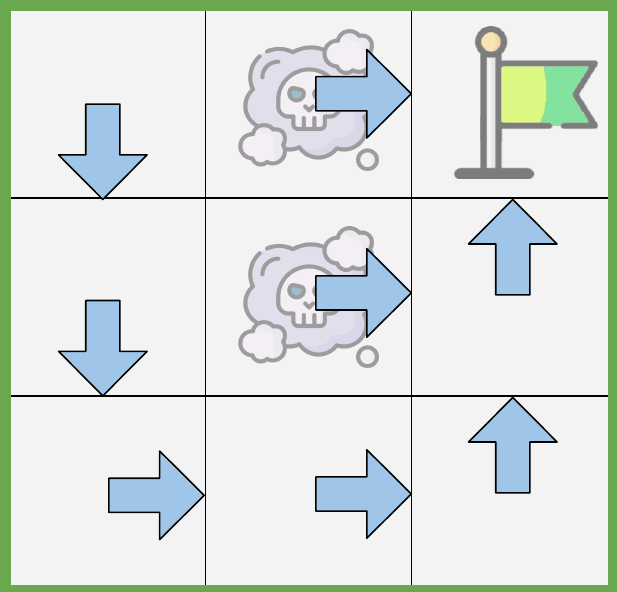}
        \caption{\label{fig:policies_joint_medium}Penalty \good}
    \end{subfigure}
    \hfill
    \begin{subfigure}[b]{.32\columnwidth}
        \centering
        \includegraphics[width=\columnwidth]{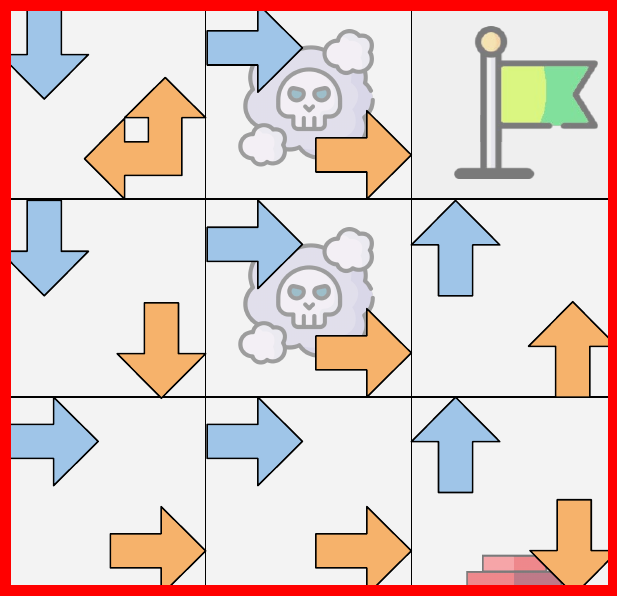}
        \caption{\label{fig:policies_joint_hard}Button \bad}
    \end{subfigure}
    \caption{\label{fig:policies_joint}\textbf{With two Q-functions and joint policy}, the agent hits the wall in the top-left corner of the Button Mon-MDP.}
\end{figure}
\subsubsection{\textbf{Algorithm 3: Two Q-functions (Joint Greedy Policy).}}
\label{subsubsec:joint}
Ignoring samples when $\rprox_t = \rewardundefined$ disregards useful information given by $\rmon_t$. 
To fix this, we decouple the value of states and actions into two Q-functions: $\qenv$ trained using proxy rewards (only when $\rprox_t \neq \rewardundefined$) and $\qmon$ using monitor rewards. This way, even if $\rprox_t = \rewardundefined$ we can still update the latter. 
This begs the question: how should the algorithm greedily select actions when there are two Q-functions?

The first strategy we propose (the Algorithm we are describing) is to select them \emph{jointly} with $(\aenv_t, \amon_t) = \arg\max_{\aenv, \amon} \{\qenv + \qmon\}$. Intuitively, the agent would try to maximize the sum of both rewards simultaneously.
As shown in Figure~\ref{fig:policies_joint}, while able to learn an optimal policy in the Simple and Penalty Mon-MDPs, this variant fails in the Button Mon-MDP. Interestingly, the policy correctly avoids penalty cells, turns \texttt{OFF} the monitor, and goes to the goal in all states \textit{but in the top-left cell when the monitor is \texttt{ON}}. 
This is due to conflicting Q-values. $\qenv$ wants to go \texttt{DOWN} and follow the safe path to the goal. On the contrary, $\qmon$ wants to go \texttt{RIGHT}, step over the penalty cells ($\qmon$ does not accumulate $\renv_t = -10$), and end the episode (to stop receiving $\rmon_t = -0.2$). 
The sum of the Q-values, however, favors neither \texttt{DOWN} nor \texttt{RIGHT}, but \texttt{UP} and \texttt{LEFT}. After all, the $\texttt{max}$ operator of the greedy policy is not linear,\footnote{Given two functions $f(x), g(x)$, $\max_x (f(x) + g(x)) \neq \max_x f(x) + \max_x g(x)$.} thus summing the two Q-functions does not guarantee maximizing both (and, indeed, a single action may not maximize both Q-functions).

\begin{figure}[b]
    \begin{subfigure}[b]{.32\columnwidth}
        \centering
        \includegraphics[width=\columnwidth]{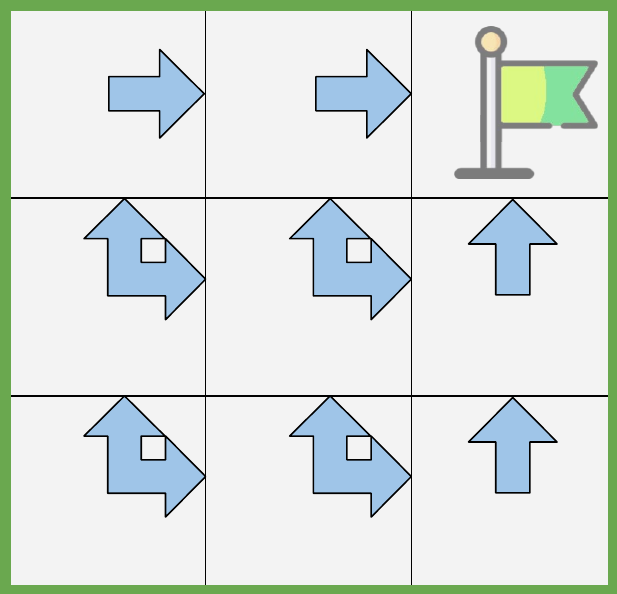}
        \caption{\label{fig:policies_sequential_easy}Simple \good}
    \end{subfigure}
    \hfill
    \begin{subfigure}[b]{.32\columnwidth}
        \centering
        \includegraphics[width=\columnwidth]{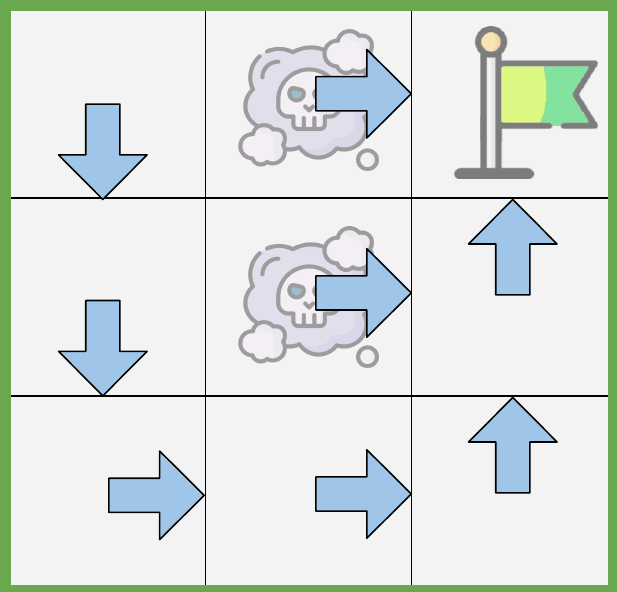}
        \caption{\label{fig:policies_sequential_medium}Penalty \good}
    \end{subfigure}
    \hfill
    \begin{subfigure}[b]{.32\columnwidth}
        \centering
        \includegraphics[width=\columnwidth]{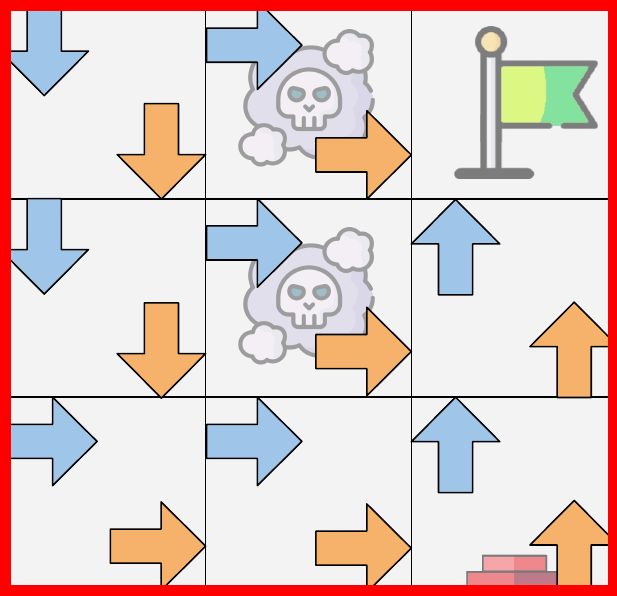}
        \caption{\label{fig:policies_sequential_hard}Button \bad}
    \end{subfigure}
    \caption{\label{fig:policies_sequential}\textbf{Using a sequential policy}, instead, the agent does not turn \texttt{OFF} monitoring in the Button Mon-MDP.}
\end{figure}
\subsubsection{\textbf{Algorithm 4: Two Q-functions (Sequential Greedy Policy).}}
\label{subsubsec:sequential}
To avoid conflicting Q-values, we modify the action selection so that the agent selects \textit{first} $\aenv_t = \arg\max_{\aenv} \qenv$ and \textit{then} $\amon_t = \arg\max_{\amon}\qmon\mid_{\aenv_t}$. 
Thus, this agent prioritizes $\qenv$, as maximizing $\qmon$ is subject to the greedy environment action. 
The policy in Figure~\ref{fig:policies_sequential} still fails in the Button Mon-MDP, as the agent does not turn \texttt{OFF} the monitor on its way to the goal. This happens because there are no explicit monitor actions and the agent must use environment actions to turn it \texttt{OFF}. Yet, going \texttt{DOWN} in the bottom-right cell to press the button is not optimal for $\qenv$. Since $\aenv$ is selected greedily using $\qenv$, the agent goes to the goal ignoring the button.

\subsubsection{\textbf{Algorithm 5: Learn a Reward Model.}}
\label{subsubsec:model}
\begin{figure}[t]
    \begin{subfigure}[b]{\columnwidth}
        \centering
        \includegraphics[trim=0 41pt 0 0, width=0.9\columnwidth]{fig/policies/legend_v5.pdf}
    \end{subfigure}
    \\
    \begin{subfigure}[b]{.32\columnwidth}
        \centering
        \includegraphics[width=\columnwidth]{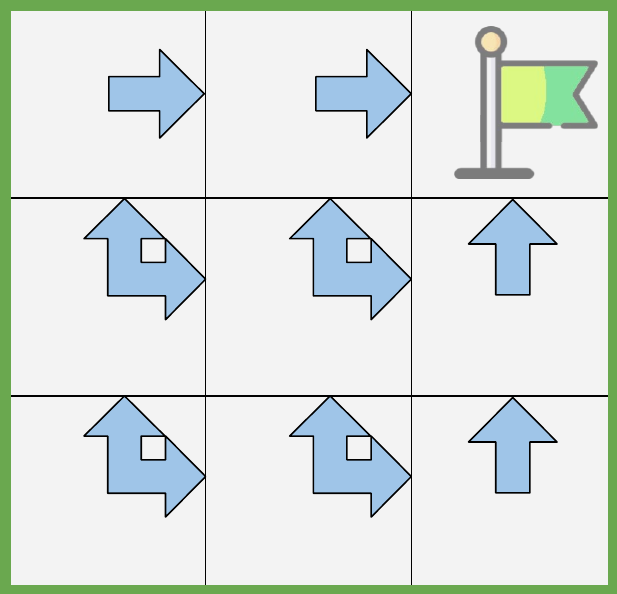}
        \caption{\label{fig:policies_model_easy}Simple \good}
    \end{subfigure}
    \hfill
    \begin{subfigure}[b]{.32\columnwidth}
        \centering
        \includegraphics[width=\columnwidth]{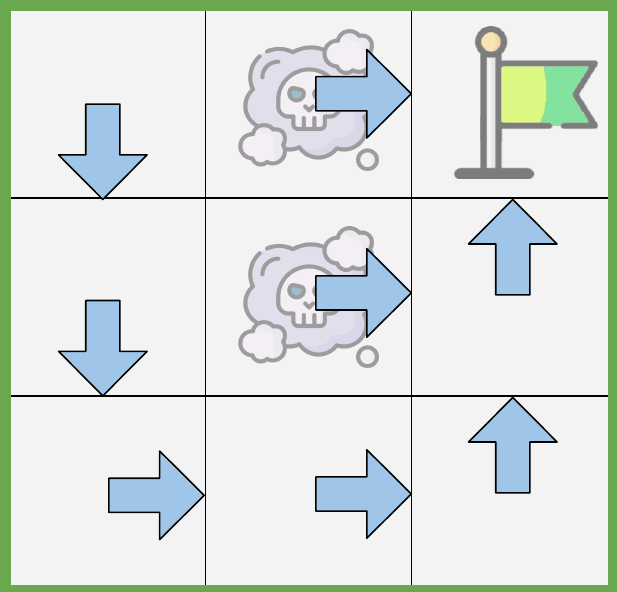}
        \caption{\label{fig:policies_model_medium}Penalty \good}
    \end{subfigure}
    \hfill
    \begin{subfigure}[b]{.32\columnwidth}
        \centering
        \includegraphics[width=\columnwidth]{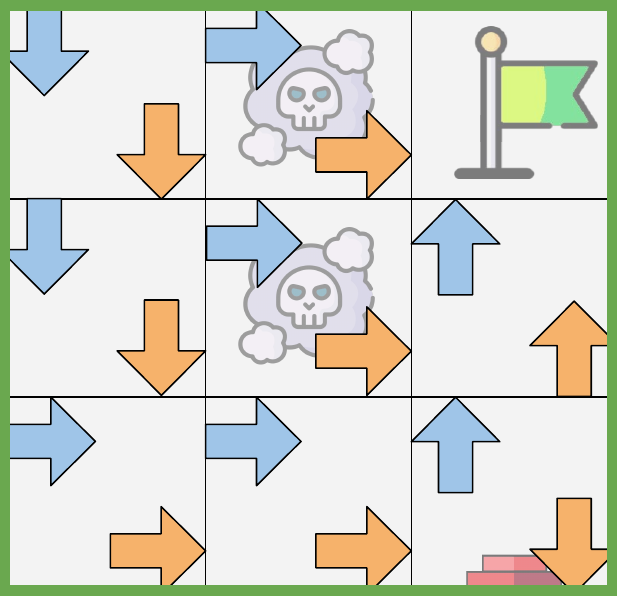}
        \caption{\label{fig:policies_model_hard}Button \good}
    \end{subfigure}
    \caption{\label{fig:policies_model}\textbf{With a reward model}, the agent learns the optimal policy in all three Mon-MDPs.}
\end{figure}

The agent replaces $\rprox_t = \rewardundefined$ with the reward predicted by a model. In discrete Mon-MDPs, the model is a table like the Q-function that stores the running mean of the environment rewards as it observes $\rprox_t = \renv_t$ (see Section \ref{subsec:proof} for more details). 
This algorithm converges to an optimal policy in all Mon-MDPs, as shown in Figure~\ref{fig:policies_model}. 
Intuitively, the reward model allows the agent to know the current reward $\renv_t$ even without observing it.
We note, however, that this algorithm works because all three Mon-MDPs satisfy the conditions of Proposition~\ref{prop:convergence}. 
In Section~\ref{subsec:proof}, we formally prove the convergence to an optimal policy of Algorithm 5 according to Proposition~\ref{prop:convergence}.

\subsubsection{\textbf{Remarks.}}
We emphasize that Algorithms 3, 4, and 5 (Joint, Sequential, Reward Model) solve the Penalty Mon-MDP because they can reason about the environment independently from the monitor. 
As discussed in Section~\ref{subsec:remarks}, because the monitor and the environment are decoupled, the agent learns that the monitor does not change the environment reward, and that walking on penalty cells is undesirable even if not monitored. 
If the agent walks on penalty cells while monitored (a similar action to spilling water with the owner at home) and observes a negative reward, it will learn that the reward would still be negative even when the monitor (the owner) is not providing it. 
Joint and Sequential decouple environment and monitor with two Q-functions, 
Reward Model with a reward model that depends only on the environment. 
However, Joint and Sequential fail in the Button Mon-MDP. 
In Appendix~\ref{app:sec:alg_details}, we discuss about stricter conditions of convergence for Sequential, and about the lack of guarantees of convergence for Joint.

\subsection{Proof of Proposition \ref{prop:convergence}}
\label{subsec:proof}

We return now to prove Proposition~\ref{prop:convergence} from Section~\ref{subsec:solving_monmdps}, and use Q-Learning with a reward model as the candidate algorithm that can solve all finite Mon-MDPs satisfying Properties~\ref{def:ergodic_mmdp},~\ref{def:env_ergodic_mon}, and~\ref{def:truthful_mon}. 

\begin{Proof}
\label{proof:q_model_conv}
Consider Q-Learning on a finite Mon-MDP that replaces the observed reward $\rprox_t$ with the running average of observed proxy reward stored in a table $\smash{\widehat{R}(\senv, \aenv)}$, i.e.,
\begingroup
\allowdisplaybreaks
\setlength{\belowdisplayskip}{2pt}
\setlength{\belowdisplayshortskip}{2pt}
\begin{align}
    \widehat{Q}(s_t,a_t) \coloneqq & \textstyle \mathbb{E}[\sum\nolimits_{i=t}^\horizon \upgamma^{i-t} \big( \widehat{R}(\senv_i, \aenv_i) + \rmon_i \big) \mid \pi, \probmodel, s_t, a_t] \nonumber
    \\[2pt]
    N_{k+1}(\senv_t, \aenv_t) \leftarrow & N_{k}(\senv_t, \aenv_t) + 1 \tag*{if $\,\rprox_t \neq \rewardundefined$}
    \\[0pt]
    \widehat{R}_{k+1}(\senv_t, \aenv_t) \leftarrow & \frac{\left(N_{k+1}(\senv_t, \aenv_t) - 1\right)\widehat{R}_{k}(\senv_t, \aenv_t) + \rprox_t}{N_{k+1}(\senv_t, \aenv_t)} \hfill \tag*{if $\,\rprox_t \neq \rewardundefined$}
    \\[0pt]
    \widehat{Q}_{k+1}(s_t, a_t) \leftarrow & (1 - \upalpha_t) \widehat{Q}_{k}(s_t, a_t) \, + \nonumber
    \\[-2pt]
    & \; \upalpha_t \big(\widehat{R}_{k+1}(\senv_t, \aenv_t) + \rmon_t + \upgamma {\max_a \widehat{Q}_{k+1}(s_{t+1}, a)}\big) \label{eq:qlearning_model} 
\end{align}
\endgroup
where $k$ denotes the $k$-th update, and $N(\senv_t, \aenv_t)$ is a count that increases every time the agent observes a reward, i.e., if $\rprox_t \neq \rewardundefined$. Then, this algorithm converges to an optimal policy in Eq.~\eqref{eq:max_pi} if (a) the policy is greedy in the limit with infinite exploration (GLIE), and (b) the learning rate $\upalpha_t$ satisfies the Robbins-Monro conditions~\citep{robbins1951stochastic}. 

\begin{enumerate}[leftmargin=2em, itemsep=1pt, topsep=3pt, label=(\Roman*)]
    \item Under a GLIE policy, the agent will visit every state infinitely often (Property~\ref{def:ergodic_mmdp}), will observe a reward for every state (Property~\ref{def:env_ergodic_mon}), and the observed reward will be the environment reward (Property~\ref{def:truthful_mon}). Therefore, the agent will observe the environment reward for every environment state-action pair infinitely often. \label{proof:1} 
    \item Under~\ref{proof:1} and by the central limit theorem, 
    \begingroup
    \setlength{\abovedisplayskip}{3pt}
    \setlength{\belowdisplayskip}{-6pt}
    \setlength{\abovedisplayshortskip}{3pt}
    \setlength{\belowdisplayshortskip}{-6pt}
    \begin{align*}
        \textstyle \lim\nolimits_{k \to \infty} \widehat{R}_k(\senv, \aenv) &= \EVV{}{\renv \mid \pi, \envprobmodel}
    \end{align*} \label{proof:2} 
    \endgroup
    \item Given \ref{proof:2} and because of linearity of expectation, maximizing the Q-function learned using rewards from $\smash{\widehat{R}(\senv, \aenv)}$ approaches the original optimization problem of Eq.~\eqref{eq:q}, i.e., \label{proof:3} 
    \begingroup
    \setlength{\abovedisplayskip}{3pt}
    \setlength{\belowdisplayskip}{-2pt}
    \setlength{\abovedisplayshortskip}{3pt}
    \setlength{\belowdisplayshortskip}{-2pt}
    \begin{align*}
        \setlength{\abovedisplayskip}{0pt}
        \setlength{\belowdisplayskip}{-20pt}
        \setlength{\abovedisplayshortskip}{0pt}
        \setlength{\belowdisplayshortskip}{-10pt}
        \textstyle \widehat{Q}_k(s_t,a_t) & \underset{\phantom{k \to \infty}}{\coloneqq} \textstyle \mathbb{E}[\sum\nolimits_{i=t}^\horizon \upgamma^{i-t} \big( \widehat{R}_k(\senv_i, \aenv_i) + \rmon_i \big) \mid \pi, \probmodel, s_t, a_t] 
        \\[-3pt]
        &\underset{\smash{k \to \infty}}{=} \textstyle \mathbb{E}[\sum\nolimits_{i=t}^\horizon \upgamma^{i-t} \big( \EVV{}{\renv_i\mid \pi, \envprobmodel} + \rmon_i \big) \mid \pi, \probmodel, s_t, a_t] 
        \\[-2pt]
        &\underset{\phantom{k \to \infty}}{=} \textstyle \mathbb{E}[\sum\nolimits_{i=t}^\horizon \upgamma^{i-t} \big( \renv_t + \rmon_i \big) \mid \pi, \probmodel, s_t, a_t] 
        \\[-3pt]
        &\underset{\phantom{k \to \infty}}{\eqqcolon} Q(s_t,a_t)
    \end{align*}
    \endgroup
    \item Given \ref{proof:3}, Eq.~\eqref{eq:qlearning} and~\eqref{eq:qlearning_model} are equivalent in the limit. Under a GLIE policy and if the learning rate $\upalpha_t$ satisfies the Robbins-Monro conditions, Eq.~\eqref{eq:qlearning} converges to the Q-function of an optimal greedy policy~\citep{dayan1992convergence, bertsekas1996neuro, melo2001convergence}.%
    \footnote{A greedy optimal policy always exists for MDPs with finite states and actions, stationary reward and transition functions, bounded rewards, and $\upgamma \in [0, 1)$~\citep{puterman1994markov}.}
\end{enumerate}
\end{Proof}

\begin{figure*}[ht]
    \captionsetup[subfigure]{aboveskip=2pt}
    \centering
    \raisebox{27pt}{\rotatebox[origin=t]{90}{\includegraphics[width=0.11\linewidth]{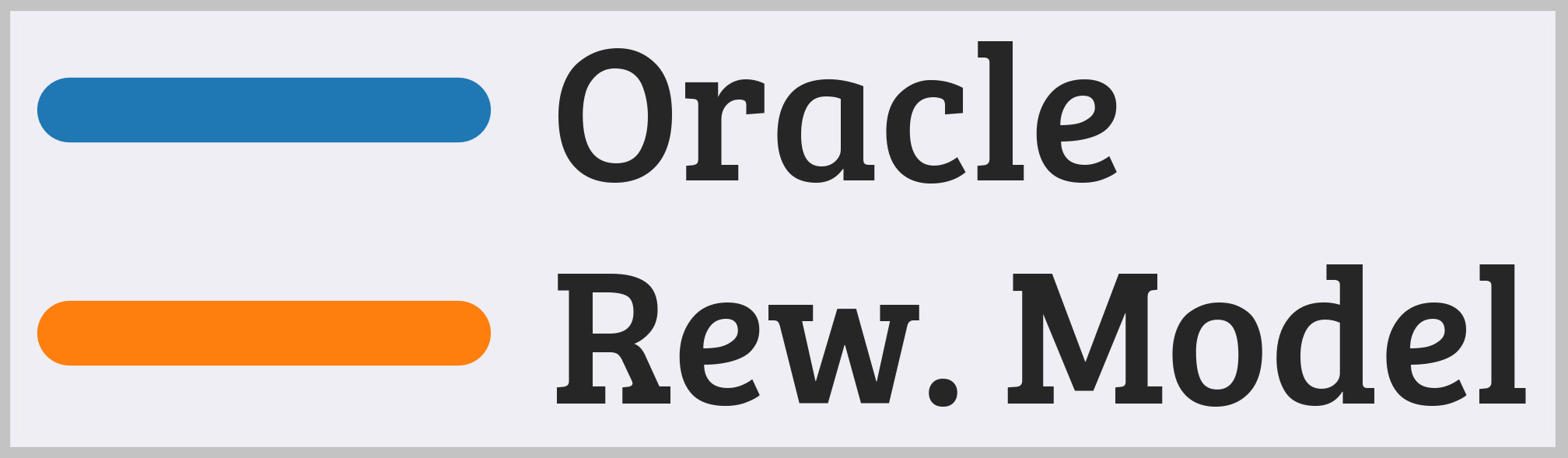}}}
    \hfill
    \begin{subfigure}[b]{0.155\linewidth}
        \centering
        \includegraphics[width=\linewidth]{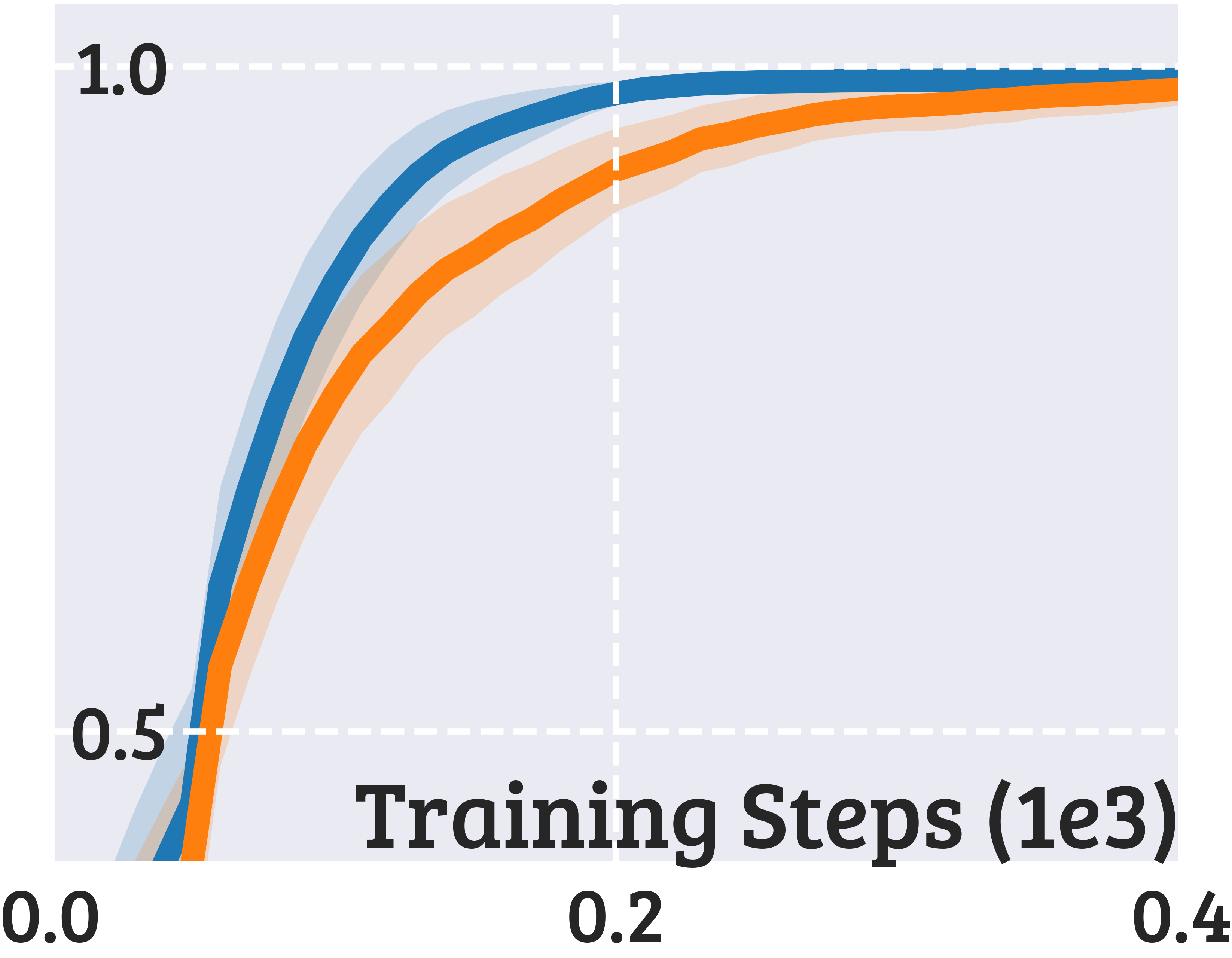}
        \caption{Simple}
    \end{subfigure}
    \hfill
    \begin{subfigure}[b]{0.155\linewidth}
        \centering
        \includegraphics[width=\linewidth]{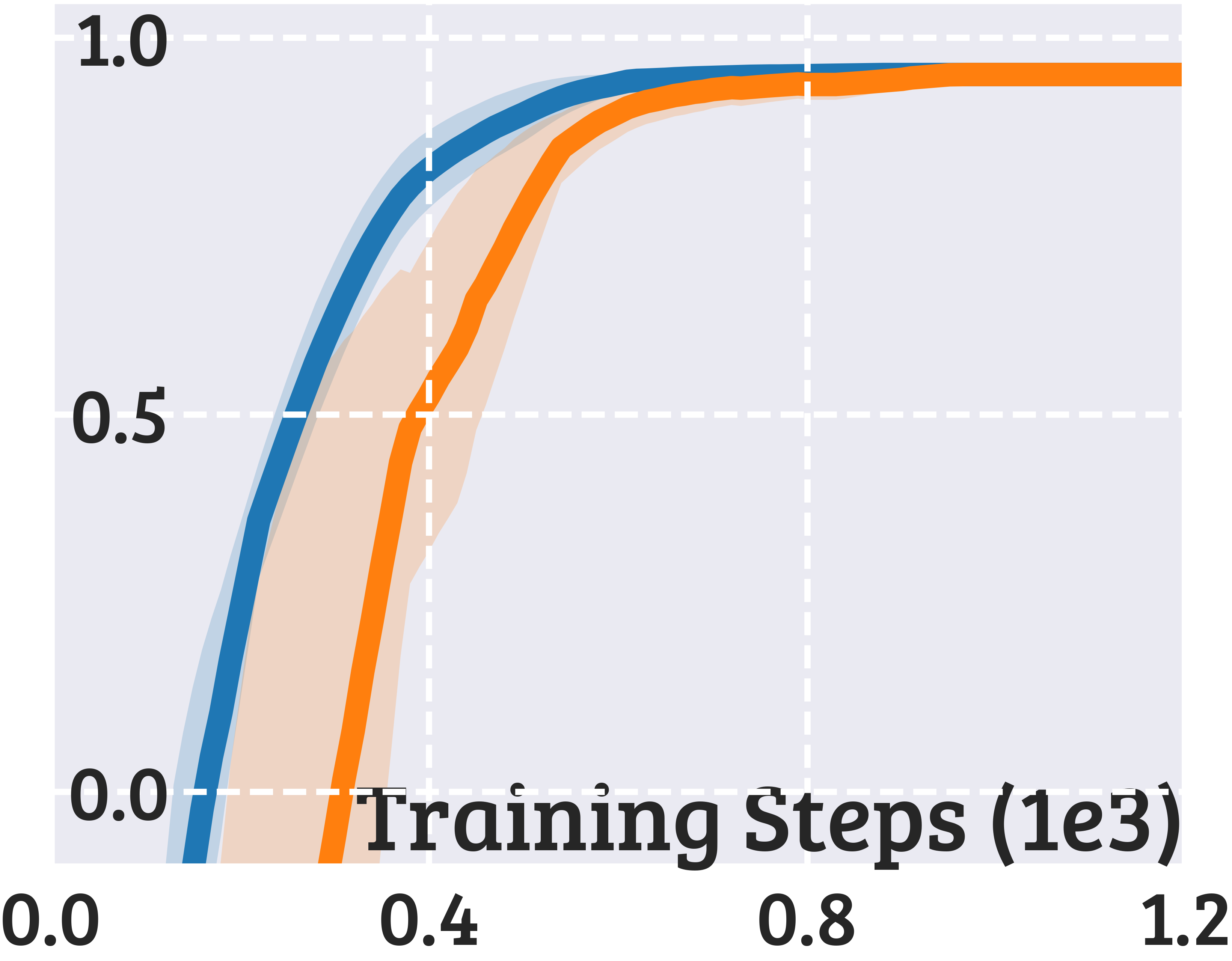}
        \caption{Penalty}
    \end{subfigure}
    \hfill
    \begin{subfigure}[b]{0.155\linewidth}
        \centering
        \includegraphics[width=\linewidth]{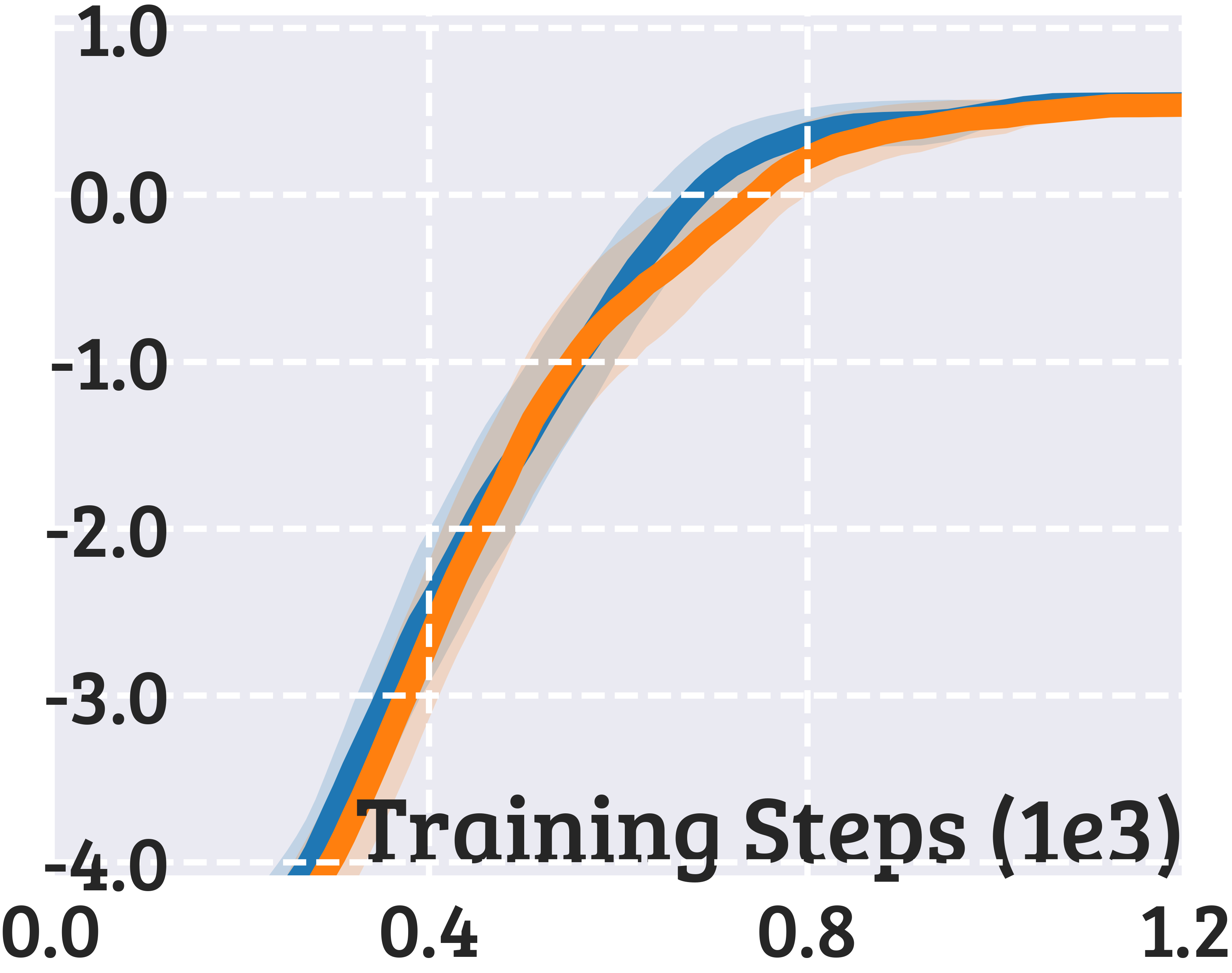}
        \caption{Button}
    \end{subfigure}
    \hfill
    \begin{subfigure}[b]{0.155\linewidth}
        \centering
        \includegraphics[width=\linewidth]{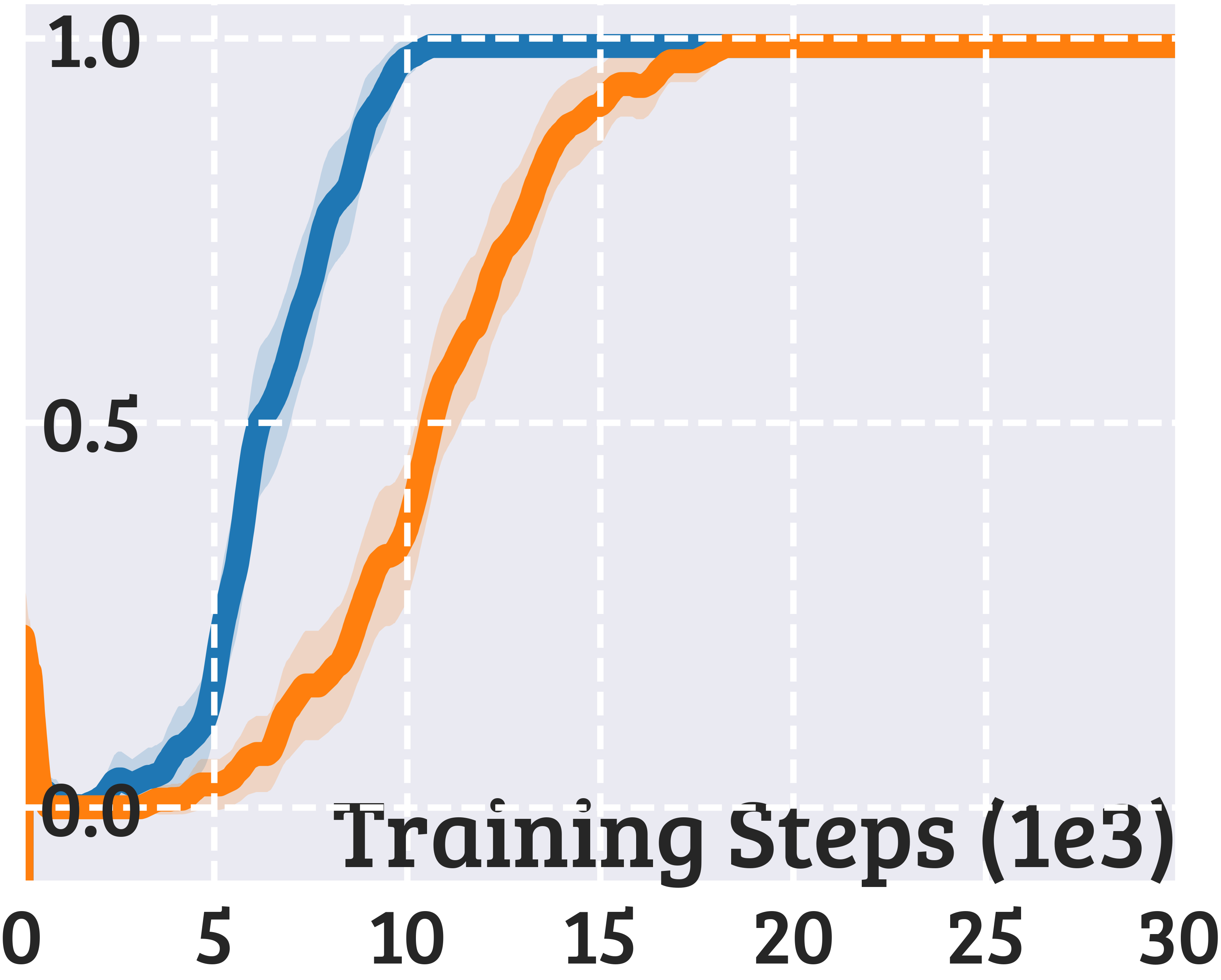}
        \caption{Simple (Noisy)}
    \end{subfigure}
    \hfill
    \begin{subfigure}[b]{0.155\linewidth}
        \centering
        \includegraphics[width=\linewidth]{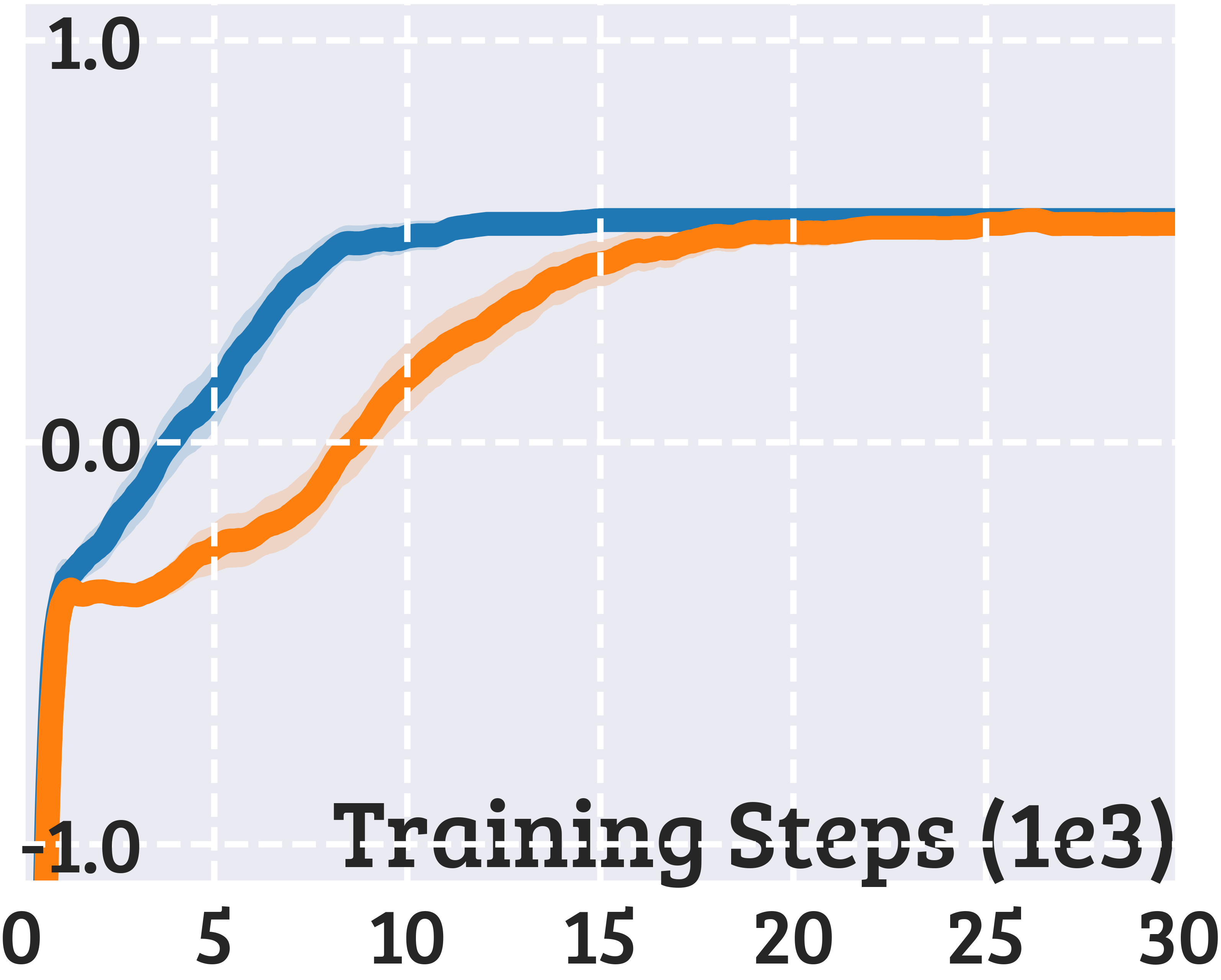}
        \caption{Penalty (Noisy)}
    \end{subfigure}
    \hfill
    \begin{subfigure}[b]{0.155\linewidth}
        \centering
        \includegraphics[width=\linewidth]{fig/plots/main/Gridworld-Medium-3x3-Stochastic-v0_mes50/iToySwitchMonitor.png}
        \caption{Button (Noisy)}
    \end{subfigure}
    \caption{\label{fig:plots_short}\textbf{Episode return $\smash{\sum_{t=1}^T\upgamma^{t-1}(\renv_t + \rmon_t)}$ of greedy policies averaged over 100 seeds (shades denote 95\% confidence interval).}}
\end{figure*}

\subsection{Empirical Rate of Convergence}
\label{subsec:rate}
One assumption needed in Proof~\ref{proof:q_model_conv} is that Q-Learning uses a GLIE policy. This, together with Mon-MDP ergodicity (Property~\ref{def:ergodic_mmdp}) guarantees that the agent will visit every state-action pair and observe every reward. But how hard is exploration in Mon-MDPs? The agent will not observe the environment reward all the time, and cannot learn an optimal policy until it has seen them all sufficiently often. 
How does this affect the rate of convergence to an optimal policy? 
Intuitively, if some rewards are unobservable learning will be slower, more so if the environment reward function is noisy. 

In this section, we empirically investigate the rate of convergence of ``Q-Learning with Reward Model'' presented in Section~\ref{subsubsec:model} against an ``Oracle'' Q-Learning. The Oracle executes monitor actions and receives monitor rewards, but always observes $\rprox_t = \renv_t$.
For each Mon-MDP, we consider also a version where the environment reward has Gaussian noise with standard deviation 0.05. 
For all details about the experiments, more plots and table, and an additional evaluation on harder Mon-MDPs, refer to Appendix~\ref{app:subsec:rate}.

Figure~\ref{fig:plots_short} shows that the \textcolor{black}{Oracle} always converges faster to an optimal policy, up to $\times$2 faster with noisy rewards. 
While $\smash{\widehat{R}}$ compensates for the unobservability of rewards, the agent still needs to observe the rewards sufficiently often --- {especially if they are noisy} --- for the model to be accurate. 
In Appendix~\ref{app:subsec:rate}, we show results on Mon-MDPs with larger monitor spaces and richer dynamics, where the gap between the Oracle and Reward Model is even larger.

\section{Future Work}
\label{sec:future}
Throughout the paper, we discussed how Mon-MDPs relate to prior work, and our empirical study has highlighted important challenges that would benefit from existing RL techniques. 
Below, we describe some of the most interesting directions of future research that this work opens up, and connect them to existing areas of research such as meta RL, model-based RL, cautious RL, and intrinsic motivation.

\textbf{Convergence, Bounds, and Connection to Partial Monitoring.}
Mon-MDPs are a new framework and therefore open to further theoretical analysis. 
First and foremost, we have presented a set of \textit{sufficient} conditions for convergence, but these may not be \textit{necessary}. For instance, the monitor may not need to be truthful, as suggested by prior work on reward shaping~\citep{ng1999policy}. Relaxing these conditions will likely pose an additional challenge that could be addressed by having a belief over the reward~\citep{marom2018belief}. 
Furthermore, it would be interesting to investigate the convergence bounds of monitored algorithms. For example, similar research proved regret bounds for many partial monitoring bandits~\citep{bartok2014partial, auer2002nonstochastic, lattimore2019information}.

\textbf{Generalization, Train-And-Deploy, and Meta RL.} 
In this work, we considered finite Mon-MDPs and assumed properties on the Mon-MDP that may not hold in real-world problems, e.g., monitor ergodicity and truthfulness. 
Can the agent learn an optimal policy even when these properties are not satisfied?

Consider the agent in Figure~\ref{fig:intro}, but this time it can \textit{never} be monitored while watering plants and --- if it spills water it will never receive a negative feedback. 
However, the agent can be monitored when cleaning dishes. Can it learn that spilling water is undesirable by receiving negative feedback for spilling water while cleaning dishes?
This requires 1) reasoning over the monitor and the environment independently --- spilling water is undesirable regardless of the monitor state --- and 2) generalization across environment states and actions --- spilling water in the kitchen and spilling water is equally undesirable. In Section~\ref{subsec:remarks}, we argued that the Mon-MDP framework already allows the former reasoning. For the latter, we need to incorporate generalization and go beyond finite Mon-MDPs.

More generally, Mon-MDPs can be further extended to consider situations where the agent must act in unmonitored environments --- where rewards are \textit{never} observable --- after being trained in a monitored environment. This is closely related to train-and-deploy and meta RL settings~\citep{wang2018prefrontal, matsushima2020deploymentefficient}, and requires the ability to generalize knowledge about rewards across states --- possibly of different environments --- to compensate for their unobservability.

\textbf{Unsolvable Mon-MDPs.}
What if the agent cannot learn an optimal policy because some rewards are \textit{never} observable? 
While it may be impossible to act optimally with respect to environment rewards, the agent should still act ``optimally'' according to what it can observe.
In this regard, it is interesting to consider algorithms that can tackle \textit{unsolvable Mon-MDPs}, i.e., can learn ``useful'' policies in Mon-MDPs where it is impossible to learn an optimal policy due to unobservability of the rewards. 
In Appendix~\ref{app:sec:taxonomy}, we formally discuss the notion of solvability from a theoretical point of view and set the stage for future directions of research.
For example, the best way to act in situations of uncertainty is still a matter of dispute in RL and relates to cautious and risk-averse RL~\citep{mohammedalamen2021learning, zhang2020cautious}.

\textbf{Exploration.} In Section~\ref{subsec:rate}, we showed that unobservable rewards make exploration significantly harder.
Clearly, naive $\varepsilon$-greedy exploration does not exploit the complexity of Mon-MDPs, and we believe there are exciting potential improvements.
In particular, as discussed in Section~\ref{subsec:remarks}, explicitly reasoning on monitor and environment separately facilitates better exploration and more advanced behaviors. 
For example, the agent could use intrinsic motivation~\citep{parisi2021interesting, mutti2021task} to prefer environment states for which it has not observed the reward yet. 
At the same time, it could try new actions in states where it knows it will be monitored.

\section{Conclusion}
\label{sec:conclusion}

MDPs offer a framework to tackle decision-making problems, but the assumption of reward observability is not descriptive of all real-world problems.
To account for situations where the agent cannot observe the rewards generated by the environment to judge its actions, we presented \textit{Monitored MDPs}. We discussed the theoretical and practical consequences of unobservable rewards, and presented toy environments and algorithms to illustrate subsequent challenges.
While prior work on active RL and partial monitoring has addressed partially observable rewards, this is --- to the best of our knowledge --- the first work that presents a generic formalism allowing for sequential decision-making without requiring the monitor to have explicit binary monitoring actions.

\vspace*{2pt}

\textit{In the same way that RL built its foundation starting from theoretical analyses on discrete MDPs and the empirical investigations of chainworlds and gridworlds, with this work we aim to set the stage for future research ranging from theoretical analysis of stronger guarantees of convergence, development of better
algorithms, and practical applications of Mon-MDPs to real-world problems.}

\section*{Acknowledgements}
This research was supported by grants from the Alberta Machine Intelligence Institute (Amii); a Canada CIFAR AI Chair, Amii; Digital Research Alliance of Canada; Huawei; Mitacs; and NSERC.

\clearpage

\balance

\bibliographystyle{ACM-Reference-Format} 
\bibliography{rl_bib, cl_bib}

\clearpage

\appendix

\newgeometry{left=2.5cm, right=2.5cm, top=3cm, bottom=3cm}

\textbf{\Huge Appendix}

\section{Table of Notation}
Table~\ref{tab:notation} summarizes the notation used in this paper. We recall that the environment and the monitor have their own spaces and rewards, and obey to their own Markovian transition functions. If we consider joint states, actions, and rewards, we have a notation similar to classic MDPs where the agent observes tuples $(s_t, a_t, \hat r_t, s_{t+1})$. 

\begin{table}[h]
\begin{minipage}[c]{0.54\textwidth}    
    \setlength\tabcolsep{2pt}
    \renewcommand{\arraystretch}{1.05}
    \centering
    \caption{\label{tab:notation}\textbf{Mon-MDP notation.}}
    \vspace*{-3pt}
    \begin{tabular}{@{\extracolsep{\fill}}|l|l|l|}
        \hline
        \textbf{Symbol} & \textbf{Definition} & \textbf{Comment} 
        \\
        \hline
        \Tstrut
        $\senv \in \envstatespace$ & Envir. state space & 
        \\ 
        $\aenv \in \envactionspace$ & Envir. action space & 
        \\
        $\envprobmodel (\senv_{t+1} \mid \senv_t , \aenv_t)$ & Envir. transition func. & 
        \\
        $\renv_t \sim \envrewardmodel (\senv_t , \aenv_t)$ & Envir. reward & $\renv \in \realspace$ 
        \Bstrut 
        \\
        \hline
        \Tstrut
        $\smon \in \monstatespace$ & Monitor state space & 
        \\
        $\amon \in \monactionspace$ & Monitor action space & 
        \\
        $\monprobmodel (\smon_{t+1} \mid \senv_t, \smon_t, \aenv_t, \amon_t)$ & Monitor trans. func. & 
        \\
        $\rmon_t \sim \monrewardmodel (\smon_t, \amon_t) $ & Monitor reward & $\rmon \in \realspace$ 
        \\
        $\rprox_t \sim \monitormodel (\renv_t, \smon_t, \amon_t) $ & Proxy reward & $\rprox \in {\realspace}_{\rewardundefined} \coloneqq \realspace \cup \{\rewardundefined\}$ 
        \\
        $\monitormodel$ & Monitor function & 
        \Bstrut
        \\
        \hline
        \Tstrut
        $s \in \joinstatespace$ & Joint state space & $\joinstatespace \coloneqq \envstatespace \times \monstatespace$ 
        \\
        $a \in \joinactionspace$ & Joint action space & $\joinactionspace \coloneqq \envactionspace \times \monactionspace$ 
        \\
        $r_t = \renv_t + \rmon_t $ & Joint reward & $r_t \in \realspace$ 
        \\
        $\hat r_t = (\rprox_t, \rmon_t)$ & Observed reward & $\hat{r}_t \in {\realspace}_{\rewardundefined} \times \realspace$ 
        \Bstrut
        \\
        \hline
        \Tstrut
        $\pi(a \mid s)$ & Joint policy & 
        \\
        $\pienv(\aenv \mid \senv)$ & Envir. policy & 
        \\
        $\pimon(\amon \mid \smon, \amon, \senv)$ & Monitor policy & 
        \Bstrut
        \\
        \hline
    \end{tabular}
\end{minipage}
\hfill
\begin{minipage}[c]{0.44\textwidth}    
    \centering
    \includegraphics[width=0.85\linewidth]{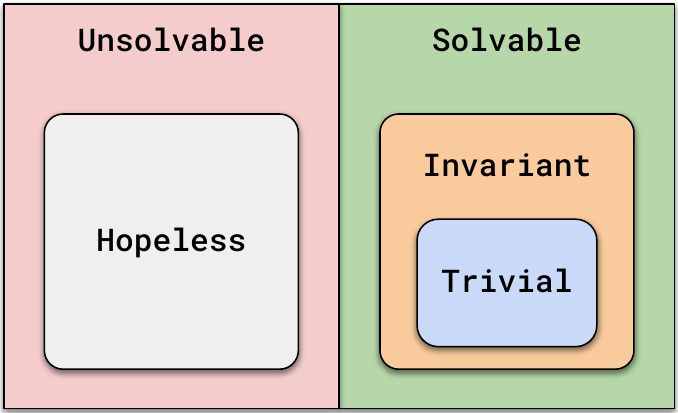}
    \captionof{figure}{\label{fig:mmdp_taxonomy} \textbf{Mon-MDPs taxonomy.} \textmd{Using the indistinguishability relation defined in Eq.~\eqref{eq:indisting}, we can classify Mon-MDP into two disjoint sets: \textit{solvable} (the agent can learn an optimal policy) and \textit{unsolvable} (the agent cannot). In \textit{hopeless} Mon-MDPs the agent cannot learn anything, e.g., because the environment reward is never observable. In contrast, in \textit{trivial} Mon-MDP the environment reward is always observable and the Mon-MDP is equivalent to a traditional MDP. Finally, in \textit{invariant} Mon-MDP, the monitor reward does not change optimality with respect to the environment MDP --- i.e., there is a policy maximizing both $\sum \upgamma^{t-1} (\renv_t + \rmon_t)$ and $\sum \upgamma^{t-1} \renv_t$.}}
\end{minipage}
\end{table}

\section{Taxonomy of Mon-MDPs}
\label{app:sec:taxonomy}
In Section~\ref{subsec:mon_mdps}, we formulated the Mon-MDP framework as generic as possible, and the monitor function --- the core of Mon-MDPs --- has no constraints. For instance, it can modify the reward randomly or even always return $\rprox_t = \rewardundefined \; \forall t$. Clearly, these are Mon-MDPs where no algorithm could learn anything meaningful, as no useful information is given to the agent. 
On the other hand, we discussed sufficient conditions for the existence of an algorithm that converges to an optimal policy, but these conditions are not always satisfied by real-world problems. 
What about other cases? Can we define when a Mon-MDP can actually be solved? And if not, is it hopeless or is there still something that can be learned?

In this section, first we formally define when Mon-MDPs are \textit{solvable} from first principles. 
We further classify Mon-MDPs according to the taxonomy presented in Figure~\ref{fig:mmdp_taxonomy}, focusing on Mon-MDPs that are not solvable. In particular, we discuss what an agent can learn in such \textit{unsolvable} Mon-MDPs, suggesting an alternative objective to learn ``reasonable'' policies.

\begin{figure}[h]
    \begin{minipage}[b]{0.425\textwidth}
        \begin{subfigure}[b]{0.495\textwidth}
            \centering
            \includegraphics[width=0.9\columnwidth]{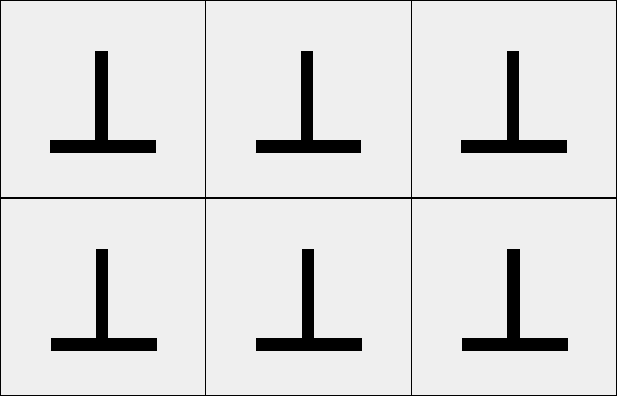}
            \caption{\label{fig:hopeless}Hopeless Mon-MDP}
        \end{subfigure}
        \hfill
        \begin{subfigure}[b]{0.495\textwidth}
            \centering
            \includegraphics[width=0.9\columnwidth]{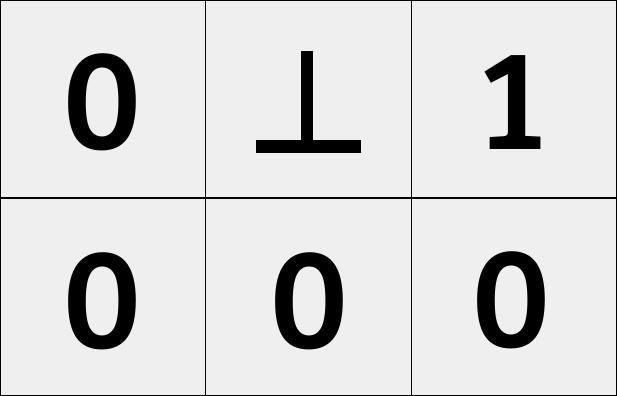}
            \caption{\label{fig:non-hopeless}Non-Hopeless Mon-MDP}
        \end{subfigure}
    \end{minipage}
    \hfill
    \begin{minipage}[b]{0.555\textwidth}
        \centering
        \caption{\label{fig:examples}\textbf{Proxy rewards of unsolvable gridworld Mon-MDPs.} \textmd{If a state always yields $\rprox_t = \rewardundefined$, the agent cannot know if it is good or bad. For all it knows, its environment reward could be anything, i.e., agent could be in any Mon-MDP. However, not all unsolvable Mon-MDPs are equally unsolvable. In Figure~\subref{fig:hopeless} \textit{all rewards are always unobservable}, thus the agent cannot learn anything about the environment --- all it can do is maximize monitor rewards. In contrast, in Figure~\subref{fig:non-hopeless} only one state yields $\renv_t = \rewardundefined$, and the agent can still learn to go to the +1 observable reward.}}
    \end{minipage}
\end{figure}

\subsection{Indistinguishability}

First, let $\monspace_{\indistingspace}$ be the set of all Mon-MDPs with the same joint state space $\statespace$, joint action space $\actionspace$, joint transition function $\probmodel$, and monitor reward function $\monrewardmodel$.
Let $\tau^{(T)} = \{s_t, a_t, \hat{r}_t, s_{t+1}\}_{t = 1 \ldots T}$ be a trajectory of length $T$ of the components observable to the agent (i.e., $\hat{r}_t$ but not $r_t$).  Let $\smash{\trajspace_{\indistingspace} \coloneqq \{\tau^{(T)} \mid \forall T\in \naturalspace^+, \, \forall m \in \monspace_{\indistingspace}\}}$ be the set of all such finite-length trajectories possible within $\monspace_{\indistingspace}$.  Finally, let $\Pi_{\indistingspace}$ be the set of all policies within $\monspace_{\indistingspace}$.  Using this notation we can define when the unobservability of rewards makes two Mon-MDPs indistinguishable.

\begin{Definition}[Indistinguishability $\indisting$] 
\label{def:indisting} 
The indistinguishability relation $\indisting$ between two Mon-MDPs $m_1, m_2 \in \monspace_{\indistingspace}$, denoted with $m_1 \,\indisting \,m_2$, is defined as
\begin{equation}
    \label{eq:indisting}
    \forall \pi \in \Pi_{\indistingspace}, \, \forall \tau \in \trajspace_{\indistingspace}: P(\tau \mid m_1, \pi) = P(\tau \mid m_2, \pi).
\end{equation}
That is, if two Mon-MDPs have the same probability of producing all possible finite-horizon trajectories of the observable components of the Mon-MDPs (that is, $s$, $a$, and $\rmon$), then the agent cannot distinguish which of two Mon-MDPs it finds itself in. 
\end{Definition}

Indistinguishability is an equivalence relation because it is:
\begin{itemize}[leftmargin=3em, itemsep=1pt, topsep=3pt]
    \item Reflexive: $\forall m \in \monspace_{\indistingspace} : m \,\indisting\, m$.
    \item Symmetric: $\forall m_1, m_2 \in \monspace_{\indistingspace} : m_1 \, \indisting \, m_2 \Rightarrow m_2 \, \indisting \, m_1$.
    \item Transitive: $\forall m_1, m_2, m_3 \in \monspace_{\indistingspace} : m_1 \,\indisting\, m_2 \, \land \, m_2 \,\indisting\, m_3 \Rightarrow m_1 \,\indisting\, m_3$.
\end{itemize}

Note that, because we consider Mon-MDPs with the same $\langle \statespace, \actionspace, \probmodel, \monrewardmodel \rangle$, the monitor function $\monitormodel$ is the only discriminating factor of a trajectory. That is, the observed proxy reward is what tells apart which Mon-MDP a trajectory is from. 
For example, consider the Mon-MDP in Figure~\ref{fig:non-hopeless}, where the proxy reward is never observable in the \texttt{TOP-CENTER} state. 
Mon-MDPs where the actual (unobservable) environment reward $\renv_t$ is either $-1, 0, 1, \ldots$ are all indistinguishable between each other, because all the agent observes is $\rprox_t = \rewardundefined$.
Figure~\ref{fig:hopeless} shows an extreme case where \textit{all} rewards are unobservable, and therefore \textit{all} Mon-MDPs with the same state-action space, dynamics (that can be inferred from observations), and monitor reward are indistinguishable from each other.

\subsection{Solvability}
Because indistinguishability is an equivalence relation, we can use it to partition the set of all Mon-MDPs into disjoint subsets such that all members of a subset are indistinguishable from each other.  Let $\{m\}_\indisting$ be the subset containing $m$.

\begin{Definition}[Solvable Mon-MDP]
\label{def:solvable}
Let $\Pi^{*}_{m}$ be the set of all optimal policies for the Mon-MDP $m$, where optimality is defined as in Eq.~\eqref{eq:max_pi}. 
A Mon-MDP $m$ is solvable if and only if 
\begin{equation}
    {\textstyle \bigcap\limits_{m' \in \{m\}_\indisting}} \Pi^{*}_{m'} \neq \varnothing, \label{eq:solvable}
\end{equation}
i.e., there exists at least one policy that is optimal for all Mon-MDPs $m'$ that are indistinguishable from $m$.
\end{Definition}

\begin{Lemma} 
A Mon-MDP $m$ is unsolvable if and only if there exists at least one Mon-MDP $m'$ that is indistinguishable from $m$, and there exists no policy that is optimal for both $m$ and $m'$, i.e.,  
\begin{equation}
    \exists {m' \in \{m\}_\indisting} \, : \Pi^{*}_{m} \cap \Pi^{*}_{m'} = \varnothing. \label{eq:unsolvable}
\end{equation}
\end{Lemma}

No algorithm can converge to an optimal policy for \textit{all} unsolvable Mon-MDPs.\footnote{For every Mon-MDP, there exists a trivial algorithm that converges to its optimal policy.  Consider an algorithm that starts from some policy and never updates it.  The algorithm that happened to start from an optimal policy for the Mon-MDP in question would then converge to an optimal policy, although clearly it only does so for that Mon-MDP (and ones that share that optimal policy).  
Instead, we are interested in algorithms that converge to optimal policies over a larger subset of Mon-MDPs, possibly with different optimal policies, such as the set of all Mon-MDPs that satisfy the conditions of Proposition~\ref{prop:convergence}.  
Since the indistinguishable equivalence class of Mon-MDPs is, by definition, those that are indistinguishable, it is sensible to consider only algorithms that converge over unions of these equivalence classes.}
In fact, no algorithm can converge to an optimal policy for even all the Mon-MDPs within the same equivalence class $\{m\}_\indisting$ of an unsolvable Mon-MDP $m$.  
If the algorithm converges to an optimal policy for one Mon-MDP in this class, it necessarily must converge to the same policy for all other indistinguishable Mon-MDPs in that class, and for one of these that policy is not optimal.

\subsection{Learning in Unsolvable Mon-MDP}
Not all {unsolvable} Mon-MDPs are equally unsolvable. There are Mon-MDPs that we define \textit{hopeless} where there is nothing the agent can learn about the environment.\footnote{In partial monitoring, ``hopeless'' games are games where the player cannot learning anything about the game's payoff~\citep{bartok2014partial, lattimore2018cleaning}.} 

\begin{Definition}[Hopeless Mon-MDP]
\label{def:hopeless}
An unsolvable Mon-MDP $m$ is hopeless if and only if
\begin{equation}
    {\textstyle \{m\}_\indisting} = \monspace_{\indistingspace}, \label{eq:hopeless}
\end{equation}
i.e., $m$ is indistinguishable from any other Mon-MDP with the same joint state-action space, transition function, and monitor rewards.
\end{Definition}

For example, in the Mon-MDP $m$ in Figure~\ref{fig:hopeless}, the monitor always returns unobservable proxy rewards $\rprox_t = \rewardundefined$ in all states. 
Intuitively, any trajectory can belong to any Mon-MDP with the same $\langle \statespace, \actionspace, \probmodel, \monrewardmodel \rangle$ because all that can be seen is $\rprox_t = \rewardundefined$ (and $\rprox_t$ is what makes trajectories distinguishable).

This leaves us some Mon-MDPs that live in-between solvable and hopeless. 
For such \textit{non-hopeless} Mon-MDPs, (1) there is not one policy that is optimal for all Mon-MDPs that are indistinguishable from $m$, and (2) some policies are suboptimal for all the Mon-MDPs indistinguishable from $m$. 
For example, in Figure~\ref{fig:non-hopeless}, a Mon-MDP where the environment reward (that is unobservable) in the \texttt{TOP-MIDDLE} state is $\renv_t = +1$ is indistinguishable from one where $\renv_t = -1$. In the former, an optimal policy walks over the \texttt{TOP-MIDDLE} state, but in the latter it avoids it. Therefore, one policy is not optimal for both Mon-MDPs. 
However, the rewards of all other states are observable and the agent can learn something. For example, collecting only 0 rewards is clearly suboptimal compared to walking only on 0s and then collecting the +1 reward. 

Following ideas from risk-averse RL, cautious RL, and partial monitoring worst-case assumptions,
we suggest the following objective is an appropriate starting point 
to tackle
non-hopeless Mon-MDPs
\begin{equation}    
\pi^*_m = \arg\max_{\pi \in \Pi_{\indistingspace}} \min_{m' \in \{m\}_\indisting} \mathbb{E}\Bigl[{\sum\nolimits_{t=1}^\horizon \upgamma^{t-1} r_t \;\Big|\; \pi, m'}\Bigr].
\end{equation}
That is, the agent should maximize the policy for the worst-case Mon-MDP among the indistinguishable ones it could be in.
For example, in Figure~\ref{fig:non-hopeless}, the agent should assume that unobservable rewards are negative and therefore walk to the goal while avoiding them.
Even though this may not be an optimal policy --- maybe unobservable rewards are positive --- it is still a reasonable policy (receiving +1 is better than not moving and receiving only 0s).

\subsection{Invariant Mon-MDPs}
We have delineated different types of unsolvable Mon-MDPs.  We can also delineate different types of solvable Mon-MDPs.  Here, we explore the situation where the monitor, while affecting the observability of rewards, does not change the optimal actions within the environment MDP.

\begin{Definition}[Invariant Mon-MDP]
\label{def:invariant}
A solvable Mon-MDP is invariant if and only if
\begin{equation}
\forall \senv, \smon, \aenv : \exists \pi^*, \pienvstar \, \EVV{}{\pi^*(\aenv, \amon \mid \senv, \smon) \mid \amon} = {\pienvstar(\aenv \mid \senv)}, \label{eq:invariant}
\end{equation}
where $\pienvstar$ is an optimal policy in $\langle \envstatespace, \allowbreak \envactionspace, \allowbreak \envprobmodel, \allowbreak \envrewardmodel, \upgamma \rangle$, i.e.,
\begin{align}
\pienvstar & \coloneqq \arg\max_{\pienv} \: \qenvpi(\senv,\aenv), \label{eq:max_pi_env}
\\
\textrm{where} \:\:
\qenvpi(\senv_t,\aenv_t) & \coloneqq \mathbb{E}\Bigl[\sum\nolimits_{i=t}^\horizon \upgamma^{i-t} \renv_{i} \;\Big|\; \pienv, \envprobmodel, \senv_t, \aenv_t\Bigr].  \label{eq:qenv}
\end{align}
\end{Definition}
That is, there exists an optimal policy $\pi^*$ that is also optimal with respect to the environment MDP. And, there exist monitor actions that, when paired with optimal environment actions, are optimal in the Mon-MDP. 
With this definition, ideas of reward shaping and potential functions~\citep{ng1999policy} are special cases of \textit{invariant} Mon-MDPs, where $\monstatespace$ and $\monactionspace$ are singleton sets, and $\monitormodel$ returns a modified reward without altering optimal policy optimality. 

A subclass of invariant Mon-MDPs are \emph{trivial} Mon-MDPs, where the monitor has no impact, i.e., $\forall t\; \rprox_t = \renv_t \mbox{~and~} \rmon_t = 0$.  This is just the setting where Mon-MDPs reduce to MDPs.\footnote{In partial monitoring, ``trivial'' games are games with zero worst-case regret~\citep{bartok2014partial, lattimore2018cleaning}.}

The taxonomy of Mon-MDPs discussed throughout this section is summarized in Figure~\ref{fig:mmdp_taxonomy}.


\clearpage

\onecolumn

\section{Algorithms Details}
\label{app:sec:alg_details}
In this section, we discuss in detail the update equations and the greedy policies of the algorithms presented in Section~\ref{subsec:algos}, and review convergence for the Sequential and Joint Q-Learning variants.

\subsection{Update Equations and Greedy Policies}
\label{app:subsec:q_and_pi}

If we consider fully-observable Mon-MDPs, i.e., $\rprox_t = \renv_t \; \forall t$, then Q-Learning updates are
\begin{equation}
    Q(s_t, a_t) \leftarrow (1 - \upalpha_t) Q(s_t, a_t) + \upalpha_t (r_t + \upgamma \max_a Q(s_{t+1}, a)), \label{eq:q_oracle} 
\end{equation}
where $s_t \coloneqq (\senv_t, \smon_t)$, $a_t \coloneqq (\aenv_t, \amon_t)$, and $r_t \coloneqq \renv_t + \rmon_t$. Note that we wrote $Q$ in place of $Q^\pi$ for the sake of simplicity, and we will keep this notation for the remainder of this section.
When the agent explores the environment, it follows an $\varepsilon$-greedy policy, i.e.,
\begin{equation}
    a_t = \begin{cases} 
        \arg\max_a Q(s_t,a) & \text{with probability $1 - \varepsilon_t$} 
        \\ 
        \text{uniform random in $\joinactionspace$} & \text{with probability $\varepsilon_t$}
    \end{cases}
    \label{eq:egreedy}
\end{equation}

This is what we call the ``\textcolor{oracle}{Oracle}'' baseline. 
With proxy rewards, the agent receives instead $\hat r_t \coloneqq (\rprox_t, \rmon_t)$. 
Below, we write the update equation and the greedy policy of the variants discussed in Section~\ref{subsec:algos}, where each variant treats $\rprox_t = \rewardundefined$ differently. We highlight each variant with the colors used later in Table~\ref{tab:rate} and Figure~\ref{fig:plots}.

\begin{itemize}[leftmargin=1em, itemsep=1pt, topsep=3pt]
    \item \textbf{\textcolor{zero}{$\boldsymbol{\rewardundefined = 0}$}.} Eq.~\eqref{eq:q_oracle} and~\eqref{eq:egreedy} are the same, but the agent replaces $\rprox_t = \rewardundefined$ with $\rprox_t = 0$. In Section~\ref{app:subsec:ablation_zero}, we test different values assigned to $\rewardundefined$.
    \item \textbf{\textcolor{ignore}{Ignore $\boldsymbol{\rprox_t = \rewardundefined}$}.} Eq.~\eqref{eq:q_oracle} and~\eqref{eq:egreedy} are the same, but the agent does not update the Q-function when $\rprox_t = \rewardundefined$.
    \item \textbf{\textcolor{joint}{Joint}.} The agent learns two Q-functions, $\qenv$ with only proxy rewards (updated only when $\rprox_t \neq \rewardundefined$) and $\qmon$ with only monitor rewards. The $\texttt{max}$ greedy operator selects $(\aenv, \amon)$ together. 
    \begin{align}
        \qenv(\senv_t, \aenv_t) \leftarrow  & (1 - \upalpha_t) \qenv(\senv_t, \aenv_t) + \upalpha_t (\rprox_t + \upgamma \max_{\aenv} \qenv(\senv_{t+1}, \aenv)) \label{eq:qe_joint}
        \\
        \qmon(\senv_t, \aenv_t, \smon_t, \amon_t) \leftarrow & (1 - \upalpha_t) \qmon(\senv_t, \aenv_t, \smon_t, \amon_t) + \upalpha_t (\rmon_t + \upgamma \textcolor{joint}{\boldsymbol{\max_{\aenv, \amon}}} \qmon(\senv_{t+1}, \textcolor{joint}{\boldsymbol{\aenv}}, \smon_{t+1}, \textcolor{joint}{\boldsymbol{\amon}})) \label{eq:qm_joint}
        \\
        \text{Greedy policy:} \;\; & (\aenv_t, \amon_t) = \arg\max_{\aenv, \amon} \{\qenv (\senv_t, \aenv) + \qmon(\senv_t, \aenv, \smon_t, \amon)\} \label{eq:pi_joint}
    \end{align}
    Note that $\qmon$ considers only monitor rewards $\rmon$ but depends also on the environment state and action $(\senv, \aenv)$. This is because the next monitor state $\smon_{t+1}$ depends also on $(\senv_t, \aenv_t)$, as defined in the transition function $\monprobmodel$.
    \item \textbf{\textcolor{sequential}{Sequential}.} The agent learns two separate Q-functions, but the $\texttt{max}$ greedy operator selects first $\aenv$ and then $\amon$.
    \begin{align}
        \qenv(\senv_t, \aenv_t) \leftarrow & (1 - \upalpha_t) \qenv(\senv_t, \aenv_t) + \upalpha_t (\rprox_t + \upgamma \max_{\aenv} \qenv(\senv_{t+1}, \aenv)) \label{eq:qe_seq} 
        \\
        \qmon(\senv_t, \aenv_t, \smon_t, \amon_t) \leftarrow & (1 - \upalpha_t) \qmon(\senv_t, \aenv_t, \smon_t, \amon_t) + \upalpha_t (\rmon_t + \upgamma \textcolor{sequential}{\boldsymbol{\max_{\amon}}} \qmon(\senv_{t+1}, \textcolor{sequential}{\boldsymbol{\arg\max_{\aenv} \qenv(\senv_{t+1}, \aenv)}}, \smon_{t+1}, \textcolor{sequential}{\boldsymbol{\amon}})) \label{eq:qm_seq}
        \\
        \text{Greedy policy:} \;\; & \text{first} \; \aenv_t = \arg\max_{\aenv} \qenv(\senv_t, \aenv), \;\; \text{then} \; \amon_t = \arg\max_{\amon} \qmon(\senv_t, \aenv_t, \smon_t, \amon) \label{eq:pi_seq}
    \end{align}
    \item \textbf{\textcolor{model}{Reward Model}.} The agent learns a model of the environment reward $\smash{\widehat{R}(\senv, \aenv})$. 
    Eq.~\eqref{eq:q_oracle} and~\eqref{eq:egreedy} are the same, but $\rprox_t$ is replaced with the $\smash{\widehat{R}(\senv_t, \aenv_t})$. The prediction model for our experiments is a table like the Q-function that keeps the running mean of the environment rewards as it observes $\rprox_t = \renv_t$. 
\end{itemize}

\subsection{\textbf{Sequential Algorithm Convergence in Invariant Mon-MDPs}}
\label{app:subsec:proof_seq}
In Section~\ref{subsec:proof}, we proved that the \textcolor{model}{Reward Model} algorithm is guaranteed to converge to an optimal policy of any Mon-MDP under Proposition~\ref{prop:convergence}. Here, we prove that the \textcolor{sequential}{Sequential} algorithm is guaranteed to converge to an optimal policy under the same conditions of Proposition~\ref{prop:convergence} but only in invariant Mon-MDPs.

\begin{Proposition}[Sequential Algorithm Convergence in Invariant Mon-MDPs]
\label{prop:convergence_seq}
For any invariant Mon-MDP with finite action and state spaces that satisfies Properties~\ref{def:ergodic_mmdp},~\ref{def:env_ergodic_mon}, and~\ref{def:truthful_mon}, the Sequential algorithm will converge to an optimal policy $\pi^*$ defined in Eq.~\eqref{eq:max_pi} of that Mon-MDP, under a GLIE policy and if $\upalpha_t$ satisfies the Robbins-Monro conditions. 
\end{Proposition}

\begin{Proof}
First, by linearity of expectation we rewrite the Mon-MDP Q-function in Eq.~\eqref{eq:q} as the sum of two separate Q-functions, i.e., 
\begin{align}
    Q^{\pi}(\senv_t,\aenv_t, \smon_t, \amon_t) \coloneqq & \, \mathbb{E}\Bigl[\sum\nolimits_{i=t}^\horizon \upgamma^{i-t} (\renv_{i} \, + \, \rmon_t) \;\Big|\; \pienv, \pimon, \envprobmodel, \monprobmodel, \senv_t, \smon_t, \aenv_t, \amon_t \Bigr] \nonumber
    \\
    = & \, {\mathbb{E}\Bigl[\sum\nolimits_{i=t}^\horizon \upgamma^{i-t} \rmon_t \;\Big|\; \pienv, \pimon, \envprobmodel, \monprobmodel, \senv_t, \smon_t, \aenv_t, \amon_t \Bigr]} + {\mathbb{E}\Bigl[\sum\nolimits_{i=t}^\horizon \upgamma^{i-t} \renv_{i} \;\Big|\; \pienv, \envprobmodel, \senv_t, \aenv_t \Bigr]} \nonumber
    \\
    = & \, \qmonenvpi(\senv_t, \aenv_t, \smon_t, \amon_t) + \qenvpi(\senv_t, \aenv_t). \label{eq:split_q}
\end{align}
We split the expectation because environment rewards do not depend on monitor states and actions (monitor rewards depend on both monitor and environment states and actions because of the monitor transition function). 
In the above Equation, $\qenvpi$ is $\qenv$ of Eq.~\eqref{eq:qe_seq}, while $\qmonenvpi$ is $\qmon$ of Eq.~\eqref{eq:qm_seq}. 
The double superscript ``$\pimon\pienv$'' denotes that the Q-function depends on two policies: $\pienv$ selects environment actions greedily first, and then $\pimon$ selects monitor actions greedily. 

Because both Q-functions depend on environment states and actions, normally we cannot maximize them independently, i.e., 
\begin{equation*}
    \max_{\pienv, \pimon}\left(\qmonenvpi(\senv_t, \aenv, \smon_t, \amon) + \qenvpi(\senv_t, \aenv)\right) \neq \max_{\pienv, \pimon}\qmonenvpi(\senv_t, \aenv, \smon_t, \amon) + \max_{\pienv}\qenvpi(\senv_t, \aenv).
\end{equation*}
However, in invariant Mon-MDP we can. First, we note that under the conditions of Proposition~\ref{prop:convergence_seq} $\qenvpi$ will converge to $\qenvstar$, i.e., the Q-function of $\pienvstar$ of Eq.~\eqref{eq:max_pi_env}. 
Second, by the definition of invariant Mon-MDPs, there exist monitor actions that, when paired with optimal greedy environment actions, are optimal as well. That is, we can replace $\pienv$ with $\pienvstar$ from Eq.~\eqref{eq:max_pi_env} and preserve optimality. This leads to
\begin{align}
    \max_{\textcolor{sequential}{\pienv}, \pimon}\left(\qmonenvpiseq(\senv_t, \textcolor{sequential}{\aenv}, \smon_t, \amon) + \qenvpiseq(\senv_t, \textcolor{sequential}{\aenv})\right) & = \underbrace{\max_{\pimon}\qmonenvpistarseq(\senv_t, \textcolor{sequential}{\aenv_t}, \smon_t, \amon)}_{\substack{\text{$\max$ is over $\amon$ only} \\ \text{because $\aenv_t$ is already greedy}}} + \,\qenvstarseq(\senv_t, \textcolor{sequential}{\aenv_t}) \qquad \text{where} \quad \aenv_t = \arg\max_{\aenv}\qenvstar. \label{eq:sequential_max_equality}
\end{align}
Because $\pienvstar$ is already optimal, Eq.~\eqref{eq:sequential_max_equality} can be solved with classic Q-Learning over monitor rewards, and the agent will learn the policy $\pimonstar$ that maximizes $\qmonenvpistar$ (under the conditions of Proposition~\ref{prop:convergence_seq}: Mon-MDP ergodicity, GLIE policy, and learning rate satisfying the Robbins-Monro conditions).  
\end{Proof}

One may think that selecting greedy monitor actions first and then greedy environment actions would also be optimal. 
However, $\qmon$ depends on $(\senv, \aenv)$, therefore $\arg\max_{\amon}\qmon$ must account for environment actions. Either $\aenv$ is selected first (as in the \textcolor{sequential}{Sequential} algorithm), or at the same time (as in the \textcolor{joint}{Joint} algorithm).

\subsection{\textbf{Joint Algorithm Counterexample}}
\label{app:subsec:joint_counter}
\begin{figure}[h]
    \begin{minipage}[b]{0.66\textwidth}
        \begin{subfigure}[b]{0.48\columnwidth}
            \centering
            \includegraphics[width=\linewidth]{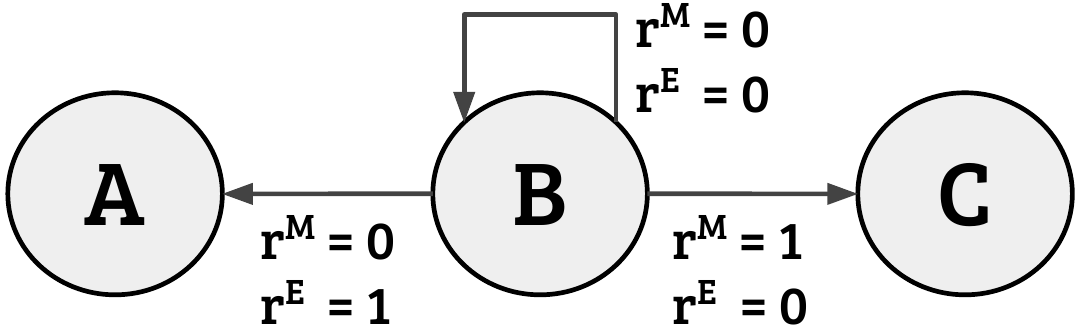}
            \caption{\label{fig:chain_r}Rewards}
        \end{subfigure}
        \hfill
        \begin{subfigure}[b]{0.48\columnwidth}
            \centering
            \includegraphics[width=\linewidth]{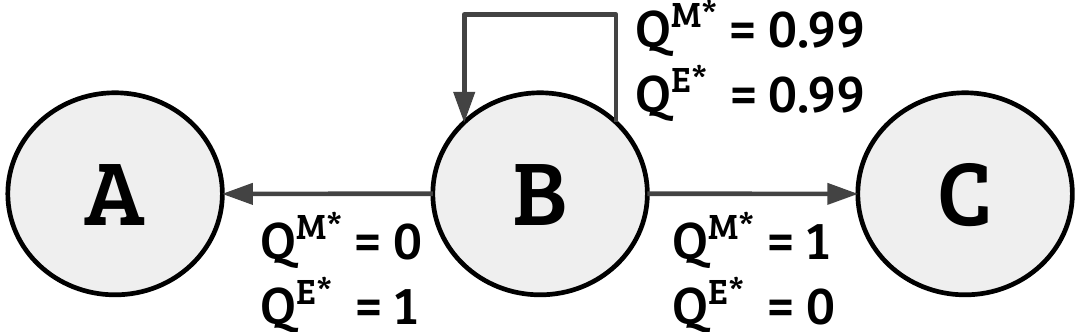}
            \caption{\label{fig:chain_q}Optimal Q-values}
        \end{subfigure}
    \end{minipage}
    \hfill
    \begin{minipage}[b]{0.32\textwidth}
        \centering
        \caption{\label{fig:joint_counterexample}\textbf{Invariant Mon-MDP where the \textcolor{joint}{Joint} algorithm will converge to a suboptimal policy. Optimal Q-values point the agent to either \texttt{A} or \texttt{C}, but the greedy policy sums them as in Eq.~\eqref{eq:pi_joint} and decides to stay in \texttt{B}.}}
    \end{minipage}
\end{figure}

Compared to Eq.~\eqref{eq:split_q}, the \textcolor{joint}{Joint} algorithm changes the optimization problem to 
\begin{equation*}
    \max_{\textcolor{joint}{\pi}, \pienv}\left(Q^{\textcolor{joint}{\pi}}(\senv_t, \aenv, \smon_t, \amon) + \qenvpi(\senv_t, \aenv)\right), 
\end{equation*}
i.e., the Q-function of monitor rewards is under a different policy than $\pienv$ --- the policy $\pi$ selects $(\aenv_t, \amon_t)$ together rather than sequentially. 
Therefore, we cannot replace $\pienv$ with $\pienvstar$ and prove its convergence to an optimal policy as we did for the \textcolor{sequential}{Sequential}. Actually, the example in Figure~\ref{fig:joint_counterexample} shows an invariant Mon-MDP where the \textcolor{joint}{Joint} algorithm does \textit{not} converge to an optimal policy. 
In this chainworld, the agent starts in state \texttt{B} and can either go to \texttt{A}, \texttt{C}, or stay in \texttt{B}. Moving to \texttt{A} or \texttt{C} gives positive rewards and ends the episode. The monitor state and action spaces are singleton sets and do not influence the observability of the rewards. 
There are infinitely many optimal policies --- any policy mixing going to \texttt{A} or \texttt{C} with any probability is optimal. The policy that always goes to \texttt{A} is also environment-optimal, therefore the Mon-MDP is invariant. 
Following Eq.~\eqref{eq:qe_joint} and~\eqref{eq:qm_joint} with $\upgamma = 0.99$, the \textcolor{joint}{Joint} algorithm converges to the Q-values shown in Figure~\ref{fig:chain_q}. However, the greedy policy in Eq.~\eqref{eq:pi_joint} prefers to stay in \texttt{B}, as the sum of the Q-values is larger than single Q-values. 
This is the same behavior displayed by the \textcolor{joint}{Joint} policy in the Button Mon-MDP discussed in Section~\ref{subsec:algos}.
Again, this is not surprising, considering that the $\texttt{max}$ operator is not linear, i.e., given two functions $f(x)$ and $g(x)$, $\max_x (f(x) + g(x)) \neq \max_x f(x) + \max_x g(x)$. 

\clearpage

\section{Experiments}
\label{app:sec:exp}

\subsection{Mon-MDP Characteristics}
\label{app:subsec:mon_details}
Together with the Mon-MDPs presented in Section~\ref{sec:experiments}, we present three new Mon-MDPs that mimic simple real-world situations. They are harder --- larger spaces and richer dynamics --- and provide further insights about the challenges arising in Mon-MDPs.

\begingroup 
\addtolength\jot{-2pt}

\begin{adjustwidth}{-0.7cm}{-0.7cm}
\setlength{\columnsep}{1em}
\begin{multicols}{2}
\setlength{\columnsep}{1em}

\begin{itemize}[leftmargin=1em, itemsep=1pt, topsep=3pt]
\item \textbf{Simple, Penalty.} The monitor is always \texttt{OFF}. The agent can observe the current environment reward with an explicit monitor action. 
\begin{align*}
\monstatespace & \coloneqq \{\texttt{OFF}\} \qquad
\monactionspace \coloneqq \{\texttt{ASK, NO-OP}\} \qquad
\smon_{t+1} = \texttt{OFF}, \, t \geq 0
\\
\rprox_t & = \begin{cases} 
    \renv_t & \text{if $\amon_t = \texttt{ASK}$}
    \\
    \rewardundefined & \text{otherwise} 
\end{cases} \qquad
\rmon_t = \begin{cases} 
    -0.2 & \text{if $\amon_t = \texttt{ASK}$} 
    \\ 
    0 & \text{otherwise} 
\end{cases}
\end{align*}
\item \textbf{Button.} The monitor is turned \texttt{ON}/\texttt{OFF} by hitting a button on the bottom wall of the bottom-right cell. The agent can do that with $\aenv = \texttt{DOWN}$ in the $\senv = \texttt{BOTTOM-RIGHT}$. 
\begin{align*}
\monstatespace & \coloneqq \{\texttt{ON, OFF}\} \qquad
\monactionspace \coloneqq \{\texttt{NO-OP}\}
\\
\smon_1 &= \text{random uniform in $\monstatespace$}
\\
\smon_{t+1} & = \begin{cases} 
    \texttt{ON} & \text{if $\smon_t = \texttt{OFF}$ and $\senv_t = \texttt{BOTTOM-RIGHT}$ and $\aenv_t = \texttt{DOWN}$} 
    \\ 
    \texttt{OFF} & \text{if $\smon_t = \texttt{ON}$ and $\senv_t = \texttt{BOTTOM-RIGHT}$ and $\aenv_t = \texttt{DOWN}$} 
    \\ 
    \smon_t & \text{otherwise} 
\end{cases}
\\
\rprox_t &= \begin{cases} 
    \renv_t & \text{if $\smon_{t+1} = \texttt{ON}$} 
    \\ 
    \rewardundefined & \text{otherwise} 
\end{cases} \qquad
\rmon_t = \begin{cases} 
    -0.2 & \text{if $\smon_{t+1} = \texttt{ON}$} 
    \\ 
    0 & \text{otherwise} 
\end{cases}
\end{align*}
Note that there are still nine environment states, but the \texttt{BOTTOM-RIGHT} cell affects the dynamics of the monitor.
Also, this Mon-MDP is not invariant --- an optimal environment policy would never bump into the wall to turn monitoring \texttt{ON/OFF}. 
\item \textbf{N-Monitor.} At every step, a random monitor is \texttt{ON}. If the agent asks to be monitored to one that is \texttt{ON}, it observes $\rprox_t = \renv_t$ but receives a negative monitor reward. 
Otherwise, it receives $\rprox_t = \rewardundefined$ but receives a smaller positive monitor reward.
\begin{align*}
    \monstatespace & \coloneqq \{\texttt{ON\textsubscript{1}}, \, \ldots, \, \texttt{ON\textsubscript{N}}\} \\
    \monactionspace & \coloneqq \{\texttt{ASK\textsubscript{1}}, \, \ldots, \, \texttt{ASK\textsubscript{N}}\} \\
    \smon_{t+1} &= \text{random uniform in $\monstatespace$}, \, t \geq 0 \\
    \rprox_t & = \begin{cases} 
        \renv_t & \text{if $\amon_t = \texttt{ASK\textsubscript{i}}$ and $\smon_t = \texttt{ON\textsubscript{i}}$ with $\texttt{i} = \texttt{1} \ldots \texttt{N}$} 
        \\
        \rewardundefined & \text{otherwise} 
    \end{cases} \\
    \rmon_t & = \begin{cases} 
        -0.2 & \text{if $\amon_t = \texttt{ASK\textsubscript{i}}$ and $\smon_t = \texttt{ON\textsubscript{i}}$ with $\texttt{i} = \texttt{1} \ldots \texttt{N}$} 
        \\
        0.001 & \text{otherwise} 
    \end{cases}
\end{align*}
In our experiments, $N = 5$. 
An optimal policy goes to the goal while asking \texttt{OFF} monitors, in order to receive the small positive monitor rewards along the way. 
The dynamics of this Mon-MDP are similar to the Simple / Penalty Mon-MDPs, i.e., the agent immediately observes the reward if it correctly asks for it. What changes is the dimensionality of the state-action space and the presence of positive monitor rewards. 
In particular, the latter discourage the agent from asking the correct monitor and observing environment rewards. Most of the time, all the agent sees is either $(\rprox_t = 0, \rmon_t = -0.2)$ for asking the right monitor, or $(\rprox = \rewardundefined, \rmon_t = 0.001)$ for asking the wrong one. Until the agent observes $(\rprox_t = 1, \rmon_t = -0.2)$ for reaching the goal while monitored, its greedy policy prefers to keep collecting small $\rmon_t = 0.001$ rather than trying to observe new rewards. As we will show in Section~\ref{app:sec:exp}, this has a severe impact on the rate of convergence with an $\varepsilon$-greedy policy. 
\item \textbf{Limited-Time.} The monitor is \texttt{ON} at the beginning of the episode, and has a small chance to turn \texttt{OFF} at any time step.\footnote{Because of the Markovian property, $\monprobmodel$ cannot depend on time or past data.
Having a probability $p$ that the monitor can turn \texttt{OFF} at every step mimics a time-dependent dynamics where the monitor stays \texttt{ON} for $1 /p$ steps on average, while keeping $\monprobmodel$ Markovian.} 
Once \texttt{OFF}, the monitor stays \texttt{OFF}. 
\begin{align*}
\monstatespace & \coloneqq \{\texttt{ON, OFF}\} \qquad
\monactionspace \coloneqq \{\texttt{NO-OP}\} 
\\
\smon_{1} & = \texttt{ON} \qquad
\smon_{t+1} = \begin{cases} 
    \texttt{ON} & \text{with probability $p$ if $\smon_t = \texttt{ON}$} 
    \\
    \texttt{OFF} & \text{otherwise} 
\end{cases}
\\
\rprox_t & = \begin{cases} 
    \renv_t & \text{if $\smon_t = \texttt{ON}$} 
    \\
    \rewardundefined & \text{otherwise} 
\end{cases} \qquad
\rmon_t = 0 
\end{align*}
In our experiments, $p = 80\%$. In this Mon-MDP, the agent cannot change the observability of the rewards (there are no monitor actions). Therefore, efficiently explore to observe new environment rewards while monitoring is \texttt{ON} is crucial. When monitoring is \texttt{OFF}, all the agent can do is to end the episode and start over. 
\item \textbf{Limited-Use.} The monitor consumes a battery when \texttt{ON}. The battery life is part of the monitor state and when it is depleted the monitor stays \texttt{OFF}. The agent can turn \texttt{OFF} the monitor to save battery. Monitoring has not cost. However, if the agent fully depletes the battery, it will receive $\rmon_t = 1$ in terminal states.\footnote{To preserve the Markovian property, terminal states are only states that end the episode regardless of the time step (e.g., the goal state in our gridworld).}
\begin{align*}
    \monstatespace & \coloneqq \overbrace{\{\texttt{ON, OFF}\}}^{\smon_t(1)} \quad \times \overbrace{{\{\texttt{0}, \, \ldots, \, \texttt{N}\}}}^{\smon_t(2) \,\, \text{is the battery life}} \\
    \monactionspace & \coloneqq \{\texttt{TURN ON, TURN OFF, NO-OP}\} \\
    \smon_{1} & = (\texttt{OFF, N}) \\
    \smon_{t+1}(2) & = \begin{cases} 
        \max(\texttt{0}, \smon_t(2) - \texttt{1}) & \text{if $\smon_t(1) = \texttt{ON}$} 
        \\
        \smon_t(2) & \text{otherwise} 
    \end{cases} \\
    \smon_{t+1}(1) & = \begin{cases} 
        \texttt{ON} & \text{if $\smon_{t}(1) = \texttt{ON}$ and $\smon_{t}(2) > \texttt{0}$ and $\amon_{t} \neq \texttt{TURN OFF}$}
        \\
        \texttt{ON} & \text{if $\smon_{t}(2) > \texttt{0}$ and $\amon_{t} = \texttt{TURN ON}$}
        \\
        \texttt{OFF} & \text{if $\amon_t = \texttt{TURN OFF}$ or $\smon_t(2) = \texttt{0}$} 
    \end{cases} \\
    \rprox_t & = \begin{cases} 
        \renv_t & \text{if $\smon_t = \texttt{ON}$} 
        \\
        \rewardundefined & \text{otherwise} 
    \end{cases} \\
    \rmon_t & = \begin{cases} 
        1 & \text{if $\smon_t(2) = \texttt{0}$ and $\senv_t$ is terminal} 
        \\
        0 & \text{otherwise} 
    \end{cases}
\end{align*}
In our experiments, $N = 7$. An optimal policy turns the monitor \texttt{ON}, waits for the battery to deplete, and goes to the goal. 
Unlike N-Monitor, the positive monitor reward does not discourage the agent from observing environment rewards. However, it makes the Mon-MDP non-invariant --- the agent must not go to the goal as quickly as possible, but has to wait for the battery to deplete, e.g., by bumping into walls and not moving. 
\end{itemize}

\end{multicols}
\end{adjustwidth}

\endgroup

\begin{table}[h]
    \setlength\tabcolsep{2pt}
    \renewcommand{\arraystretch}{1.0}
    \centering
    \caption{\label{tab:monitors}\textbf{Summary of the Mon-MDPs used in our experiments. All satisfy the conditions for convergence to an optimal policy of Proposition~\ref{prop:convergence}, i.e., joint state-action ergodicity, environment reward ergodicity, and proxy reward truthfulness.}}
    \begin{tabular}{@{\extracolsep{\fill}}|l|c|c|c|c|c|}
        \hline
         & \textbf{Dimensionality} & \textbf{Explicit}  & & \textbf{Positive} &
        \\
        \textbf{Mon-MDP} & $\boldsymbol{||\,\monstatespace \times \monactionspace\,||}$ & \textbf{Mon. Actions}  & \textbf{Invariant} & \textbf{Mon. Rewards} & \textbf{Difficulty} 
        \\
        \hline
        \Tstrut
        Simple, Penalty & 2 & \good & \good & \bad & $\bullet$ 
        \\
        Button & 2 & \bad & \bad & \bad & $\bullet$$\bullet$ 
        \\
        N-Monitor & 25 & \good & \good & \good & $\bullet$$\bullet$$\bullet$$\bullet$ 
        \\
        Limited-Time & 2 & \bad & \good & \bad & $\bullet$$\bullet$$\bullet$$\bullet$ 
        \\
        Limited-Use & 36 & \good & \bad & \good & $\;\bullet$$\bullet$$\bullet$$\bullet$ 
        \Bstrut
        \\
        \hline
    \end{tabular}
\end{table}

\vspace*{-2pt}

Table~\ref{tab:monitors} highlights important features of the Mon-MDPs presented in this section, i.e., the size of their monitor spaces, if the agent has dedicated actions to explicitly control monitoring, if they are invariant, if monitor rewards are positive, and their overall level of difficulty. 
Below, we briefly summarize the main characteristics and challenges of these Mon-MDPs.

\begin{itemize}[leftmargin=1em, itemsep=1pt, topsep=3pt]
    \item \textbf{Simple, Penalty.} Simple binary monitor action.
    \item \textbf{Button.} No monitor action, the agent must use environment actions to activate/deactivate monitoring, non-invariant. 
    \item \textbf{N-Monitor.} Larger spaces, small positive monitor rewards that do not encourage the discovery of environment rewards. 
    \item \textbf{Limited-Time.} Random chance of not being able to observe the rewards at all, limited monitoring.
    \item \textbf{Limited-Use.} Larger spaces, limited monitoring, positive monitor reward for reaching the goal, non-invariant. 
\end{itemize}

\subsection{Empirical Rate of Convergence}
\label{app:subsec:rate}
This section expands the results of Section~\ref{subsec:rate}. Here, we include the three additional Mon-MDPs presented in the previous section --- N-Monitor, Limited-Time, and Limited-Use --- and show the performance of all algorithms.
For each Mon-MDP, we consider two settings, one with deterministic and one with noisy environment rewards. The latter adds Gaussian noise with standard deviation 0.05 to the environment reward. 
The hyperparameters of the experiment settings are the following. 

\begin{itemize}[leftmargin=1em, itemsep=1pt, topsep=3pt]
\item Q-values are initialized pessimistically to -10. In Section~\ref{app:subsec:ablation_q0} we investigate different initialization values.
\item Discount factor $\upgamma = 0.99$.
\item Constant learning rate $\upalpha_t = 1$.
\item The coefficient $\varepsilon_t$ of the $\varepsilon$-greedy exploration policy linearly decays from 1 (beginning of training) to 0 (end of training).
\item Training lasts for 10,000 steps in the deterministic setting, 100,000 in the noisy one. 
\item Every 10 training steps, we estimate the \textit{expected discounted return} $\smash{\mathbb{E}[{\sum_{t=1}^T\upgamma^{t-1}(\renv_t + \rmon_t) \mid \pi]}}$ of the current greedy policy (presented in Section~\ref{app:sec:alg_details}). In Mon-MDPs with noisy reward, we turn off the noise at evaluation to determine if a policy has converged (see below). 
\item We assume that a policy has converged if its expected discounted return does not change for the last 2,000 (deterministic setting) or 20,000 (noisy setting) training steps. For example, if a policy is optimal after 3,000 training steps and stays optimal until the end of training, then we say that ``the algorithm converged to an optimal policy after 3,000 steps.''
\item We average the results over 100 training seeds and report 95\% confidence intervals.
\end{itemize}

\begin{figure*}[ht]
    \setlength{\belowcaptionskip}{0pt}
    \setlength{\abovecaptionskip}{3pt}
    \centering
    \includegraphics[width=0.6\linewidth]{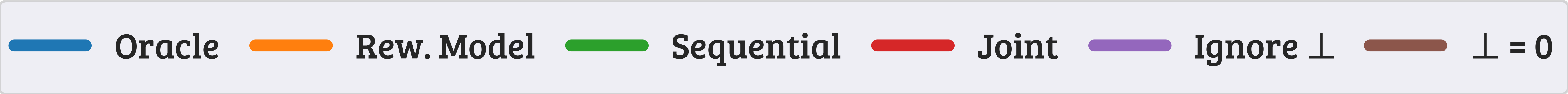}
    \\[3pt]
    \makebox[\linewidth][c]{%
    \raisebox{24pt}{\rotatebox[origin=t]{90}{\fontfamily{qbk}\footnotesize\textbf{Determ. Reward}}}
    \hfill
    \begin{subfigure}[b]{0.17\linewidth}
        \centering
        {\fontfamily{qbk}\footnotesize\textbf{Simple}}\\[3pt]
        \includegraphics[width=\linewidth]{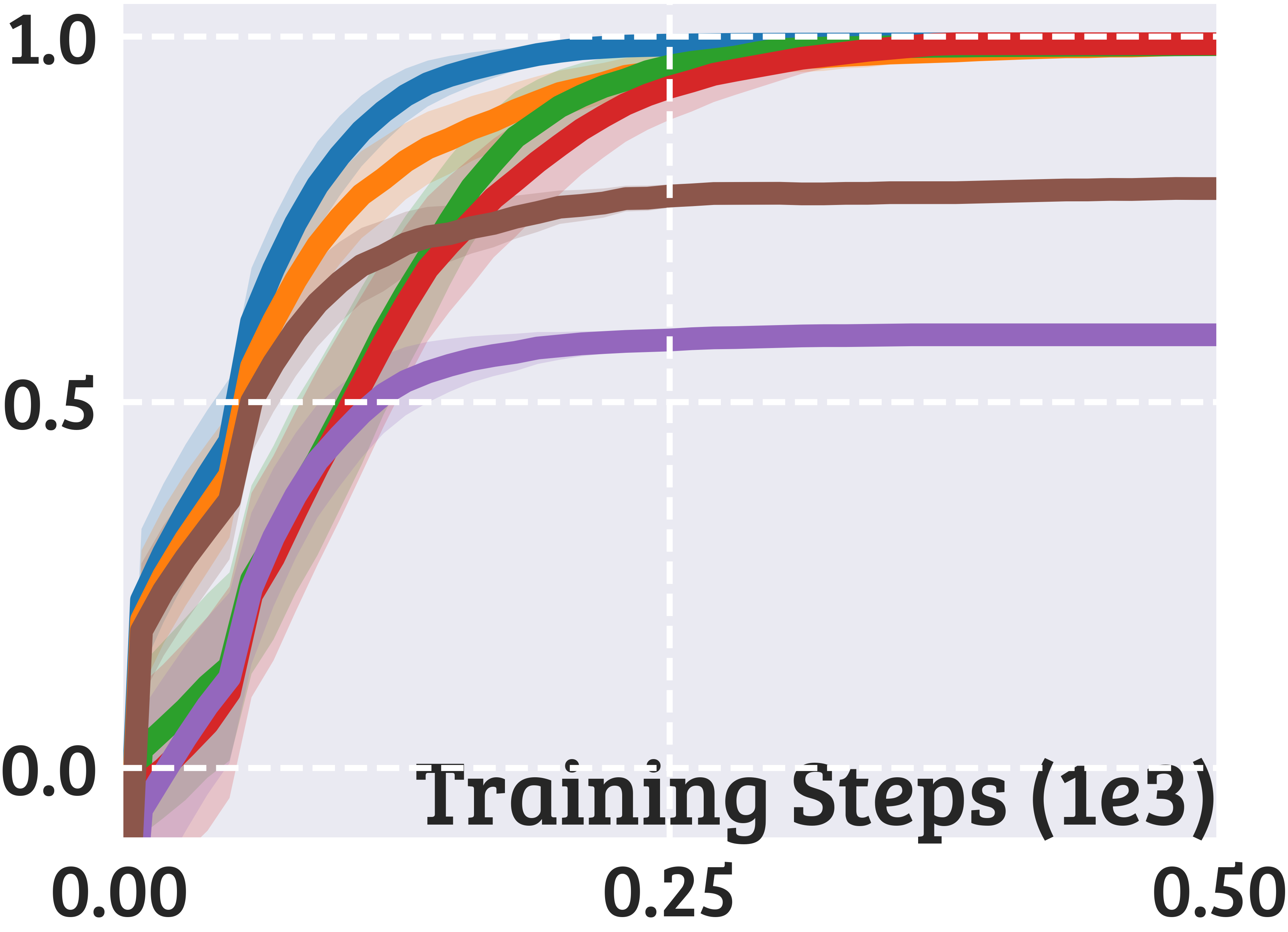}
    \end{subfigure}
    \hfill
    \begin{subfigure}[b]{0.17\linewidth}
        \centering
        {\fontfamily{qbk}\footnotesize\textbf{Penalty}}\\[3pt]
        \includegraphics[width=\linewidth]{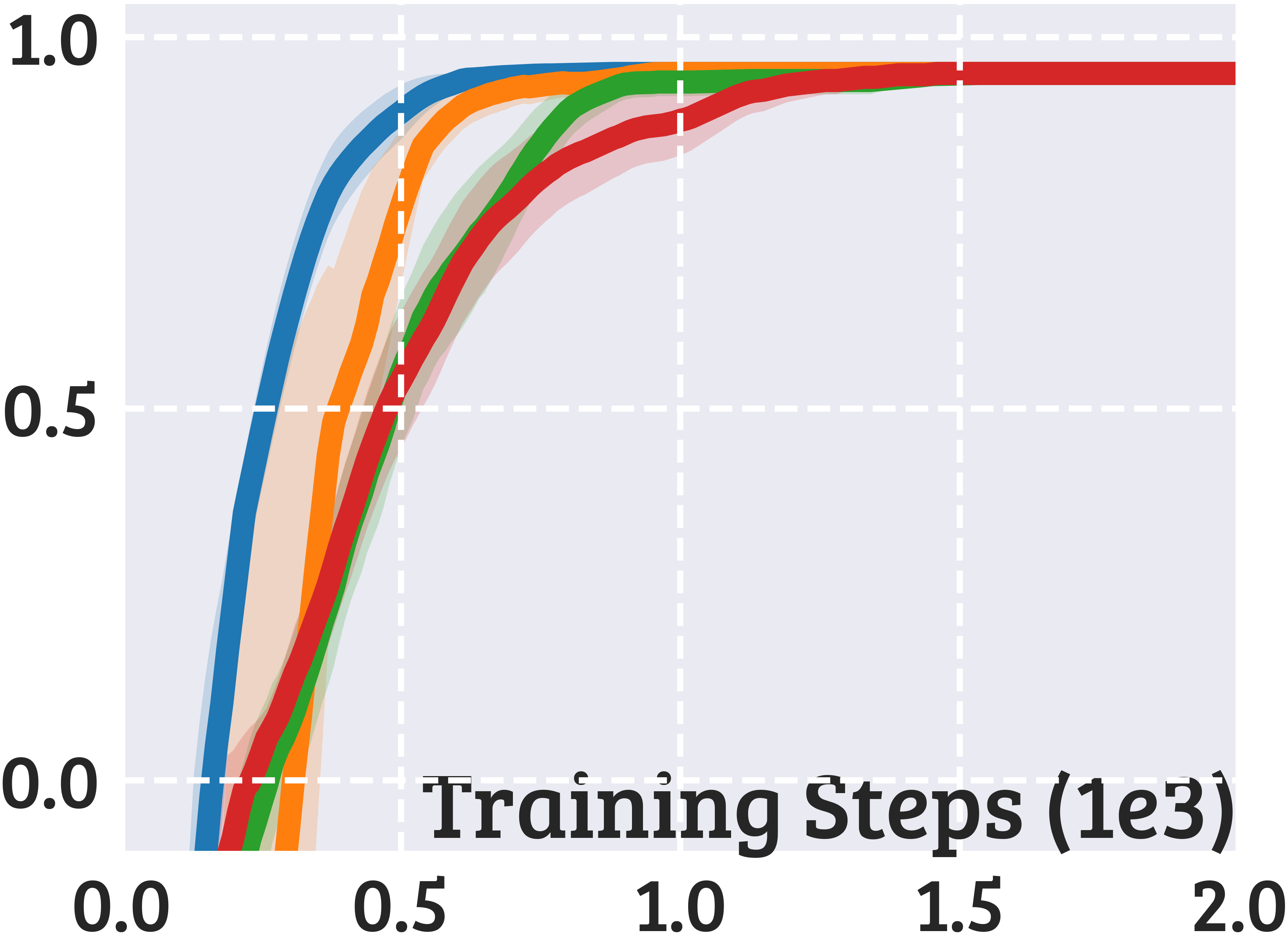}
    \end{subfigure}
    \hfill
    \begin{subfigure}[b]{0.17\linewidth}
        \centering
        {\fontfamily{qbk}\footnotesize\textbf{Button}}\\[3pt]
        \includegraphics[width=\linewidth]{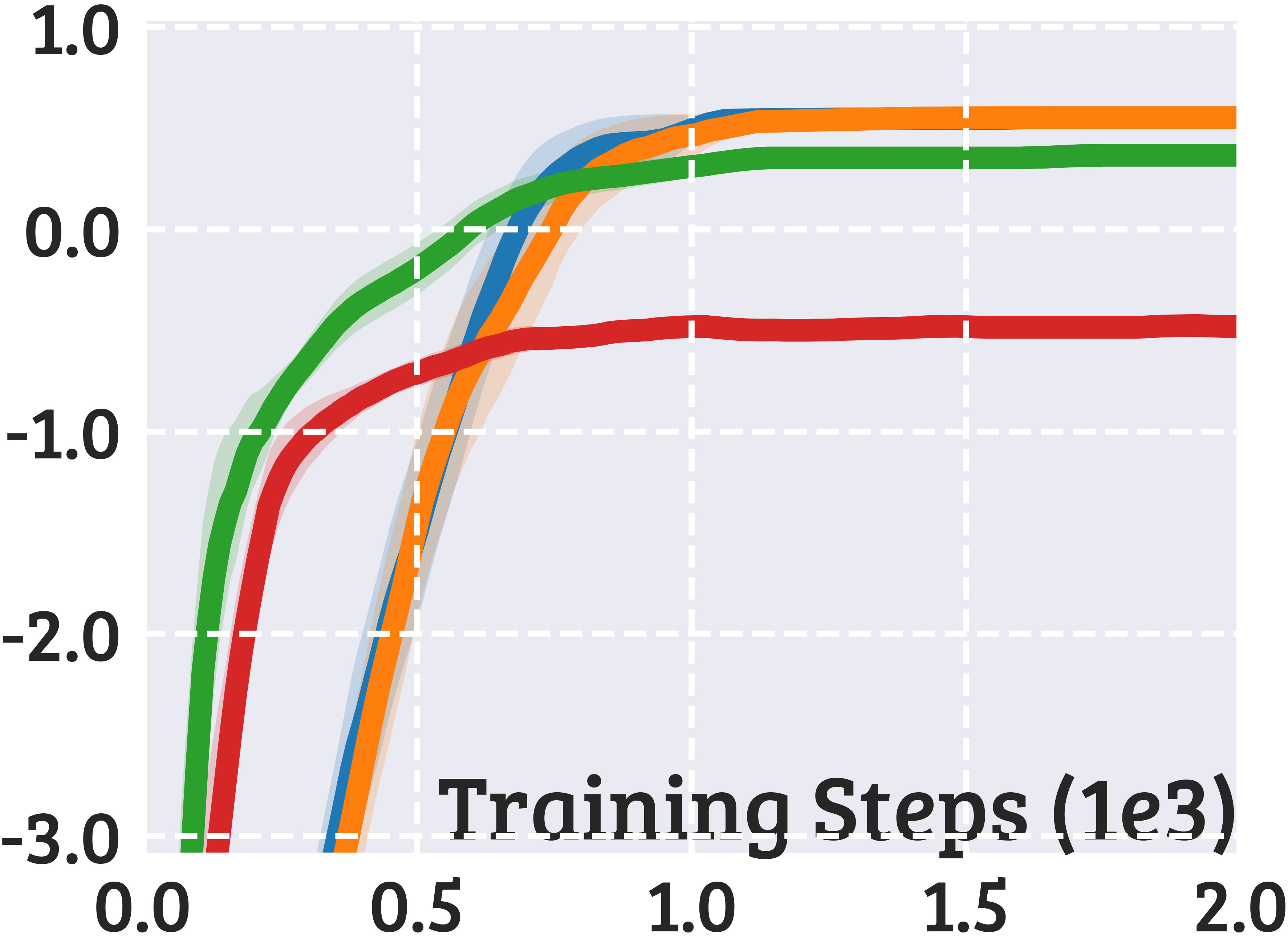}
    \end{subfigure}
    \hfill
    \begin{subfigure}[b]{0.17\linewidth}
        \centering
        {\fontfamily{qbk}\footnotesize\textbf{N-Monitor}}\\[3pt]
        \includegraphics[width=\linewidth]{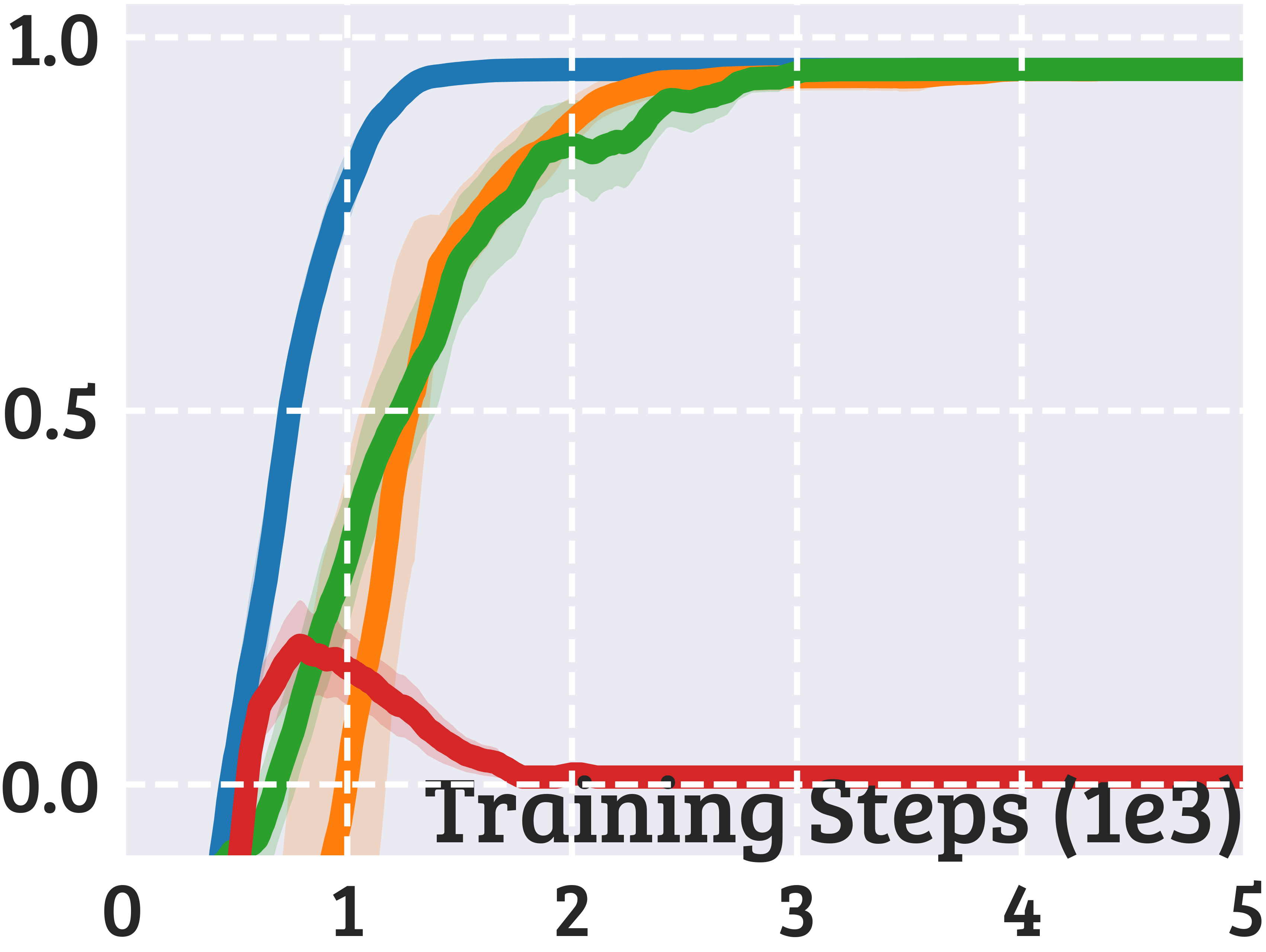}
    \end{subfigure}
    \hfill
    \begin{subfigure}[b]{0.17\linewidth}
        \centering
        {\fontfamily{qbk}\footnotesize\textbf{Limited-Time}}\\[3pt]
        \includegraphics[width=\linewidth]{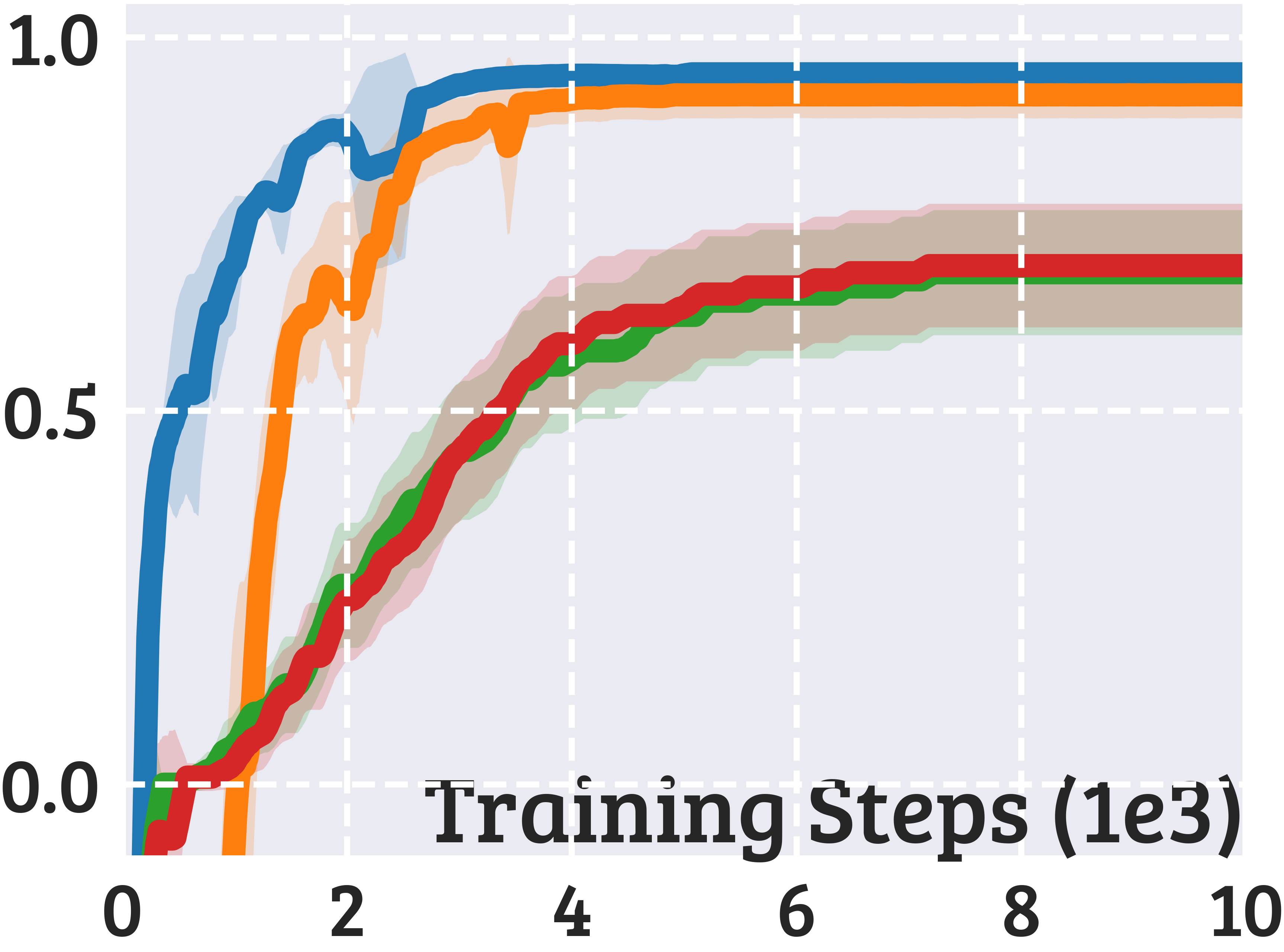}
    \end{subfigure}
    \hfill
    \begin{subfigure}[b]{0.17\linewidth}
        \centering
        {\fontfamily{qbk}\footnotesize\textbf{Limited-Use}}\\[3pt]
        \includegraphics[width=\linewidth]{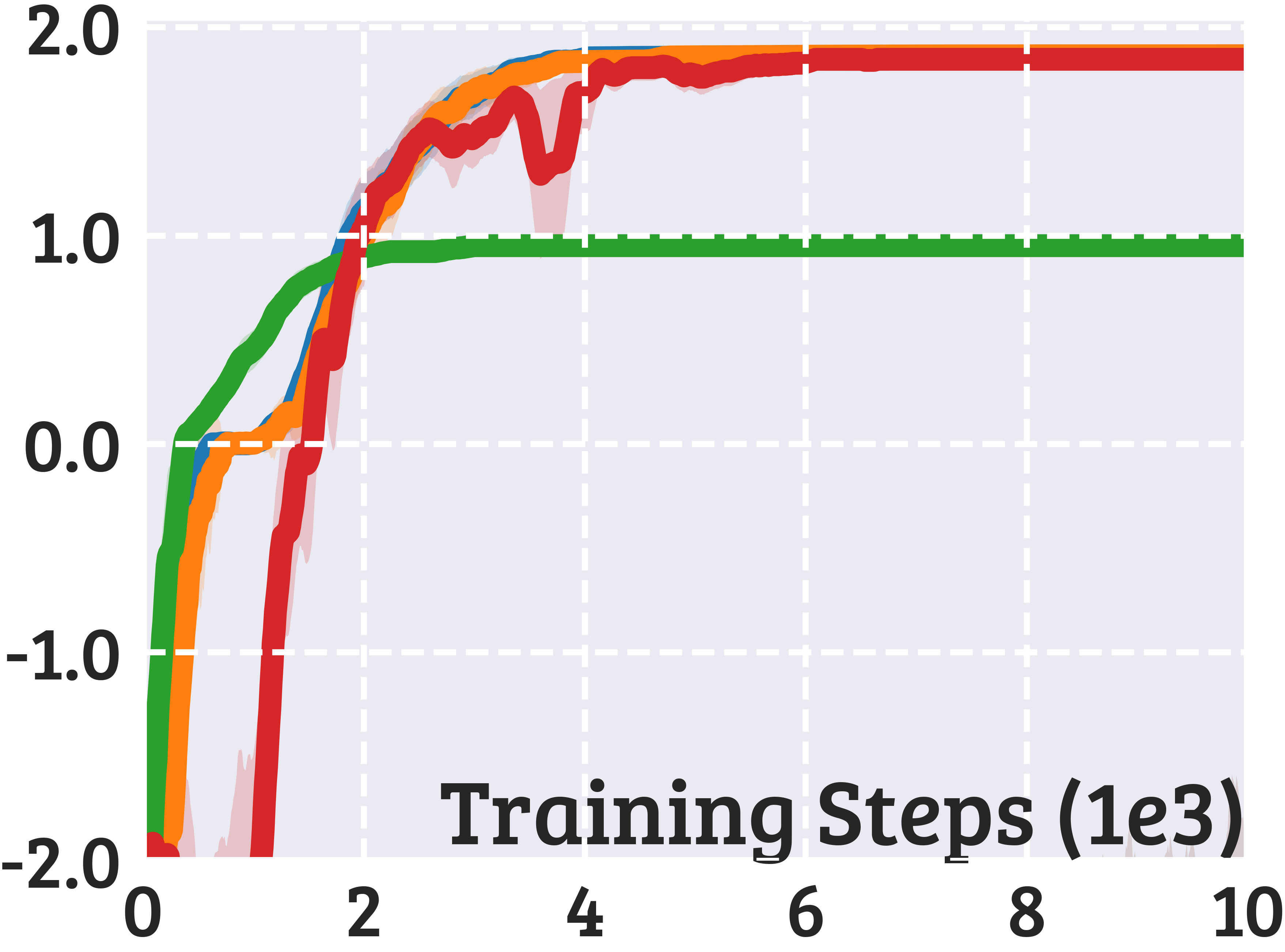}
    \end{subfigure}%
    }
    \\[3pt]
    \makebox[\linewidth][c]{%
    \raisebox{24pt}{\rotatebox[origin=t]{90}{\fontfamily{qbk}\footnotesize\textbf{{Noisy Reward}}}}
    \hfill
    \begin{subfigure}[b]{0.17\linewidth}
        \centering
        \includegraphics[width=\linewidth]{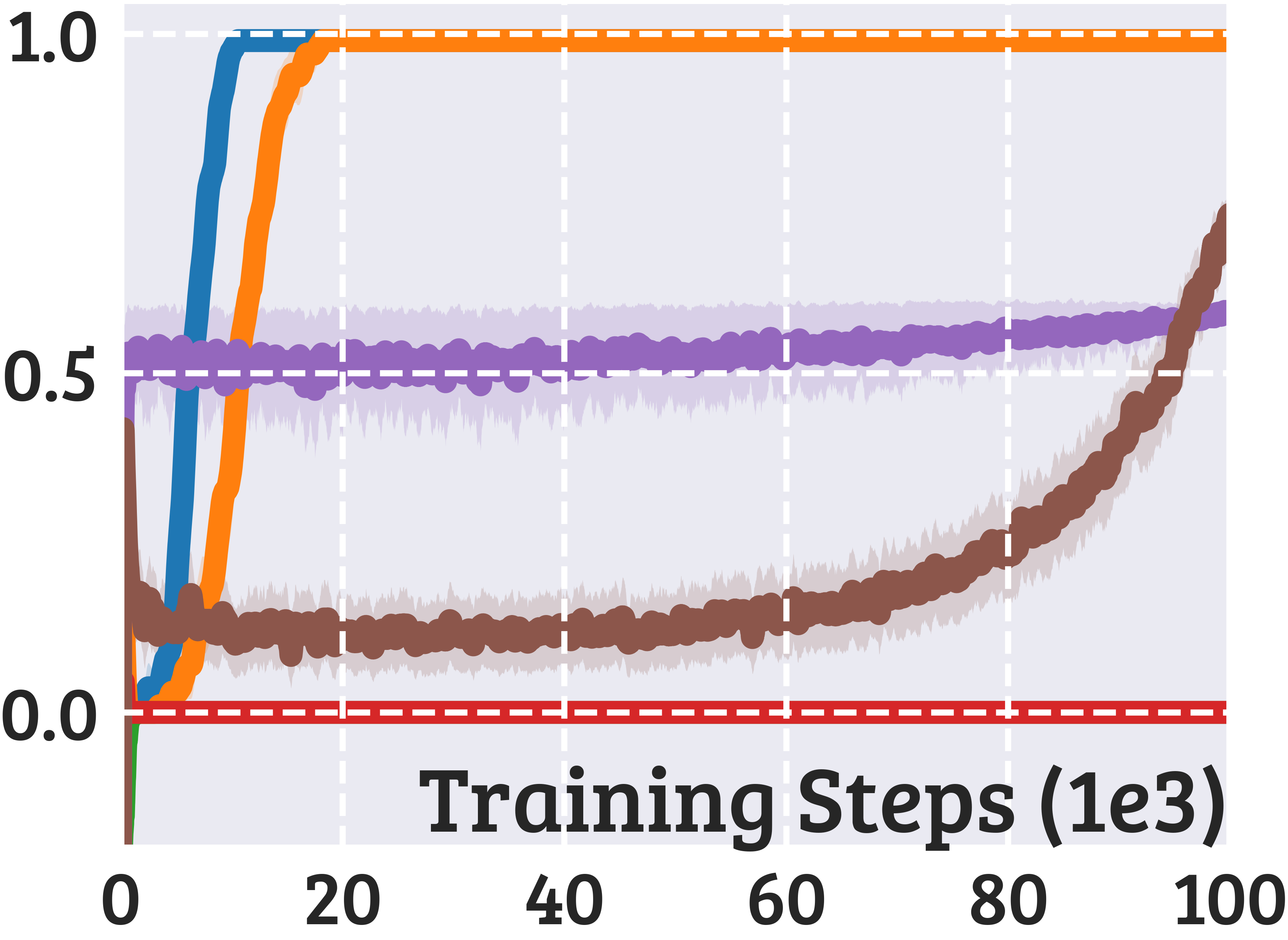}
    \end{subfigure}
    \hfill
    \begin{subfigure}[b]{0.17\linewidth}
        \centering
        \includegraphics[width=\linewidth]{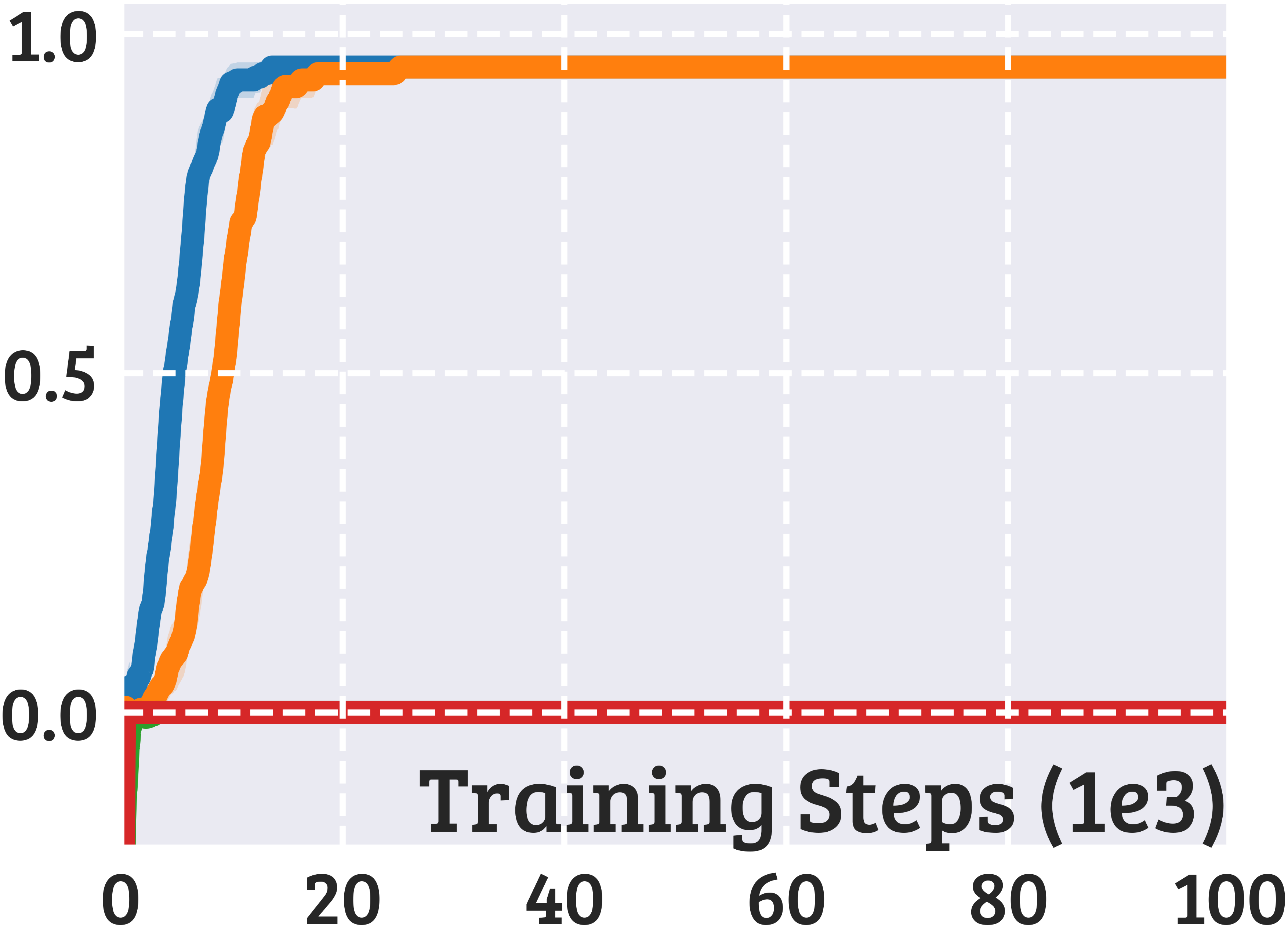}
    \end{subfigure}
    \hfill
    \begin{subfigure}[b]{0.17\linewidth}
        \centering
        \includegraphics[width=\linewidth]{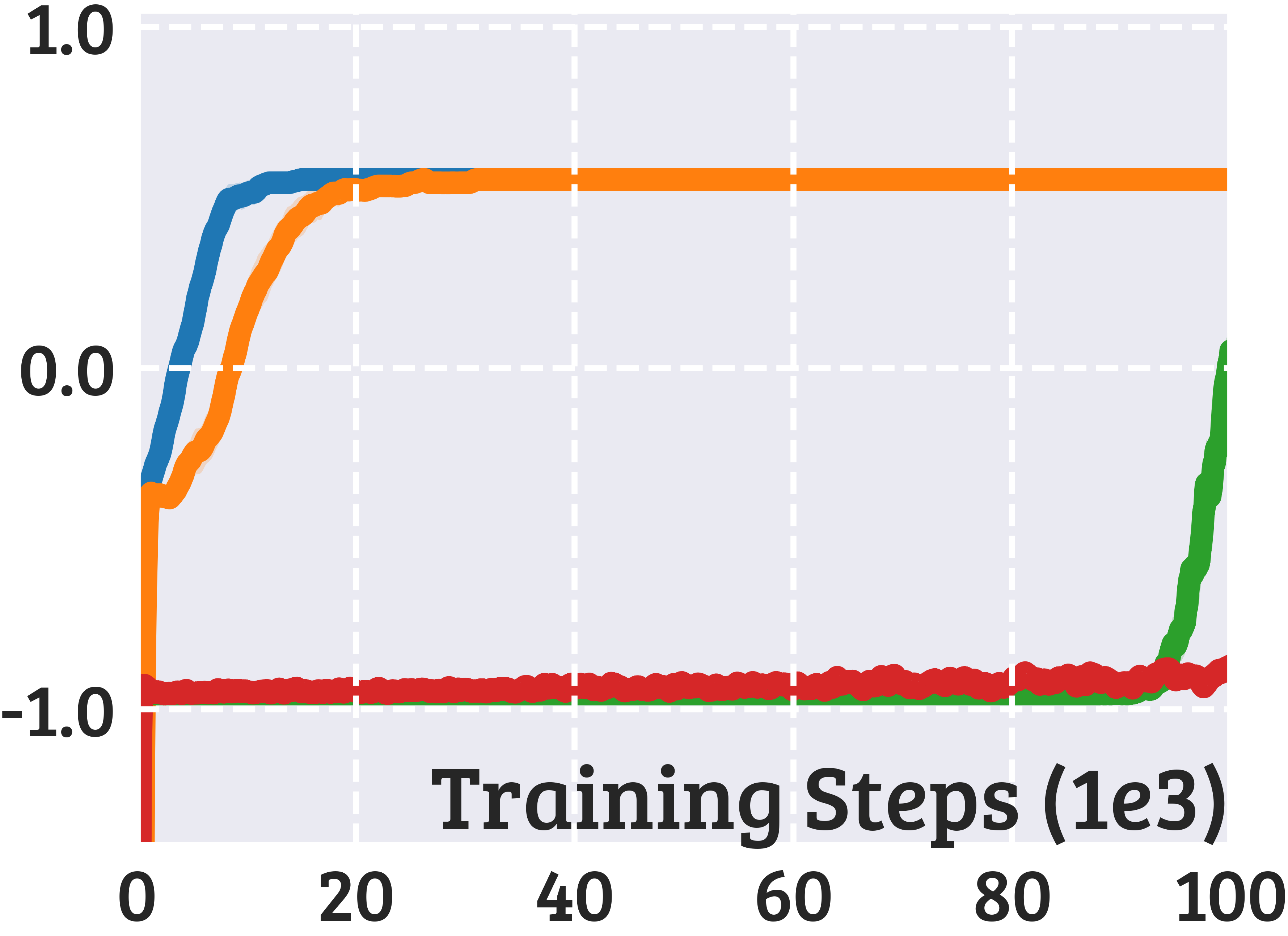}
    \end{subfigure}
    \hfill
    \begin{subfigure}[b]{0.17\linewidth}
        \centering
        \includegraphics[width=\linewidth]{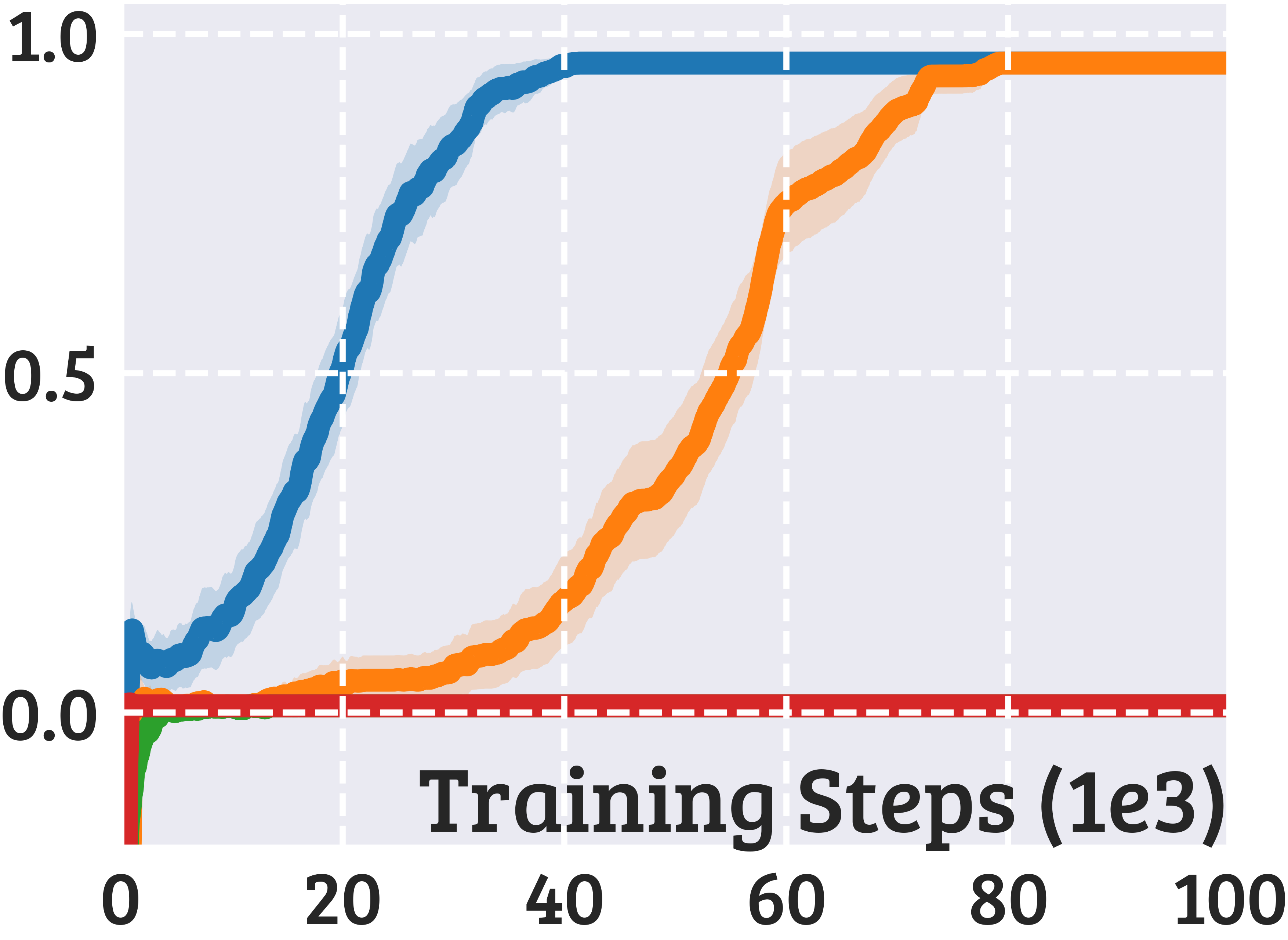}
    \end{subfigure}
    \hfill
    \begin{subfigure}[b]{0.17\linewidth}
        \centering
        \includegraphics[width=\linewidth]{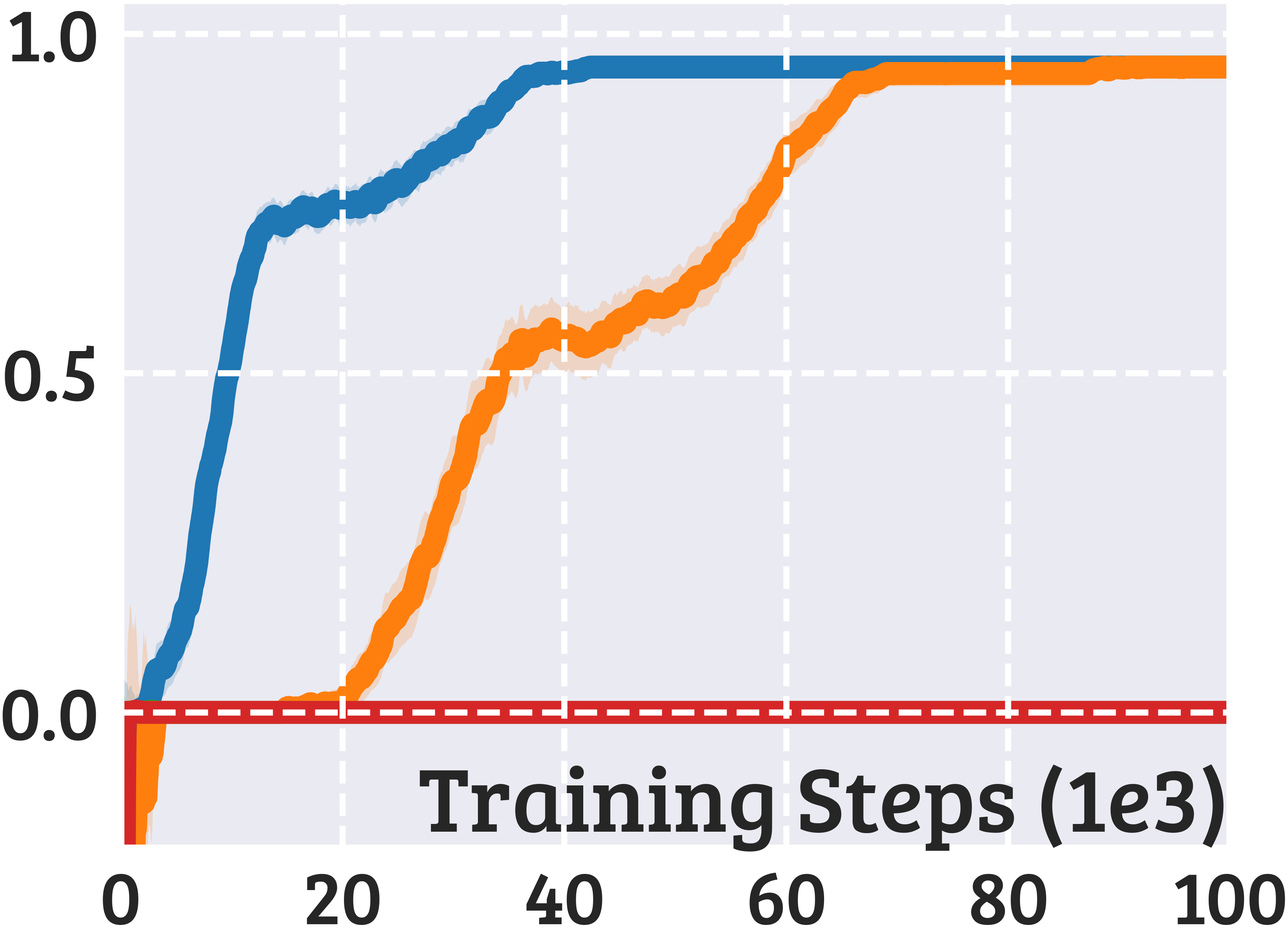}
    \end{subfigure}
    \hfill
    \begin{subfigure}[b]{0.17\linewidth}
        \centering
        \includegraphics[width=\linewidth]{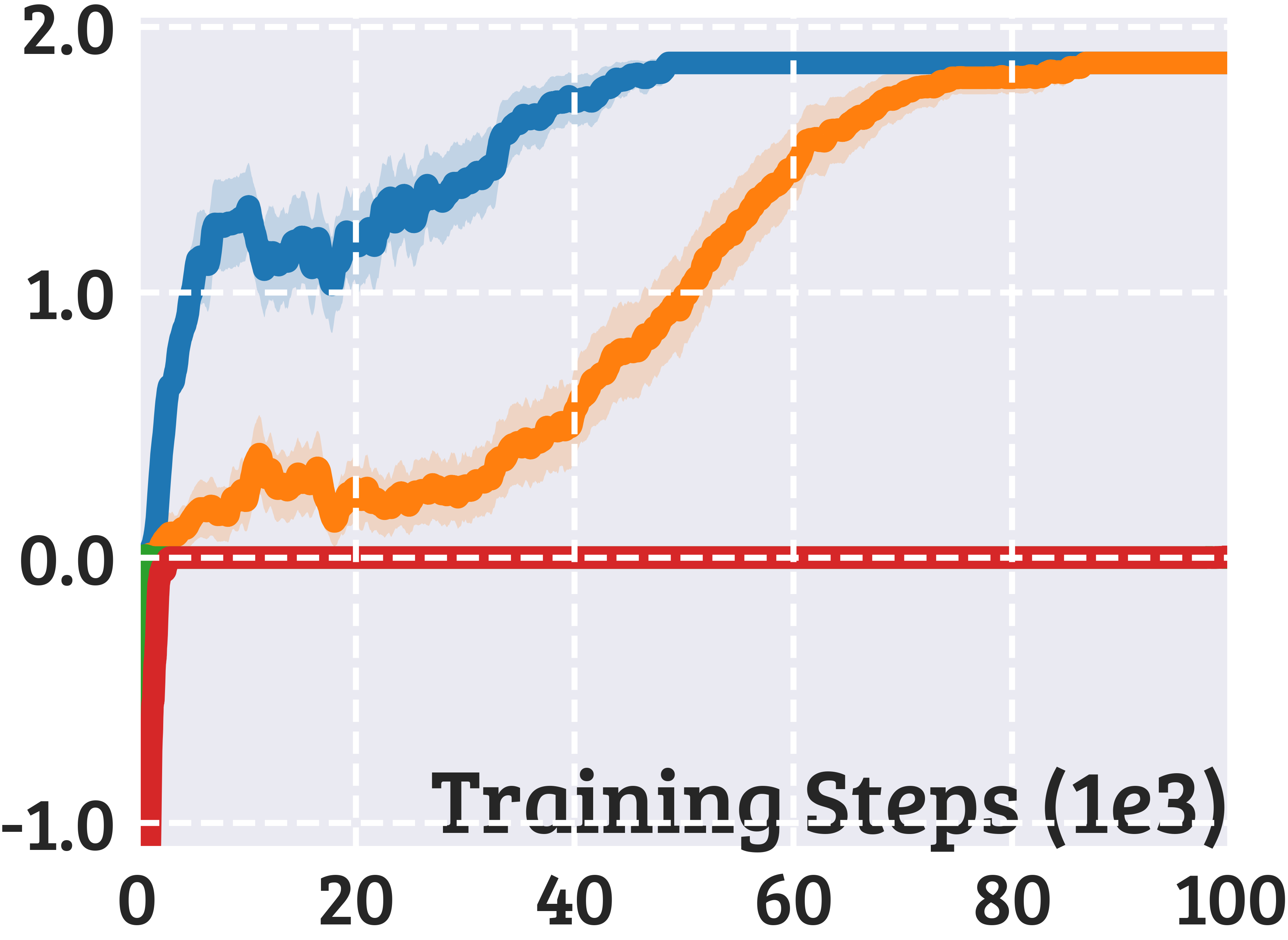}
    \end{subfigure}%
    }
    \caption{\label{fig:plots}\textbf{Expected discounted return of the greedy policies averaged over 100 seeds (shaded areas denote 95\% confidence interval).} \textmd{As expected from Table~\ref{tab:rate}, there is a large gap between the \textcolor{oracle}{Oracle} and algorithms receiving unobservable rewards. 
    In particular, \textcolor{zero}{$\boldsymbol{\rewardundefined = 0}$} and \textcolor{ignore}{Ignore $\rewardundefined$} perform extremely poorly and their curves are far below the others.
    The gap is smaller --- but still significant --- between the \textcolor{oracle}{Oracle} and \textcolor{model}{Reward Model}. The model $\smash{\widehat{R}(\senv, \aenv)}$, indeed, can mitigate the effect of noisy rewards. Nonetheless, because the agent will often observe $\rprox_t = \rewardundefined$ --- especially in N-Monitor, Limited-Time, and Limited-Use --- and because of poor $\varepsilon$-greedy exploration, \textcolor{model}{Reward Model} needs more samples to converge to an optimal policy. 
    }}
\end{figure*}

In Table~\ref{tab:rate}, we report the steps that each algorithm needed to converge to an optimal policy. If an algorithm did not converge to an optimal policy even once, it is not reported. 
In Figure~\ref{fig:plots}, we report the trend of the greedy policy expected discounted return against training steps.
We recall that the \textcolor{oracle}{Oracle} is Q-Learning with fully-observable rewards (the agent always receives $\rprox_t = \renv_t$) that still executes monitor actions and receives monitor rewards. For a fair comparison against \textcolor{model}{Reward Model}, the \textcolor{oracle}{Oracle} also learns $\smash{\widehat{R}(\senv, \aenv)}$. As a matter of fact, when rewards are noisy, a reward model can significantly help the agent.\footnote{With deterministic rewards, instead, using the reward model did not change the performance of the \textcolor{oracle}{Oracle} in our experiments.}

Results show that the \textcolor{oracle}{Oracle} clearly converges faster than all other algorithms. 
Table~\ref{tab:rate} reports that only the \textcolor{oracle}{Oracle} converges in all Mon-MDPs within the steps limit in all 100 seeds, and Figure~\ref{fig:plots} shows that there is a large gap in how fast the \textcolor{oracle}{Oracle} and other baselines learn. 
Results also stress how harder N-Monitor, Limited-Time, and Limited-Use are compared to the other Mon-MDPs.

\begin{itemize}[leftmargin=1em, itemsep=1pt, topsep=3pt]
\item \textbf{N-Monitor.} This Mon-MDP has larger state-action spaces, and positive monitor rewards for \textit{not} observing environment rewards. Until the agent observes the +1 goal reward, monitor rewards act like ``distractors'', discouraging the agent from asking to be monitored. Indeed, \textcolor{joint}{Joint} completely fails, as it exhibits the behavior discussed in Section~\ref{app:subsec:joint_counter}. Furthermore, only $1$ out of $N$ monitor actions allows the agent to observe environment rewards. The larger $N$, the less effective $\varepsilon$-greedy exploration is.
\item \textbf{Limited-Time.} In this Mon-MDP, exploration is crucial because after some steps the agent will stop being monitored until the end of the episode. 
\textcolor{sequential}{Sequential} and \textcolor{joint}{Joint} fail in many seeds. \textcolor{model}{Reward Model} almost always converges within the steps limit, as $\smash{\widehat{R}(\senv,\aenv)}$ compensates for the unobservability of the reward. However, it needs significantly more samples than the \textcolor{oracle}{Oracle} because it observes environment rewards less frequently, and therefore updates its reward model less often. This is a primary example of why $\varepsilon$-greedy exploration works poorly in Mon-MDPs --- rather than trying new actions randomly with probability $\varepsilon$, the agent should try actions for which it has not observed the reward yet, making the most out of the time monitoring is \texttt{ON}.
\item \textbf{Limited-Use.} Exploration in this Mon-MDP is hard for two reasons. First, it is non-invariant, and this is the reason why \textcolor{sequential}{Sequential} fails --- rather than going straight to the goal, the agent must wait for the battery to deplete, e.g., by bumping into walls. 
Second, the agent can turn the monitor \texttt{OFF} at any time. 
On one hand, this allows the agent to balance battery consumption and observe more environment rewards. On the other hand, the agent must deplete the battery to observe $\rmon_t = 1$. 
In such a scenario, a random $\varepsilon$-greedy exploration is clearly not the best strategy. For example, a smart agent would turn \texttt{ON} the monitor in states where it has not observed environment rewards yet. Or even, the agent may plan long-term actions knowing that it can observe rewards only $N$ times during an episode (i.e., one time per battery level). 
\end{itemize}

\textit{These results shown are primary examples of why better exploration strategies must be developed for Mon-MDPs.}

\begin{table*}[hb]
    \renewcommand{\arraystretch}{1.05}
    \setlength\tabcolsep{3pt}
    \centering
    \begin{adjustbox}{center}
    \begin{tabular}{@{\extracolsep{\fill}}|l|c|c|c|c|c|c|c|c|c|c|c|c|}
\multicolumn{13}{c}{\textbf{With Deterministic Environment Reward (Max 10,000 Training Steps)}}
\\
\hline
& 
\multicolumn{2}{c|}{\textbf{Simple}} & 
\multicolumn{2}{c|}{\textbf{Penalty}} & 
\multicolumn{2}{c|}{\textbf{Button}} & 
\multicolumn{2}{c|}{\textbf{N-Monitor}} & 
\multicolumn{2}{c|}{\textbf{Limited-Time}} &
\multicolumn{2}{c|}{\textbf{Limited-Use}}
\\
\hline
& 
Steps & \% & 
Steps & \% & 
Steps & \% & 
Steps & \% & 
Steps & \% & 
Steps & \%
\\
\hline
\hline
\textcolor{oracle}{\textbf{Oracle}} & $97 \pm 16$ & 100 & $353 \pm 26$ & 100 & $612 \pm 44$ & 100 & $1,568 \pm 59$ & 100 & $1,911 \pm 156$ & 100 & $2,157 \pm 122$ & 100
\\
\hline
\textcolor{model}{\textbf{Rew. Model}} & $160 \pm 21$ & 100 & $445 \pm 30$ & 100 & $662 \pm 43$ & 100 & $2,146 \pm 150$ & 100 & $2,109 \pm 170$ & 97 & $2,252 \pm 123$ & 100
\\
\hline
\textcolor{sequential}{\textbf{Sequential}} & $134 \pm 17$ & 100 & $533 \pm 38$ & 100 & --- & 0 & $1,639 \pm 102$ & 100 & $2,728 \pm 337$ & 72 & --- & 0
\\
\hline
\textcolor{joint}{\textbf{Joint}} & $130 \pm 14$ & 100 & $535 \pm 46$ & 100 & --- & 0 & --- & 0 & $2,755 \pm 309$ & 73 & $2,940 \pm 238$ & 99
\\
\hline
\multicolumn{13}{c}{}
\\[-4pt]
\multicolumn{13}{c}{\textbf{With Noisy Environment Reward (Max 100,000 Training Steps)}}
\\
\hline
\textcolor{oracle}{\textbf{Oracle}}  & $6,703 \pm 340$ & 100 & $5,686 \pm 409$ & 100 & $5,640 \pm 462$ & 100 & $23,087 \pm 1,508$ & 100 & $26,977 \pm 1,652$ & 100 & $20,197 \pm 2,705$ & 100
\\
\hline
\textcolor{model}{\textbf{Rew. Model}}  & $10,790 \pm 599$ & 100 & $9,793 \pm 640$ & 100 & $11,312 \pm 1,032$ & 100 & $56,346 \pm 2,460$ & 97 & $58,790 \pm 1,153$ & 99 & $47,342 \pm 2,866$ & 97
\\
\hline
    \end{tabular}
    \end{adjustbox}
    \caption{\label{tab:rate}\textbf{Empirical rate of convergence to an optimal policy.} \textmd{%
    ``\%'' denotes in how many seeds an algorithm successfully converged to an optimal policy. 
    ``Steps'' is the number of steps an algorithm needed to converge to an optimal policy averaged over the successful seeds with 95\% confidence interval.
    For example, ``Steps $58,790.6 \pm 1,153.3$ and 99\%'' means that the algorithm converged to an optimal policy in 99/100 seeds, and in those 99 seeds it converged within 58,790.6 steps on average. 
    Algorithms not reported did not converged to an optimal policy even once.
    N-Monitor, Limited-Time, and Limited-Use are clearly the hardest Mon-MDPs. In N-Monitor, \textcolor{joint}{Joint} never converged to an optimal policy --- positive monitor rewards make the agent behave as discussed in Section~\ref{app:subsec:joint_counter}. In Limited-Use, \textcolor{sequential}{Sequential} did not converge as well --- the Mon-MDP is not invariant. In Limited-Time, both \textcolor{sequential}{Sequential} and \textcolor{joint}{Joint} converged to an optimal policy only 72-73\% of the seeds. 
    In the noisy setting, only \textcolor{model}{Reward Model} converged to optimal policies, but not 100\% of the seeds, and on average it needed 2-2.5 times more samples than the \textcolor{oracle}{Oracle}. 
    }
    }
\end{table*}

\vspace*{-5pt}

\clearpage

\newgeometry{left=1.4cm, right=1.4cm, top=1.5cm, bottom=1.5cm}

\twocolumn

\subsection{``Assign a Constant Value to $\rewardundefined$'' Ablation}
\label{app:subsec:ablation_zero}
\begin{figure}[h]
    \setlength{\belowcaptionskip}{0pt}
    \setlength{\abovecaptionskip}{3pt}
    \centering
    \includegraphics[width=0.5\linewidth]{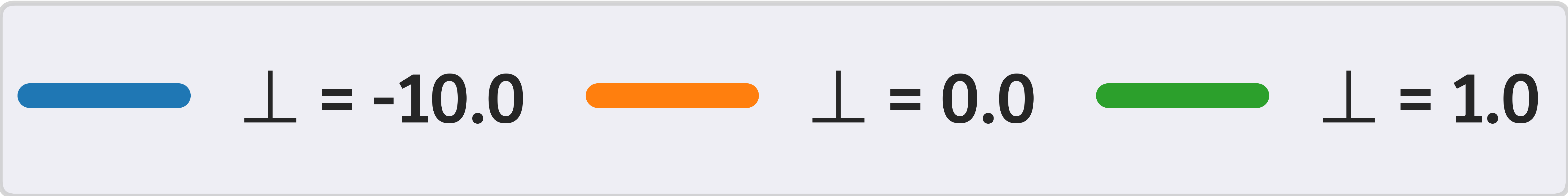}
    \\[3pt]
    \begin{subfigure}[b]{0.32\linewidth}
        \centering
        {\fontfamily{qbk}\footnotesize\textbf{Simple}}\\[3pt]
        \includegraphics[width=\linewidth]{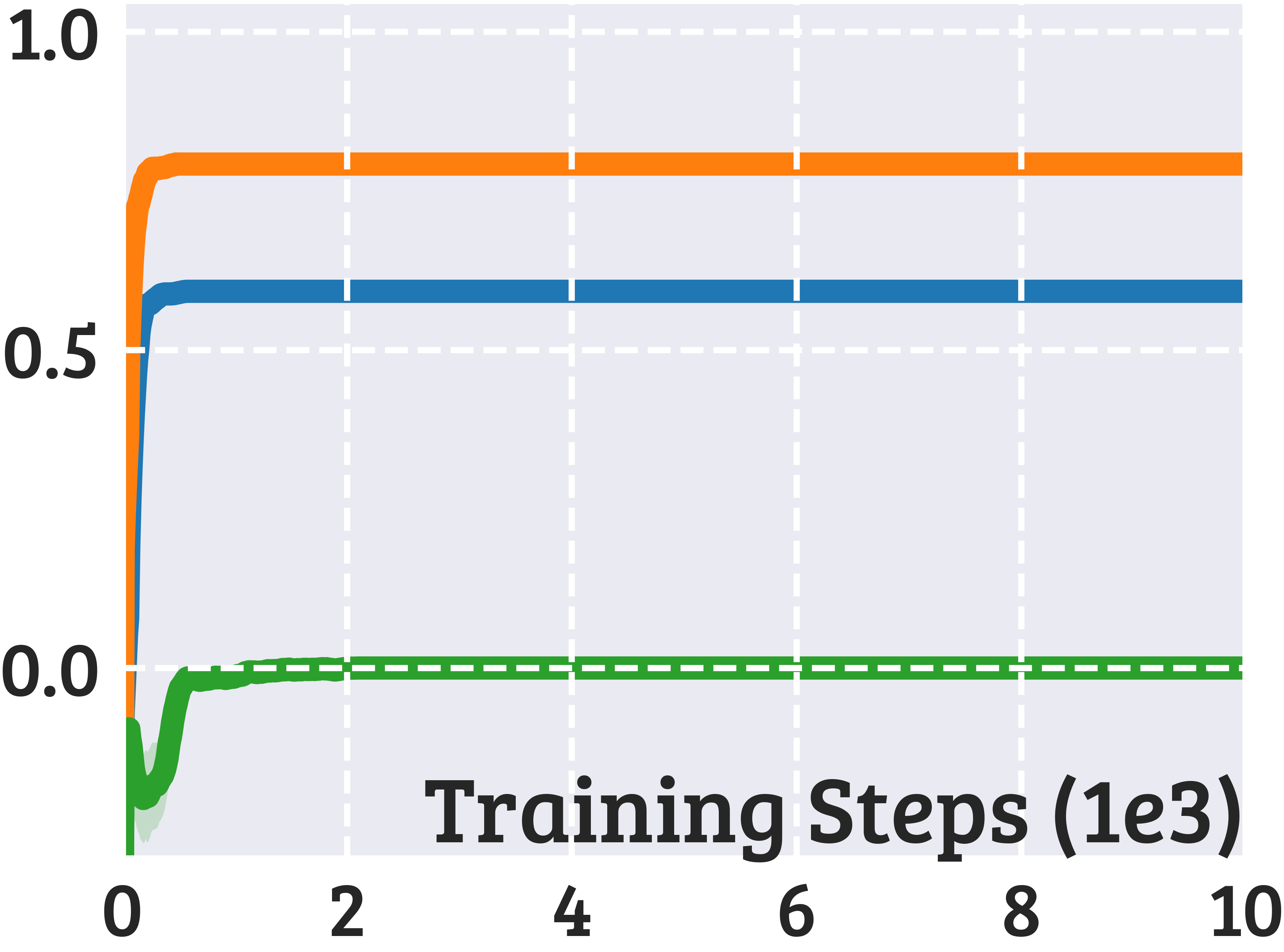}
    \end{subfigure}
    \hfill
    \begin{subfigure}[b]{0.32\linewidth}
        \centering
        {\fontfamily{qbk}\footnotesize\textbf{Penalty}}\\[3pt]
        \includegraphics[width=\linewidth]{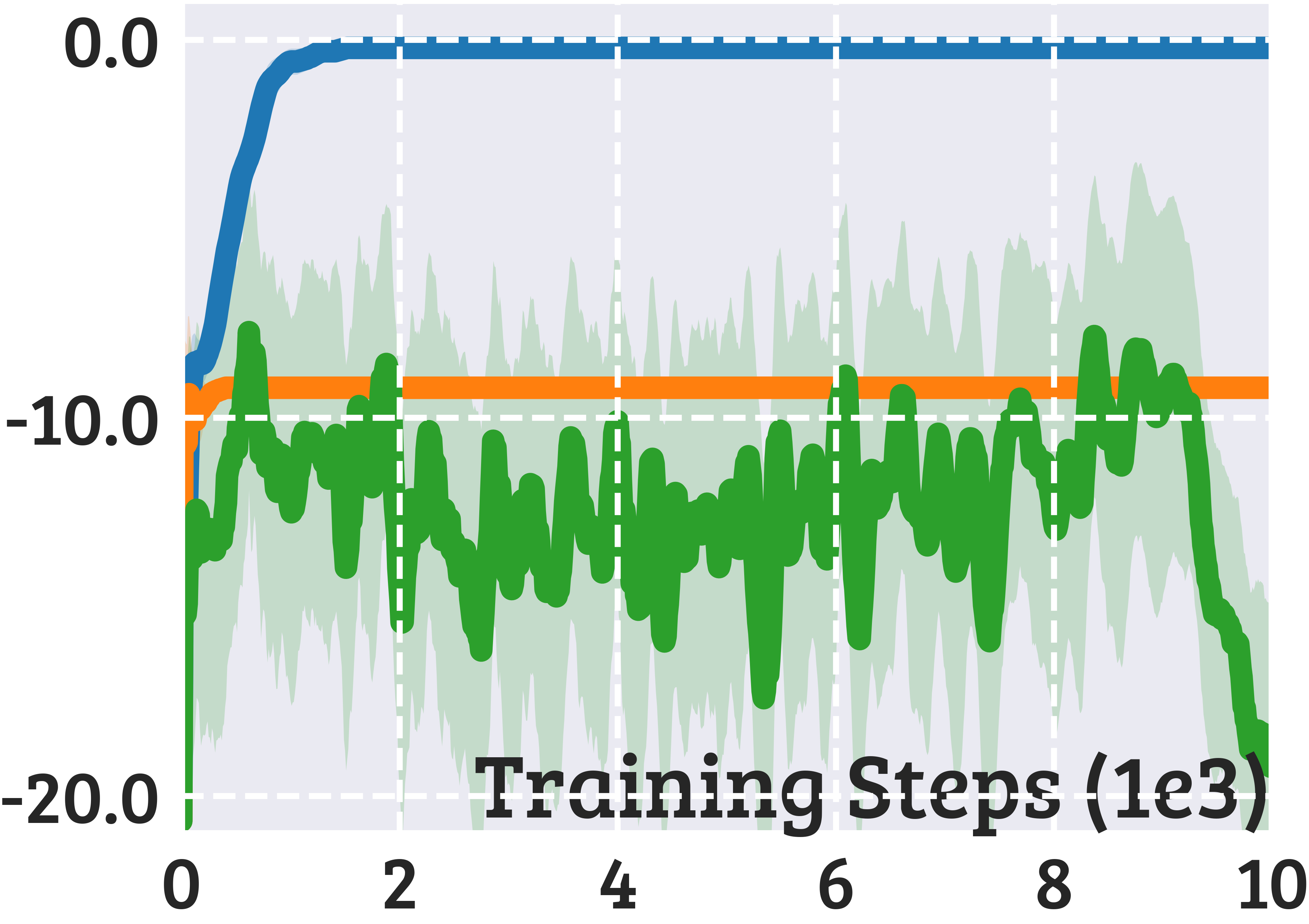}
    \end{subfigure}
    \hfill
    \begin{subfigure}[b]{0.32\linewidth}
        \centering
        {\fontfamily{qbk}\footnotesize\textbf{Button}}\\[3pt]
        \includegraphics[width=\linewidth]{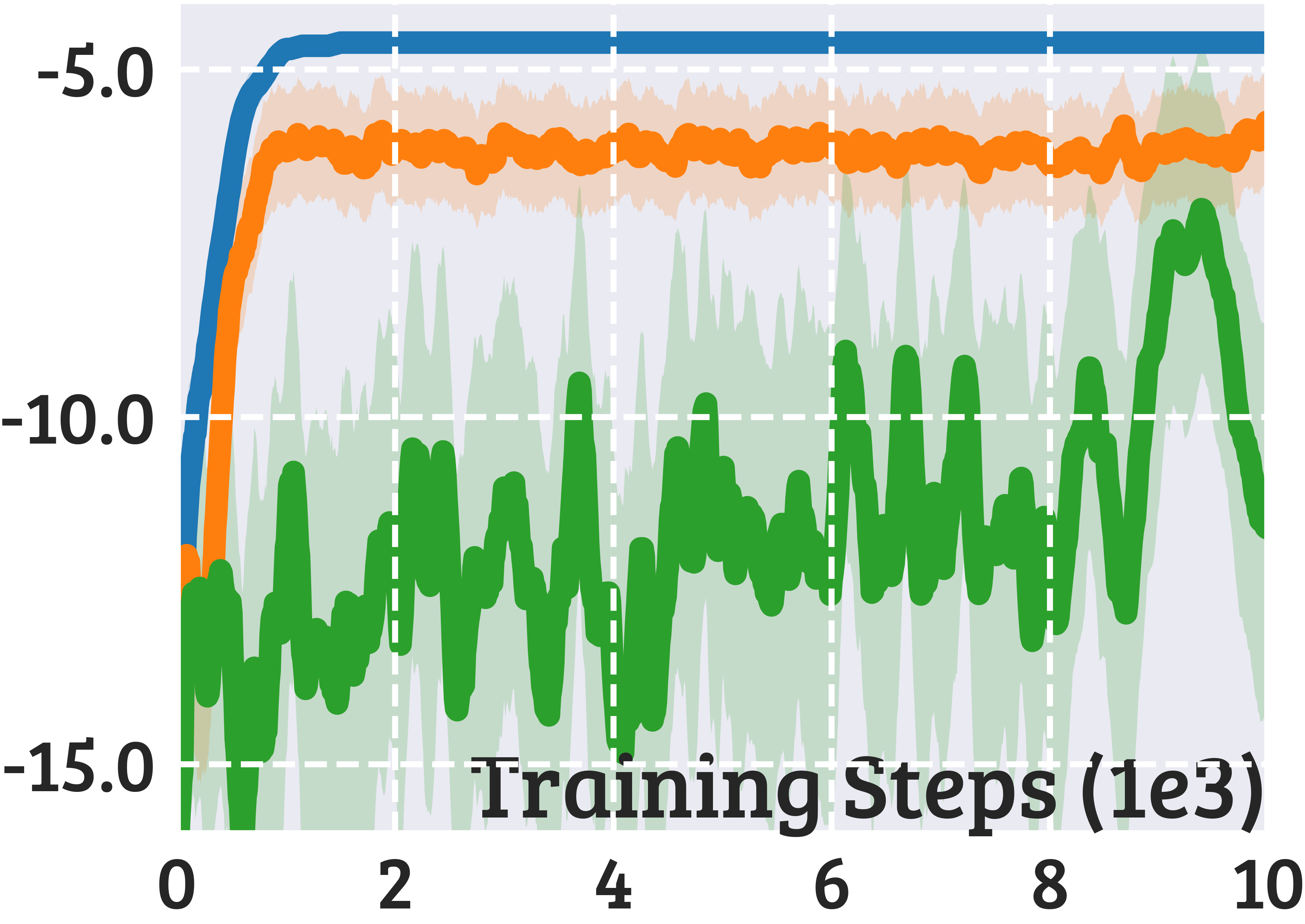}
    \end{subfigure}
    \caption{\label{fig:ablation_zero}\textbf{Expected discounted return on varying the constant value assigned to $\rewardundefined$.} \textmd{Shades denote 95\% confidence interval over 100 seeds. In no case, the algorithm converged to an optimal policy, as emphasized by the policy plots below.}}
\end{figure}

\begin{figure}[h]
\centering
    \setlength{\belowcaptionskip}{0pt}
    \setlength{\abovecaptionskip}{3pt}
    \centering
    \includegraphics[trim=0 33pt 0 0, width=0.9\columnwidth]{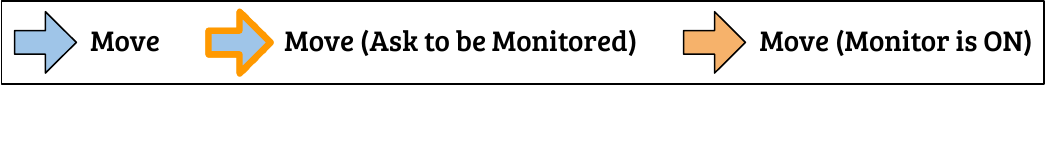}
    \\[3pt]
    \includegraphics[width=.32\columnwidth]{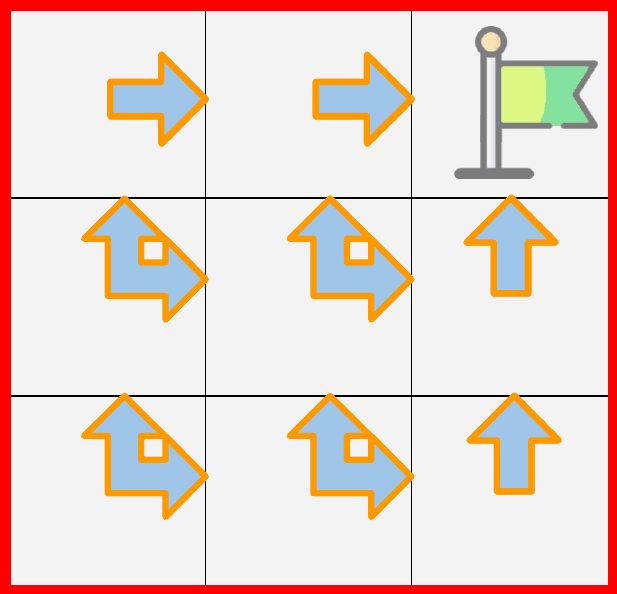}
    \hfill
    \includegraphics[width=.32\columnwidth]{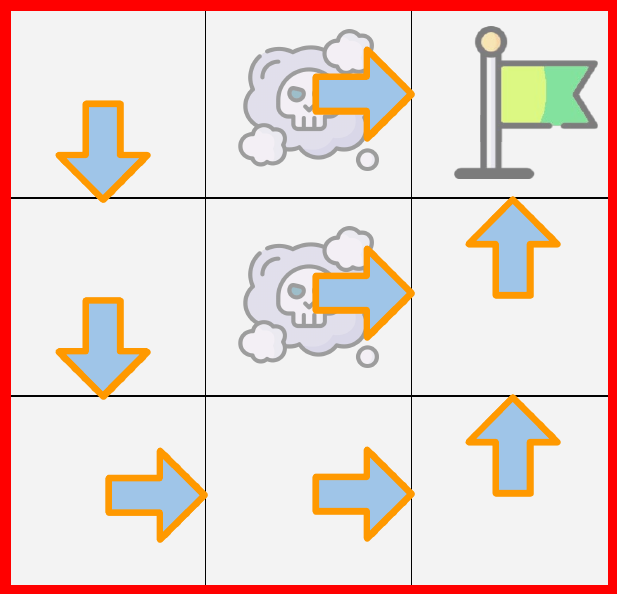}
    \hfill
    \includegraphics[width=.32\columnwidth]{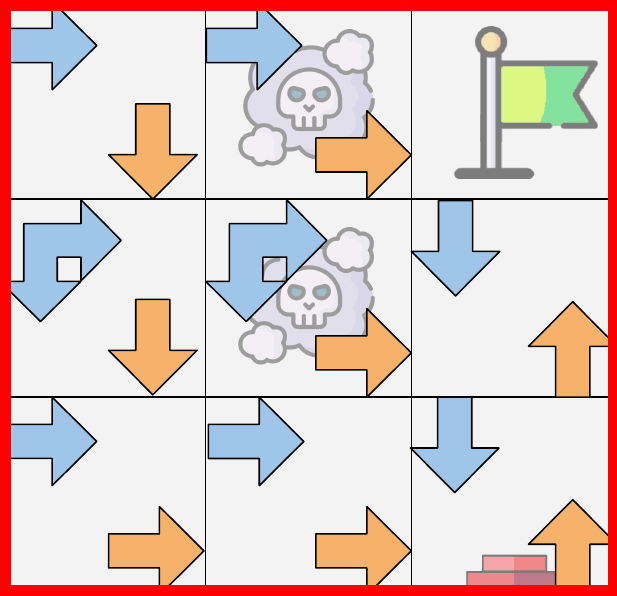}
    \caption{\label{fig:policies_ablation_-10}Assigning $\boldsymbol{\rewardundefined = -10}$ \textmd{can be considered a pessimistic strategy, because no environment reward is worse than -10. As a result, the agent always seeks monitoring.}}
    \bigskip
    \includegraphics[width=.32\columnwidth]{fig/policies/policies_zero_easy_v3.pdf}
    \hfill
    \includegraphics[width=.32\columnwidth]{fig/policies/policies_zero_medium_v3.pdf}
    \hfill
    \includegraphics[width=.32\columnwidth]{fig/policies/policies_zero_hard_v3.pdf}
    \caption{\label{fig:policies_ablation_0}Assigning $\boldsymbol{\rewardundefined = 0}$ \textmd{can be considered a neutral strategy. However, the agent ends up not asking to be monitored if the environment reward would be worse, as in penalty cells.}}
    \bigskip
    \includegraphics[width=.32\columnwidth]{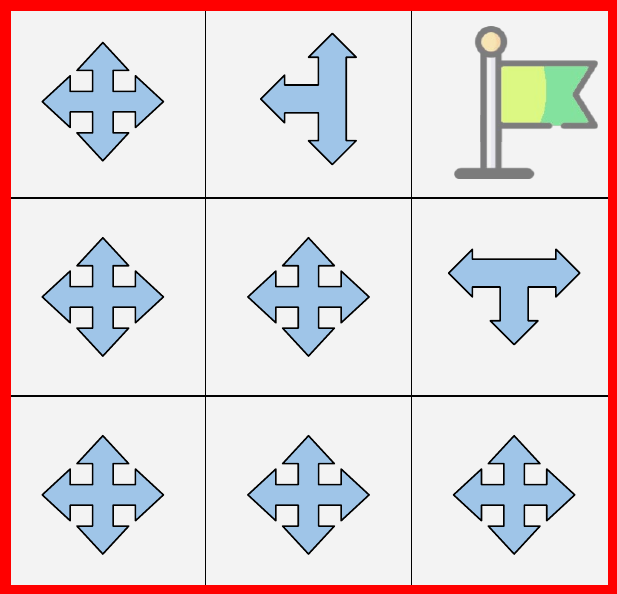}
    \hfill
    \includegraphics[width=.32\columnwidth]{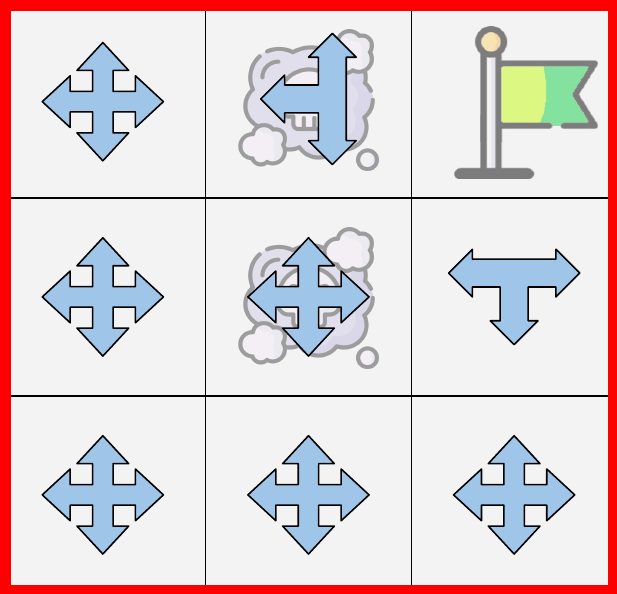}
    \hfill
    \includegraphics[width=.32\columnwidth]{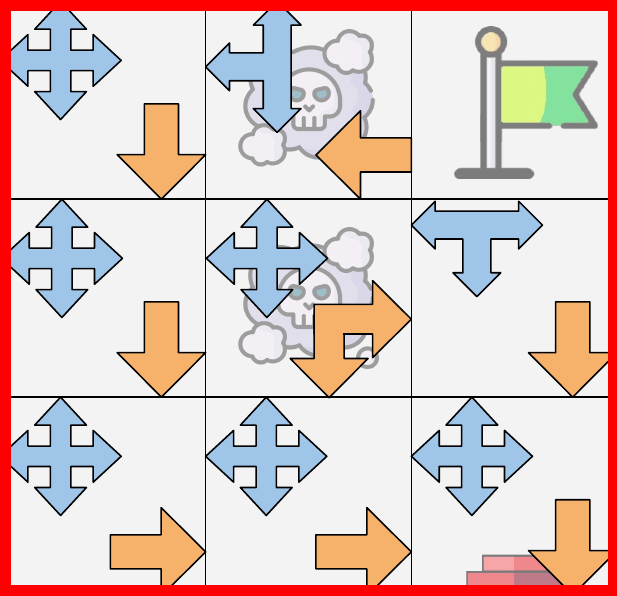}
    \caption{\label{fig:policies_ablation_1}Assigning $\boldsymbol{\rewardundefined = 1}$ \textmd{can be considered an optimistic strategy, as 1 is the highest environment reward the agent can observe. This is the worst strategy, because the agent is overly optimistic about what cannot observe. Rather than observing smaller environment rewards, the agent prefers to never seek monitoring and to never go the goal. This way, it can ``pretend'' to receive positive rewards forever. 
    For example, in the Button Mon-MDP, when monitoring is \texttt{ON} the agent goes to the button cell to turn it \texttt{OFF} (avoiding penalty cells). Then, it keeps walking around without ever going to the goal (not even avoiding penalty cells because monitoring is now \texttt{OFF}).}
    }
\end{figure}

Training curves in Figure~\ref{fig:ablation_zero} and policy plots in Figures~\ref{fig:policies_ablation_-10},~\ref{fig:policies_ablation_0}, and~\ref{fig:policies_ablation_1} show that the Q-Learning variant that assigns a constant value to $\rewardundefined$ performs poorly regardless of the value assigned. 

\subsection{Initial Q-Value Ablation}
\label{app:subsec:ablation_q0}

Here, we investigate how the Q-function initialization affects learning in the Simple, Penalty, and Button Mon-MDPs with deterministic rewards. We consider initial Q-values -10, 0, 1. The former and the latter correspond to pessimistic and optimistic initialization, respectively (-10 is less or equal than the lowest Q-value of the optimal Q-functions, while 1 is greater or equal than the highest Q-value). Note that for \textcolor{sequential}{Sequential} and \textcolor{joint}{Joint}, both $\qenv$ and $\qmon$ are initialized with the same value.

\begin{figure}[h]
    \setlength{\belowcaptionskip}{0pt}
    \setlength{\abovecaptionskip}{3pt}
    \centering
    \includegraphics[width=0.95\linewidth]{fig/plots/supp/legend.png}
    \\[2pt]
    \raisebox{18pt}{\rotatebox[origin=t]{90}{\fontfamily{qbk}\footnotesize\textbf{Init -10}}}
    \hfill
    \begin{subfigure}[b]{0.30\linewidth}
        \centering
        {\fontfamily{qbk}\footnotesize\textbf{Simple}}\\[3pt]
        \includegraphics[width=\linewidth]{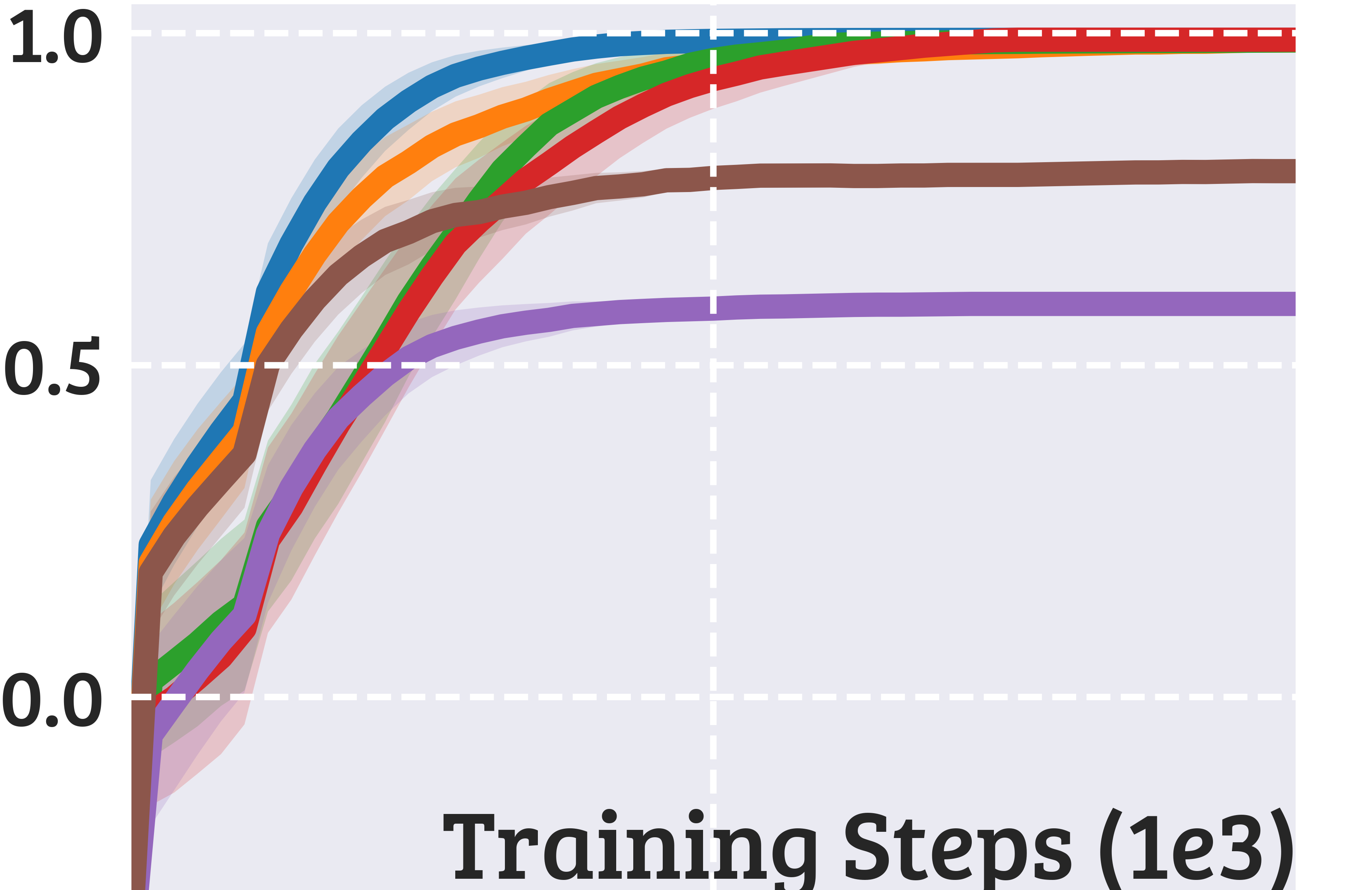}
    \end{subfigure}
    \hfill
    \begin{subfigure}[b]{0.30\linewidth}
        \centering
        {\fontfamily{qbk}\footnotesize\textbf{Penalty}}\\[3pt]
        \includegraphics[width=\linewidth]{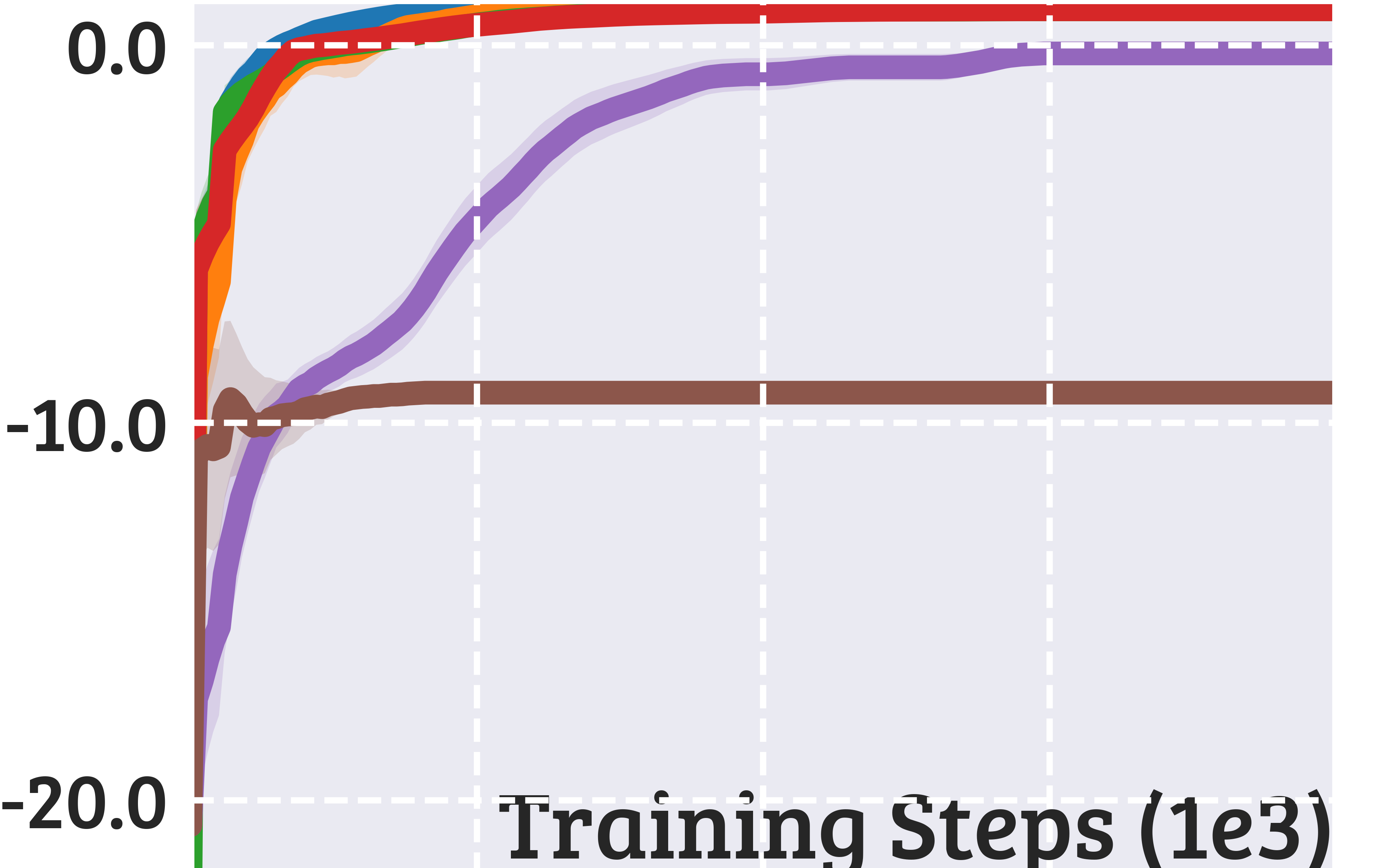}
    \end{subfigure}
    \hfill
    \begin{subfigure}[b]{0.30\linewidth}
        \centering
        {\fontfamily{qbk}\footnotesize\textbf{Button}}\\[3pt]
        \includegraphics[width=\linewidth]{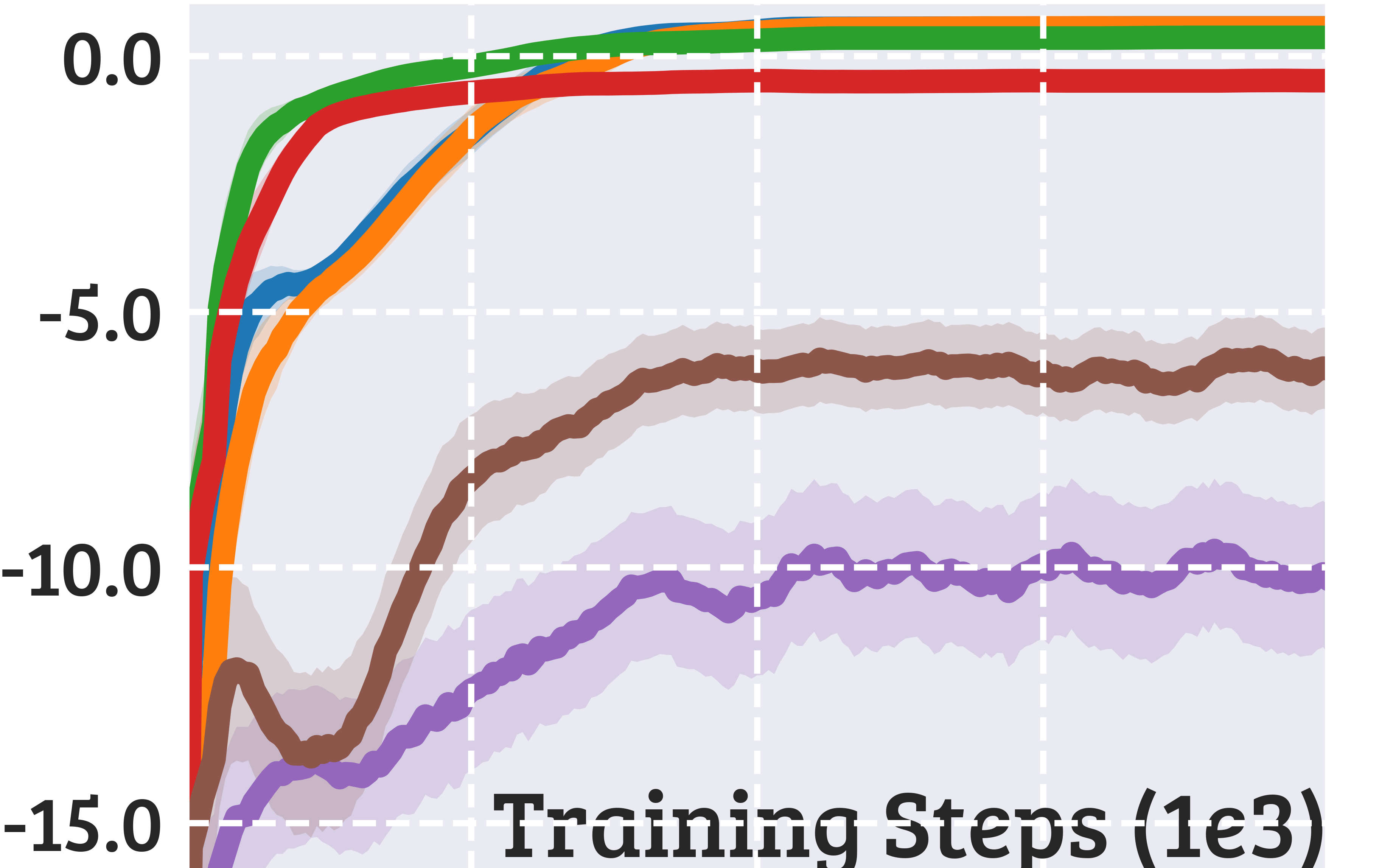}
    \end{subfigure}
    \\[2pt]
    \raisebox{18pt}{\rotatebox[origin=t]{90}{\fontfamily{qbk}\footnotesize\textbf{Init 0}}}
    \hfill
    \begin{subfigure}[b]{0.30\linewidth}
        \centering
        \includegraphics[width=\linewidth]{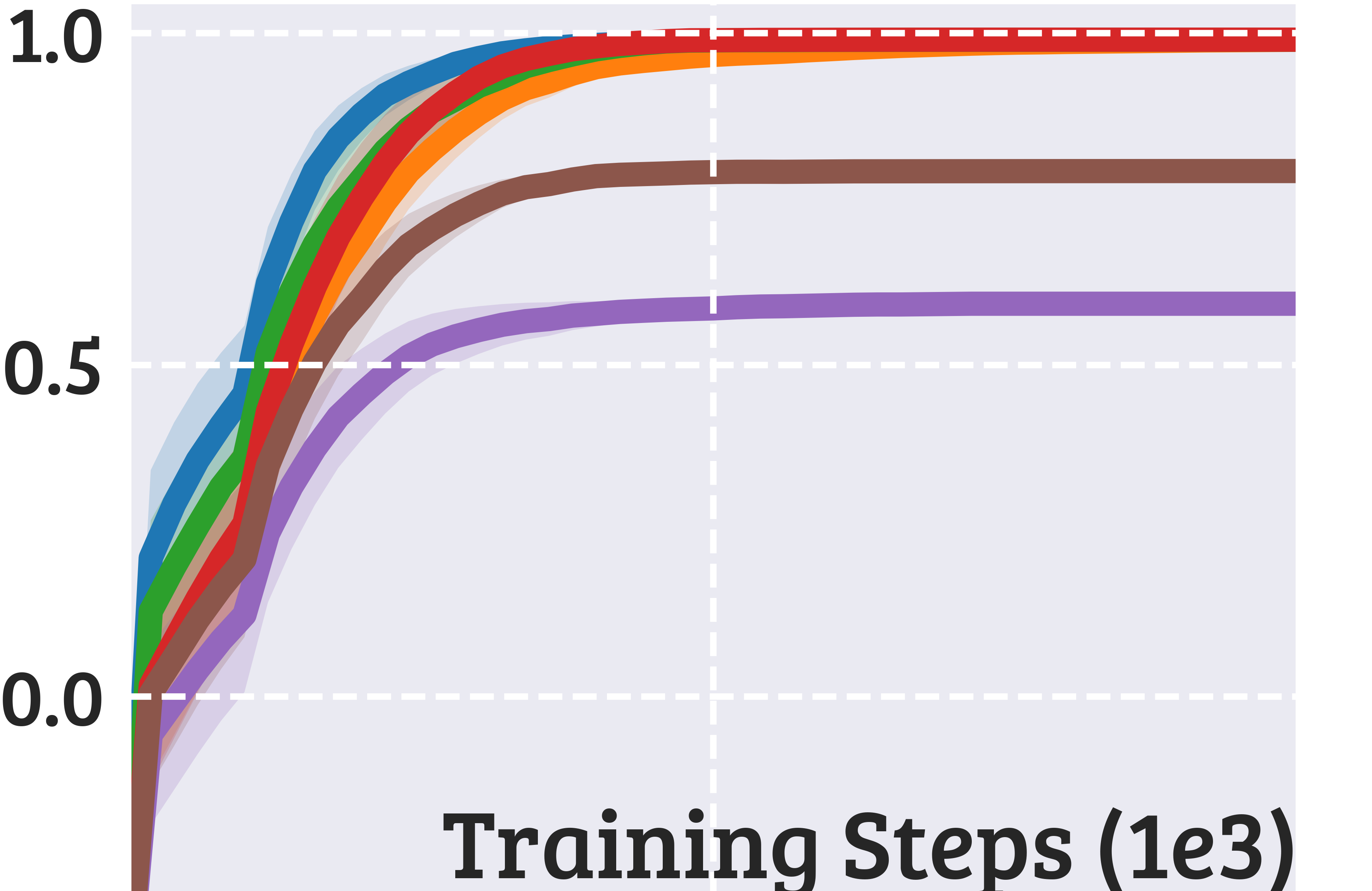}
    \end{subfigure}
    \hfill
    \begin{subfigure}[b]{0.30\linewidth}
        \centering
        \includegraphics[width=\linewidth]{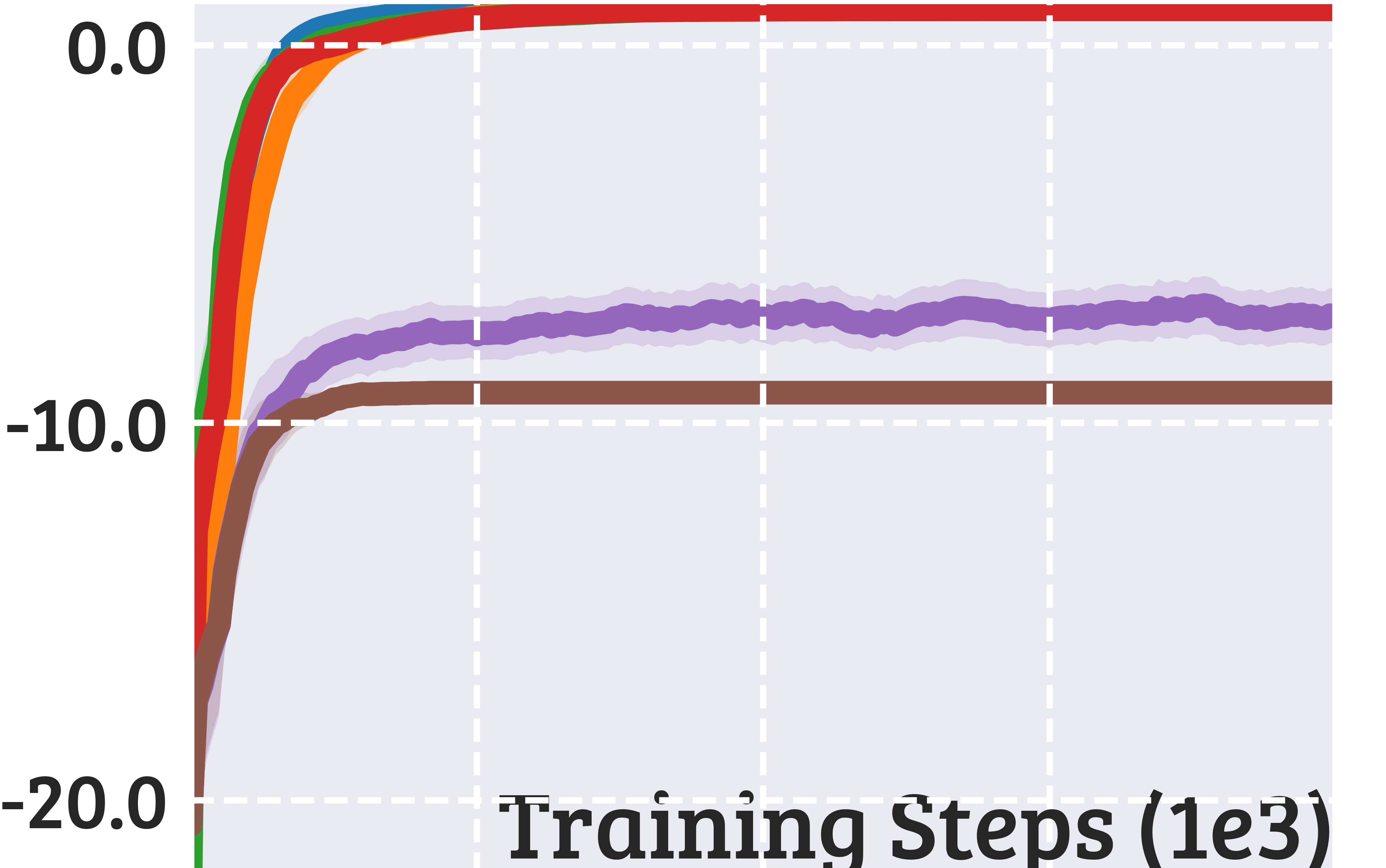}
    \end{subfigure}
    \hfill
    \begin{subfigure}[b]{0.30\linewidth}
        \centering
        \includegraphics[width=\linewidth]{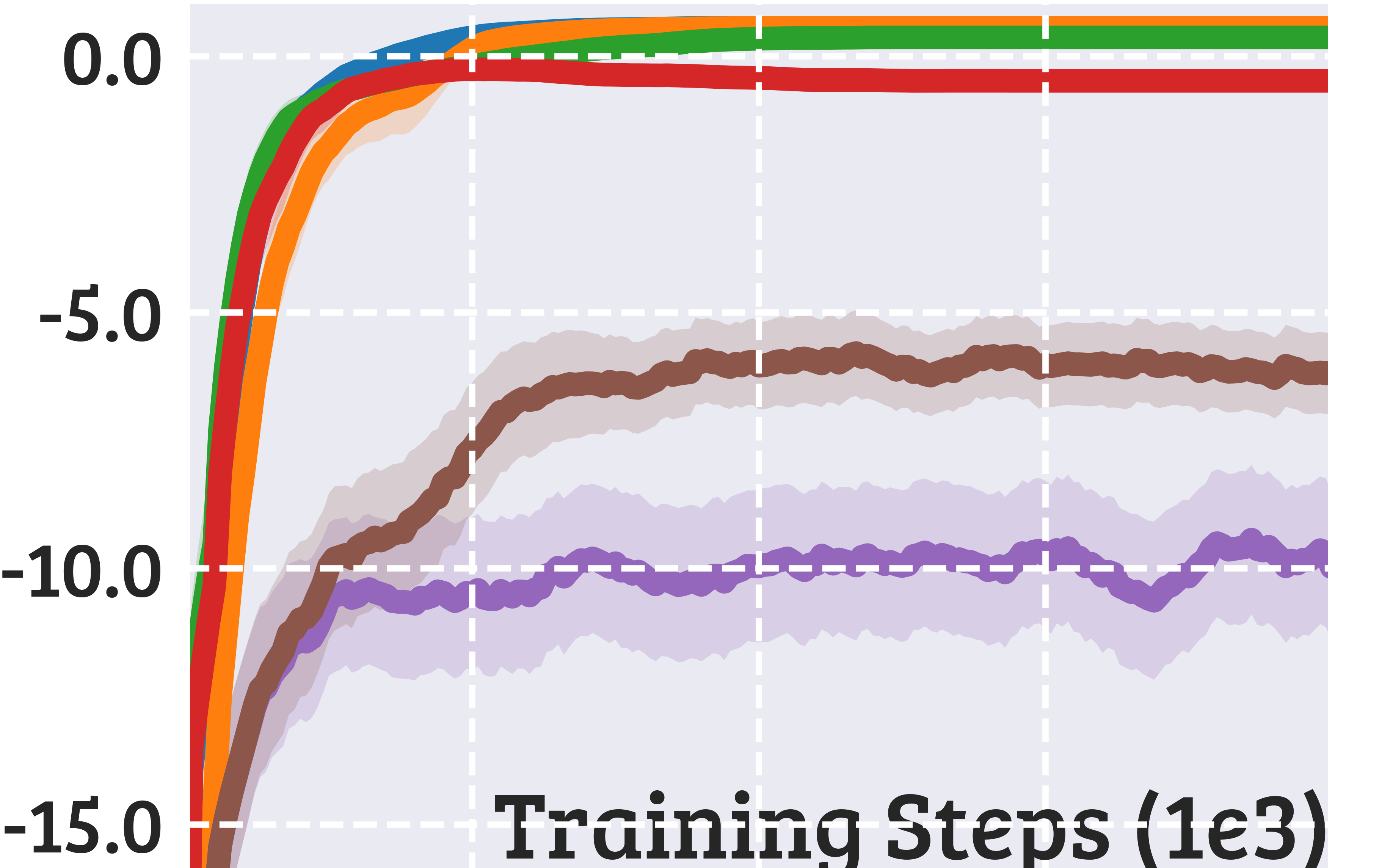}
    \end{subfigure}
    \\[2pt]
    \raisebox{24pt}{\rotatebox[origin=t]{90}{\fontfamily{qbk}\footnotesize\textbf{Init 1}}}
    \hfill
    \begin{subfigure}[b]{0.30\linewidth}
        \centering
        \includegraphics[width=\linewidth]{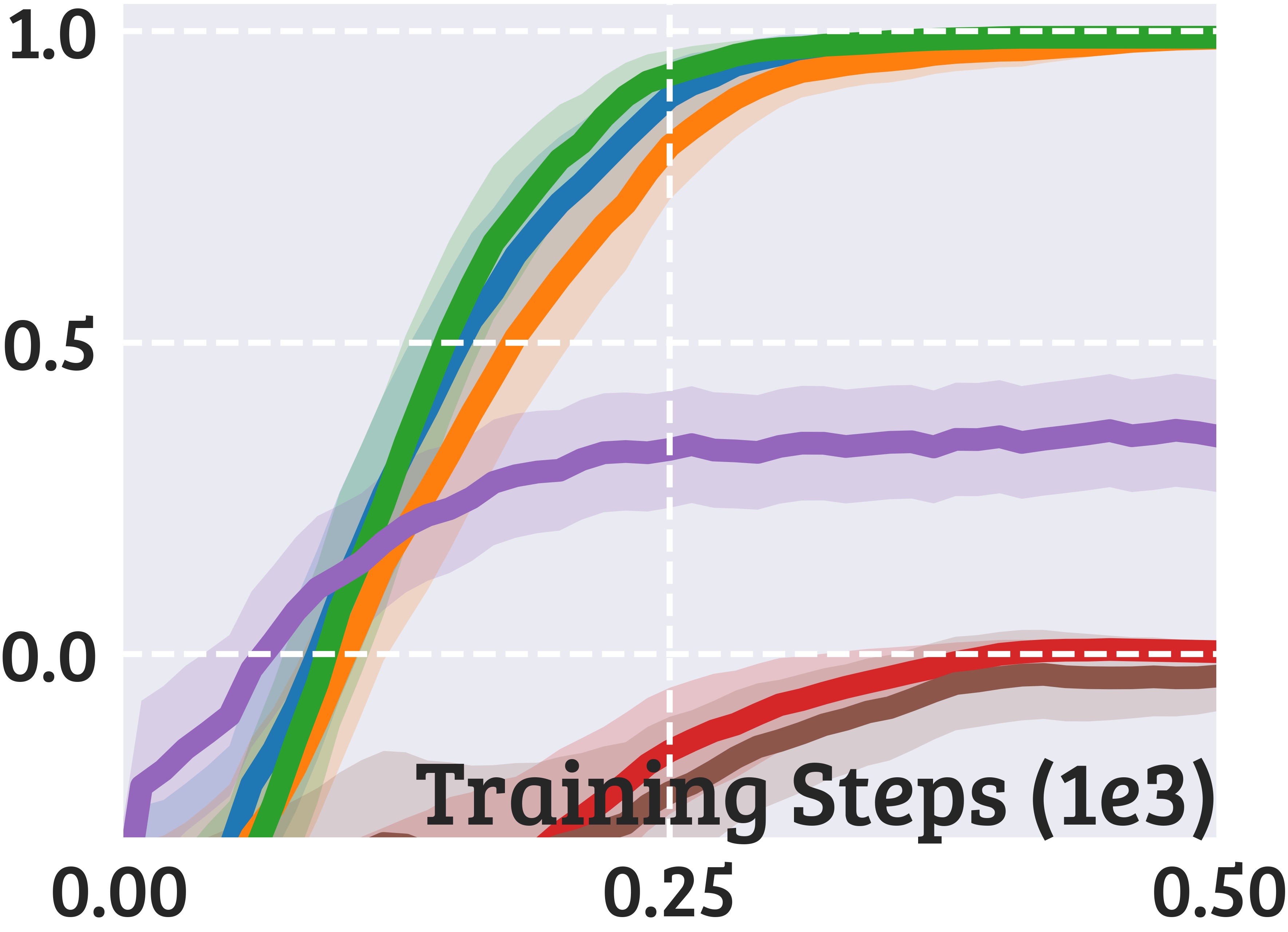}
    \end{subfigure}
    \hfill
    \begin{subfigure}[b]{0.30\linewidth}
        \centering
        \includegraphics[width=\linewidth]{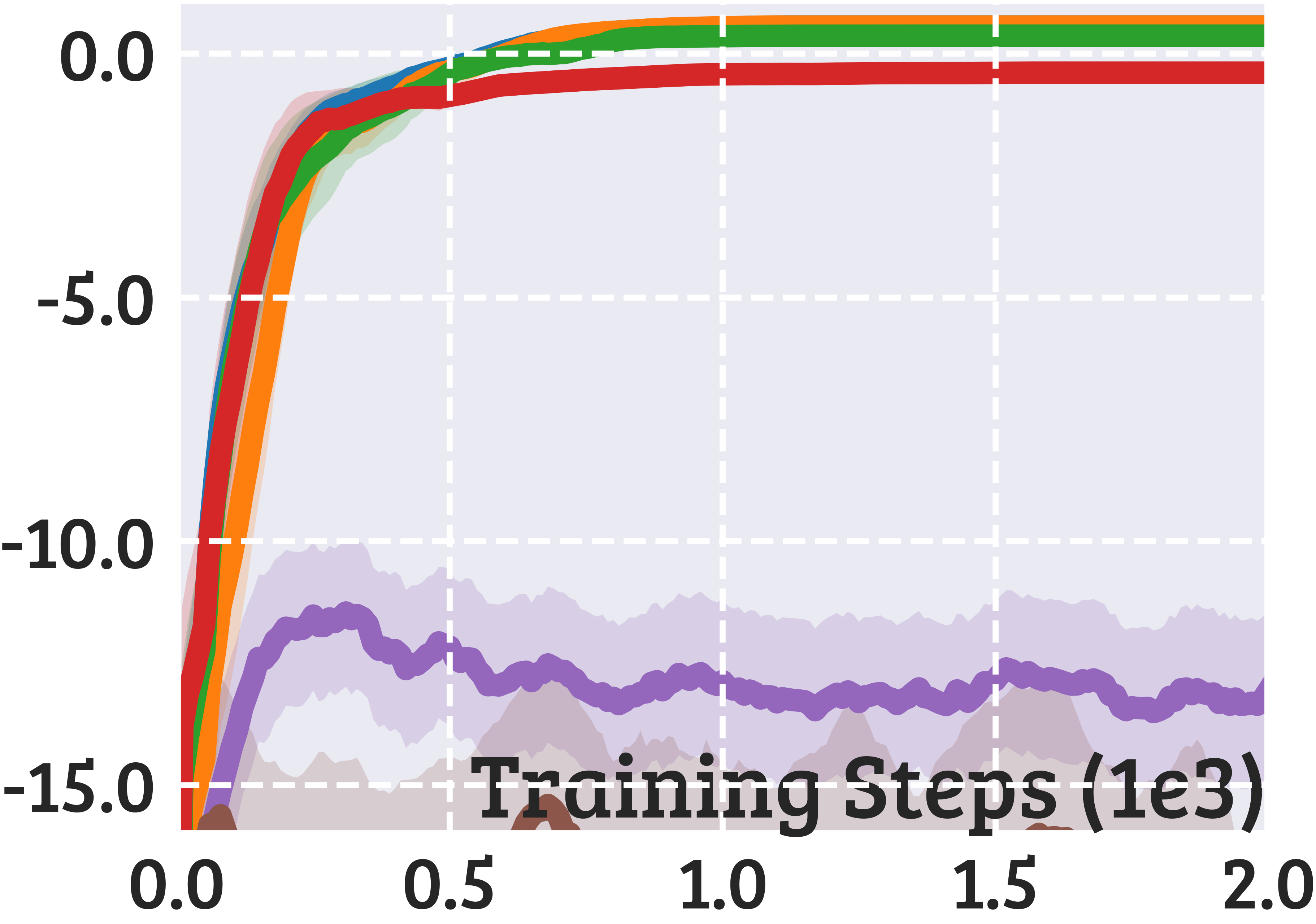}
    \end{subfigure}
    \hfill
    \begin{subfigure}[b]{0.30\linewidth}
        \centering
        \includegraphics[width=\linewidth]{fig/plots/supp/ablation_q0/Gridworld-Medium-3x3-v0_mes50_iToySwitchMonitor_1.0.png}
    \end{subfigure}
    \caption{\label{fig:ablation_q0}\textbf{Expected discounted return on varying the Q-function initialization.} 
    \textmd{%
    Shades denote 95\% confidence interval over 100 seeds.
    Optimistic initialization results in slower convergence and even severely harms learning in \textcolor{joint}{Joint}.}}
\end{figure}

Figure~\ref{fig:ablation_q0} shows that optimistic initialization results in slower rate of convergence. Because the agent's starting position is fixed and the only positive reward is given at the goal, with pessimistic initialization the agent needs to observe such reward only once for the Q-function to greedily guide the agent to the goal.
With optimistic initialization, instead, the agent needs to visit every state-action pair at least once to lower its (overly) optimistic Q-values. Until then, the greedy policy will bring the agent to unvisited states. 

Nonetheless, optimistic initialization does not significantly affect \textcolor{oracle}{Oracle}, \textcolor{model}{Reward Model} and \textcolor{sequential}{Sequential}. In contrast, it highly impacts \textcolor{joint}{Joint} and \textcolor{ignore}{Ignore $\rewardundefined$}. 
\textcolor{joint}{Joint} cannot converge to the optimal policy in the Simple and Penalty Mon-MDPs within 10,000 training steps. As discussed in Section~\ref{app:subsec:joint_counter}, acting greedily over the sum $\qenv + \qmon$ causes unexpected behaviors. This issue is even more prominent when both Q-functions are optimistic initialized, resulting in an extremely slow rate of convergence.\footnote{\textcolor{joint}{Joint} does still converge to the optimal policy in all 100 seeds in the Simple and Penalty Mon-MDPs, but it needs more than 100,000 training steps on average.}
\textcolor{ignore}{Ignore $\rewardundefined$}, instead, converges to a different policy. Because the algorithm does not update the Q-function when it cannot observe rewards, the value of some state-action pairs will never change, and will dictate the behavior of the greedy policy forever. For example, with -10 or 0 initialization, the agent will eventually discover that there are actions with higher values and will learn to reach the goal (always asking for monitoring, though). In contrast, initializing the Q-function optimistically with 1 causes the policy to never ask for monitoring and always act randomly --- the initial value 1 for non-monitoring actions will never be lowered via updates and will always be the highest. 

\end{document}